\documentclass[review]{elsarticle}

\usepackage[caption=false]{subfig}
\usepackage{graphicx}
\usepackage{verbatim} 

\usepackage{algorithm}
\usepackage{algorithmic}
\usepackage{listings}
\usepackage{amsmath}

\journal{Neural Networks}

\begin{document}
\lstset{language=Python}
\begin{frontmatter}

\title{Visualising Basins of Attraction for the Cross-Entropy and the Squared Error Neural Network Loss Functions}

\author[1]{Anna Sergeevna Bosman\corref{myfootnote}}
\ead[myfootnote]{annar@cs.up.ac.za}
\author[2,3]{Andries Engelbrecht}
\author[1]{Mard\'e Helbig}

\address[1]{Department of Computer Science, University of Pretoria, Pretoria, South Africa}
\address[2]{Department of Industrial Engineering, Stellenbosch University,  Stellenbosch, South Africa}
\address[3]{Department of Computer Science, Stellenbosch University,  Stellenbosch, South Africa}

\cortext[myfootnote]{Corresponding author}
\begin{abstract}
Quantification of the stationary points and the associated basins of attraction of neural network loss surfaces is an important step towards a better understanding of neural network loss surfaces at large. This work proposes a novel method to visualise basins of attraction together with the associated stationary points via gradient-based random sampling. The proposed technique is used to perform an empirical study of the loss surfaces generated by two different error metrics: quadratic loss and entropic loss. The empirical observations confirm the theoretical hypothesis regarding the nature of neural network attraction basins. Entropic loss is shown to exhibit stronger gradients and fewer stationary points than quadratic loss, indicating that entropic loss has a more searchable landscape. Quadratic loss is shown to be more resilient to overfitting than entropic loss. Both losses are shown to exhibit local minima, but the number of local minima is shown to decrease with an increase in dimensionality. Thus, the proposed visualisation technique successfully captures the local minima properties exhibited by the neural network loss surfaces, and can be used for the purpose of fitness landscape analysis of neural networks.
\end{abstract}

\begin{keyword}
fitness landscape analysis \sep neural networks \sep cross-entropy \sep squared error \sep local minima \sep loss functions
\end{keyword}

\end{frontmatter}


\section{Introduction}
In the wake of the deep learning research explosion in the artificial neural network (NN) research community~\cite{ref:LeCun:2015,ref:Schmidhuber:2015}, it becomes
increasingly important to develop a better general understanding of NN training as a non-convex optimisation problem. Lack of understanding causes practitioners to make arbitrary choices for various hyperparameters, yielding potentially subpar performance. Failure or success of a particular combination of NN architecture and training
algorithm parameters is hard to predict. Specifically, the nature of the error landscapes associated with the NN loss functions is still poorly understood \cite{ref:Choromanska:2015a, ref:Kordos:2004,
ref:Shen:2016}. There are on-going debates and theories regarding the presence or absence of local minima in NN error landscapes, as well as the properties of stationary points and the associated basins of attraction in the search space~\cite{ref:Dauphin:2014, ref:Kawaguchi:2016, ref:Sagun:2017}. Such lack of understanding hinders the development of new training algorithms that would take the discovered properties of the search space into consideration.

One of the main reasons for this lack of understanding is the inherent high dimensionality of NN problems. High-dimensional spaces are not intuitively visualisable, thus other means of analysis have to be employed. Theoretical analysis, however, often relies on unrealistic assumptions, sometimes causing erroneous conclusions. For example, papers were published claiming that XOR has no local minima~\cite{ref:Hamey:1998}, to be subsequently followed by other publications that explicitly listed all local minima of the XOR problem~\cite{ref:Sprinkhuizen:1999}. Sprinkhuizen et al.~\cite{ref:Sprinkhuizen:1999} have also stated that the listed local minima are in fact saddle points~\cite{ref:Sprinkhuizen:1999}. More recent studies confirm that local optima are indeed present in the NN error landscapes~\cite{ref:Swirszcz:2016}, although saddle points are likely to become more prevalent as the dimensionality of the problem increases~\cite{ref:Dauphin:2014,ref:Choromanska:2015}. Similarly to local minima, the properties of the NN basins of attraction are being actively studied and questioned~\cite{ref:Sagun:2017,ref:Xing:2018}.

The number of local minima, as well as the properties of local minima, were theoretically shown to depend on the chosen error
metric~\cite{ref:Solla:1988}, among other parameters. Solla et
al.~\cite{ref:Solla:1988} analysed two common NN loss functions,
quadratic loss and entropic loss, and came to the conclusion that quadratic loss exhibits a higher density of local minima, and entropic loss has steeper gradients, which is likely to benefit gradient-based training. Entropic loss has gained popularity in the deep learning community due to speeding up gradient descent convergence, and providing more robust results than squared loss~\cite{ref:Glorot:2010, ref:Golik:2013}. However, studies have been published advocating the hybrid use of both entropic and squared loss, as squared loss was shown to be able to refine the solution discovered with entropic loss~\cite{ref:Golik:2013}.

This study aims to explore the properties of the stationary points and the associated basins of attraction exhibited by the NN loss functions. The stationary points of the NN error surfaces are visualised using sampling-based techniques developed for fitness landscape analysis (FLA). Hessian matrix analysis is further employed to classify the discovered stationary points into minima, maxima, and saddles. A simple visualisation method is proposed, which enables intuitive insights into the nature of the error landscapes. Squared loss and entropic loss are shown to exhibit different landscape properties on a selection of classification problems. The experiments correlate well with the current theoretical insights, and provide ground for further hypothesis. 

The novel contributions of this paper are summarised as follows:
\begin{itemize}
  \item A low-dimensional visualisation of the NN stationary points is proposed.
  \item A simple numerical metric to quantify the number and extent of the basins of attraction is proposed.
  \item Squared loss is empirically shown to exhibit more stationary points than entropic loss on a selection of classification problems.
  \item Both loss functions are shown to exhibit a single attractor  at the global optimum on most problems considered, indicating that, regardless of the number of optima present in the search space, the quality of the optima is largely the same.
  \item Empirical evidence is obtained indicating that the global optimum generally does not have good generalisation properties.
  \item Squared loss is shown to be less prone to overfitting
    than the entropic loss function.
\end{itemize}

The rest of the paper is structured as follows:
Section~\ref{sec:minima} reviews the previously published literature on local minima, stationary points, and attraction basins in the NN error landscapes. Section~\ref{sec:error} discusses the two loss functions considered in this study. Section~\ref{sec:fla} describes FLA in the context of NN training problems, and discusses the sampling technique used. Section~\ref{sec:proposed} proposes: (1) a novel method to visualise stationary points and the associated basins of attraction of NN loss surfaces in 2-dimensional space, and (2) two metrics to numerically quantify the discovered basins of attraction. Section~\ref{sec:setup} details the experimental procedure. Section~\ref{sec:results} presents a visual and numerical analysis of stationary points and basins of attraction of the quadratic and the entropic error landscapes. Finally, Section~\ref{sec:conclusions} concludes the paper and proposes some topics for future research.

\section{Local Minima and Basins of Attraction in Neural Networks}\label{sec:minima}
Many studies of local minima in NNs were carried out on the XOR
(exclusive-or) problem. XOR is a simple, but linearly non-separable problem that can be solved by a feedforward NN with at least two hidden neurons. As such, XOR is often used to analyse the basic properties of NNs. Studies of the XOR error landscape are especially interesting, because researchers have arrived at somewhat contradictory conclusions. Hamey~\cite{ref:Hamey:1998} has claimed that the NN error surface associated with XOR has no local minima. A year later, Sprinkhuizen-Kuyper et al.~\cite{ref:Sprinkhuizen:1999, ref:Sprinkhuizen:1999a} have shown that stationary points are present in the XOR NN search space, but that the stationary points are in fact saddle points. A more recent study of the XOR error surface was published by Mehta et al.~\cite{ref:Mehta:2018}, where techniques developed for potential energy landscapes were used to quantify local minima of the XOR problem under a varied number of hidden neurons and regularisation coefficient values. Mehta et al.~\cite{ref:Mehta:2018} have shown that the XOR problem exhibits local minima, and that the number of local minima grows with the increase in the size of the hidden layer.

Further theoretical analysis performed for more complex problems than XOR has highlighted the fact that saddle points are more prevalent in high-dimensional spaces than local minima, and that the number of local minima decreases with an increase in dimensionality~\cite{ref:Dauphin:2014,ref:Choromanska:2015}. Counter examples have also been published, artificially constructing problems with difficult local minima that can potentially trap the training algorithm~\cite{ref:Swirszcz:2016}. Current understanding of the stationary points in NN error surfaces remains incomplete, partially due to the lack of empirical evidence and intuitive visualisations.

The discovery of the prevalence of saddle points in NN error landscapes has lead researchers to question the nature of the basins of attraction associated with the stationary points~\cite{ref:Sagun:2017}. It has been observed that NN error landscapes are comprised of wide and narrow valleys, and that the solutions discovered at the bottom of such valleys may have different generalisation behaviour~\cite{ref:Gallagher:2000, ref:Keskar:2017,ref:Chaudhari:2017}. It has also been observed that it may be possible to find a path of non-increasing error value that connects any two valleys, thus indicating that the valleys may all be part of a single manifold, or attraction basin~\cite{ref:Draxler:2018}. This study estimates the properties of the basins of attraction associated with two different loss functions, namely quadratic and entropic, discussed in the next section.

\section{Loss Functions}\label{sec:error}
The modality of a NN search space, i.e. the number of local minima, as well as the properties of local minima and the associated basins of attraction, were theoretically shown to depend on the chosen error metric~\cite{ref:Solla:1988}, among other parameters. The two most widely used error metrics are the quadratic loss function and the entropic loss function, discussed in this section.

Quadratic loss, also referred to as the sum squared error (SSE),
simply calculates the sum of squared errors produced by the NN:
\begin{equation}E_{sse}=\sum_{p=1}^{P}\sum_{k=1}^{K}(t_{k,p} - o_{k,p})^2\end{equation}
where $P$ is the number of data points, $K$ is the number of outputs, $t_{k,p}$ is the $k$'th target value for data point $p$, and $o_{k,p}$ is the $k$'th output obtained for data point $p$. Minimisation of the SSE minimises the overall error produced by the NN.

If the outputs of the NN can be interpreted as probabilities, then the cross-entropy between two distributions can be calculated, i.e. the distribution of the desired outputs (targets), and the distribution of the actual outputs. Entropic loss, also referred to as the cross-entropy (CE) error, is formulated as follows:
\begin{equation}E_{ce}=\sum_{p=1}^{P}\sum_{k=1}^{K}\big(t_{k,p}\log{o_{k,p}} + (t_{k,p} - 1)\log{(o_{k,p} - 1)}\big)\end{equation}
Minimisation of the cross-entropy leads to convergence of the two distributions, i.e. the actual output distribution resembles the
target distribution more and more, thus minimising the NN error.

Solla et al.~\cite{ref:Solla:1988} analysed quadratic loss and  entropic loss theoretically, and came to the conclusion that quadratic loss exhibits a higher density of local minima. Solla et al.~\cite{ref:Solla:1988} further showed that entropic loss must generate a ``steeper'' landscape with stronger gradients, which may be the reason for the observed faster convergence of gradient descent on CE compared to SSE. Faster convergence of entropic loss has lead to entropic loss becoming more popular than quadratic loss in the deep learning community~\cite{ref:Glorot:2010,ref:Golik:2013}. In addition to faster convergence, entropic loss was shown to exhibit better statistical properties, such as more precise estimation of the true posterior probability on average~\cite{ref:Kline:2005}.

From a theoretical standpoint, however, the global minima of both SSE and CE will correspond to the true posterior probability derived from the given dataset~\cite{ref:Bourlard:1993}. Thus, if a global minimum is found on either of the error landscapes, the quality of either minimum will be equally good. A study by Golik et al.~\cite{ref:Golik:2013} has shown that, although squared loss may cause the training algorithm to converge to a poor minimum, this behaviour is only exhibited if the algorithm was initialised poorly. Golik et al.~\cite{ref:Golik:2013} demonstrated the benefit of applying gradient descent to the error landscape generated by entropic loss at first, and then ``switching'' to quadratic loss to further refine the solution discovered on the entropic loss surface. Such a training scheme may be successful due to the fact that entropic loss is known to turn flat around the global minimum~\cite{ref:Auer:1996}.

\section{Fitness Landscape Analysis}\label{sec:fla}
The concept of fitness landscape analysis (FLA) comes from the
evolutionary context, where quantitative metrics have been developed to study the landscapes of combinatorial problems~\cite{ref:Jones:1995,ref:Merz:2000}. FLA of continuous
fitness landscapes has recently attracted a significant amount of research~\cite{ref:Malan:2009,ref:Malan:2014,ref:Munoz:2015,ref:Sun:2014}. Various fitness landscape properties, such as ruggedness, neutrality, modality, and searchability, can be estimated by taking multiple samples of the search space, calculating the objective function value for every point in each sample, and analysing the relationship between the spatial and the qualitative characteristics of the sampled points. If the samples cover the search space in a meaningful way, the characteristics of the fitness landscape captured by the sampling will apply to the fitness
landscape at large. 

The NN search space is defined as all possible real-valued weight combinations. Thus, samples of the weight space can be taken and analysed to make conclusions about the search space properties. Several studies have been conducted showing FLA to be a useful tool for analysis and visualisation of the NN error surfaces~\cite{ref:Rakitianskaia:2016,ref:Bosman:2016,  ref:Bosman:2017,ref:vanAardt:2017}. However, none of the previous FLA studies have attempted to quantify the modality of the NN error landscapes, i.e. the presence and characteristics of local minima.

One of the simplest FLA approaches to estimate the presence of local minima is to take a uniform random sample of the search space, and then to calculate the proportion of local minima within the sample~\cite{ref:Alyahya:2016}. To identify minima, stationary points need to be identified first. Since the loss functions are differentiable, the gradient can be calculated for each point in the sample. Points with a gradient of zero are stationary points. Stationary points can be further categorised into local minima, local maxima, and saddle points by calculating the eigenvalues of the corresponding Hessian matrix. A positive-definite Hessian is indicative of a local minimum~\cite{ref:Edwards:1973}.

However, an earlier study by Bosman et al.~\cite{ref:Bosman:2018} has demonstrated that random samples capture very few points of high fitness even for such a simple problem as XOR, and thus are unlikely to discover local or global minima. Additionally, random samples do not capture the neighbourhood relationship between individual sample points, which is crucial to the analysis of the basins of attraction. Besides simply identifying the presence or absence of local minima, the possibility of escaping the minima, as well as the structure of the minima, should also be quantified. 

An alternative to a uniform random sample is a sample generated by a random walk. To perform a random walk, a random point is chosen within range, and consecutive steps in randomised directions are taken to generate the sample. This way, the sampled points will be related to each other topographically. However, a random walk faces the same problem as the uniform random sample; in that random traversal of the search space provides no guarantee of locating areas of good fitness~\cite{ref:Bosman:2018}.

Instead of analysing random walks, the trajectory of a training algorithm can be analysed. However, such an approach will make the observations biased towards the performance of the specific algorithm under specific hyperparameter values. To address this problem, Bosman et al.~\cite{ref:Bosman:2018} proposed an algorithm-agnostic sampling method called a progressive gradient walk. A progressive gradient walk uses the numeric gradient of the loss function to determine the direction of each step. The size of the step is randomised per dimension within predefined bounds. The advantage of this sampling approach is that gradient information is combined with stochasticity, preventing convergence, yet guiding the walk towards areas of higher fitness.

For combinatorial landscapes, application of local search to a selection of random starting points, and then analysing the best fitness solutions as discovered by the local search from different starting points, has been proposed as a method to estimate the total number of optima and the size of the corresponding basins of attraction~\cite{ref:Garnier:2001, ref:Garnier:2001b}. The method was adapted to continuous spaces~\cite{ref:Caamano:2010,ref:Caamano:2013}, and involves performing local searches with gradient-based updates, and then calculating the Euclidean distance between discovered minima to determine whether the minima belong to the same basin of attraction. However, it has been demonstrated by Morgan and Gallagher~\cite{ref:Morgan:2014} that using Euclidean distance in high-dimensional spaces can be very misleading, since all distances begin to look the same when the dimensionality of the problem is high enough. Due to the inherently high dimensionality of NN error landscapes, metrics based on Euclidean distance must be avoided. Additionally, Sagun et al.~\cite{ref:Sagun:2017} have challenged the notion of multiple basins of attraction in NN error landscapes, illustrating through Hessian matrix analysis that minima that lie away from one another may still be connected by a flat region, and should thus be treated as members of the same basin.

This study applies progressive gradient walks to sample NN error landscapes across multiple classification problems for the two different error metrics discussed in Section~\ref{sec:error}. The
obtained samples are studied to empirically establish and quantify the presence and structure of local minima and the associated basins of attraction in NN error landscapes. 

\section{Fitness Landscape Analysis Techniques for Visualisation and Quantification of Neural Network Attraction Basins}\label{sec:proposed}
This section describes the two novel FLA techniques proposed in this study for the purpose of visualisation and quantification of the stationary points and the associated basins of attraction exhibited by NN loss surfaces. Section~\ref{subsec:hessian} introduces the loss-gradient clouds, which offer a 2-dimensional visualisation of the stationary points discovered by the gradient walks. Section~\ref{subsec:basins} proposes two metrics to quantify the properties of the basins of attraction encountered by the gradient walks.

\subsection{Loss-Gradient Clouds}\label{subsec:hessian} 
A simple way to visualise stationary points discovered by a search trajectory is to plot the error, or loss values, against the corresponding gradient in a 2-dimensional scatterplot,  referred to as the loss-gradient cloud, or l-g cloud. All points of zero gradient are stationary points. Stationary points of non-zero loss can be either local minima, local maxima, or saddle points. To determine if a particular stationary point is a local minimum, local maximum, or a saddle point, local curvature information can be derived from the eigenvalues of the corresponding Hessian matrix. If the eigenvalues of the Hessian are positive, the point is a maximum. If the eigenvalues are negative, the point is a minimum. If the eigenvalues are positive as well as negative, the point is a saddle. If any of the eigenvalues are zero, i.e. if the Hessian is indefinite, the test is considered inconclusive.

The main benefit of l-g clouds is the 2-dimensional representation of the high-dimensional search space. Studying the discovered stationary points in 2-dimensional space allows the identification of the total number of attractors corresponding to different loss values. The gradient behaviour of the attractors is also visualised by the l-g clouds, and can provide useful insights. Since the distance between sampled points is not represented in the l-g clouds, the actual number of distinct local minima and other attractors cannot be estimated using this technique.

\subsection{Quantifying Basins of Attraction}\label{subsec:basins}
A progressive gradient walk samples the search space by taking randomised steps in the general direction of the steepest gradient descent. If a step taken in the direction of the negative gradient is too large, the step may miss the area of low error, and result in an area of higher error. Thus, a progressive gradient walk will not necessarily produce a sequence of points with strictly non-increasing error values. In fact, any gradient-based sample or algorithm trajectory is likely to exhibit oscillatory behaviour if the gradient in some dimensions is significantly steeper than in others \cite{ref:Rumelhart:1985}.

Even though the gradient step sequence will not necessarily be strictly decreasing in error, the sample is nonetheless expected to travel in the general direction of the global minimum. The areas of the landscape where a gradient-based walk oscillates or otherwise fails to reduce the error for a number of steps are the stationary areas of the search space that may hinder the optimisation process. Quantifying the number and extent of such areas will provide an indication of the ``difficulty'' of the search space, as well as an empirical estimate of the landscape modality. Thus, an important error landscape property to estimate is the number of times that the
sampling algorithm will get ``stuck'' along the way. 

To smooth out the potential oscillations of the sample, an exponential moving average of the sample can be calculated. An exponentially weighted moving average (EWMA)~\cite{ref:Winters:1960} is a smoothing filter commonly used for time series prediction. EWMA calculates the moving average for each step in the time series by taking all previous steps into account, and assigning exponentially decaying weights to the previous steps, such that the weight for each older step in the series decreases exponentially, never reaching zero. Given a sequence $T = \{T_i\}_{i = 1}^Z$, the EWMA-smoothed sequence $T'$ is given by:
\begin{equation}
 T'_i = 
  \begin{cases} 
   T_i & \text{if } i = 1 \\
   \alpha \cdot T_i + (1 - \alpha)\cdot T'_{i-1} & \text{if } i > 1
  \end{cases}
\end{equation}
The decay coefficient $\alpha\in[0,1]$ determines the degree of
smoothing, where larger values of $\alpha$ facilitate faster decay and weaker smoothing, and smaller values of $\alpha$ facilitate slower decay and thus stronger smoothing.

To identify the sections of the sample where the behaviour is stagnant, the standard deviation of the smoothed sample is calculated first. Then, a sliding window approach is used to generate a sequence of the moving standard deviations of the sample. If the standard deviation of the values in the current window is less than the standard deviation of the entire sample for a number of steps, then these steps can be said to form a stagnant sequence. The average number of stagnant regions encountered per sample, $n_{stag}$, and the average length of the stagnant regions, $l_{stag}$, can be used to quantify the number and size of the basins of attraction present in the search space.

\begin{figure}
  \begin{subfloat}
    [Window of 6\label{fig:example:6}]{\includegraphics[height=1.7in]{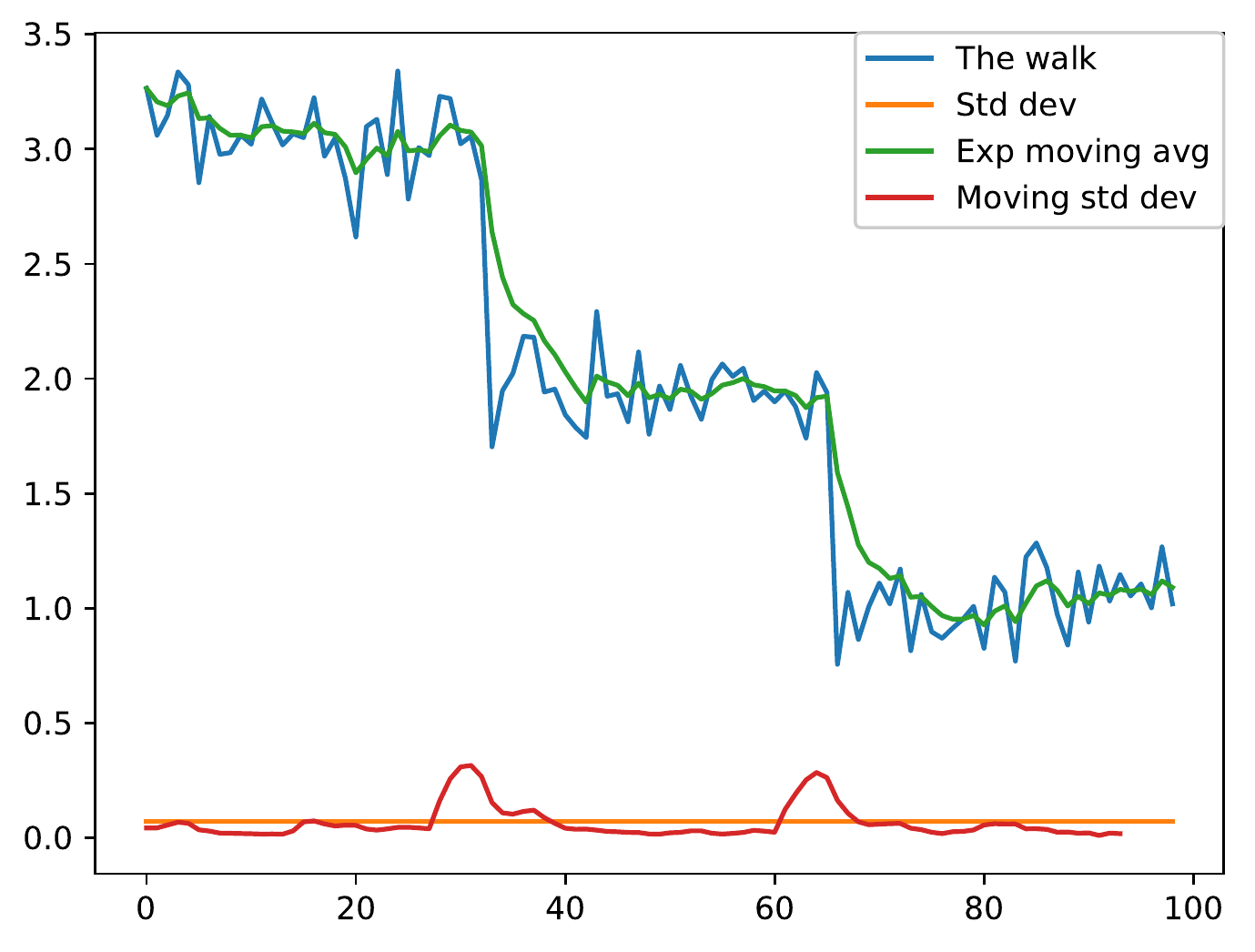}}
  \end{subfloat}
  \begin{subfloat}
    [Window of 8\label{fig:example:8}]{\includegraphics[height=1.7in]{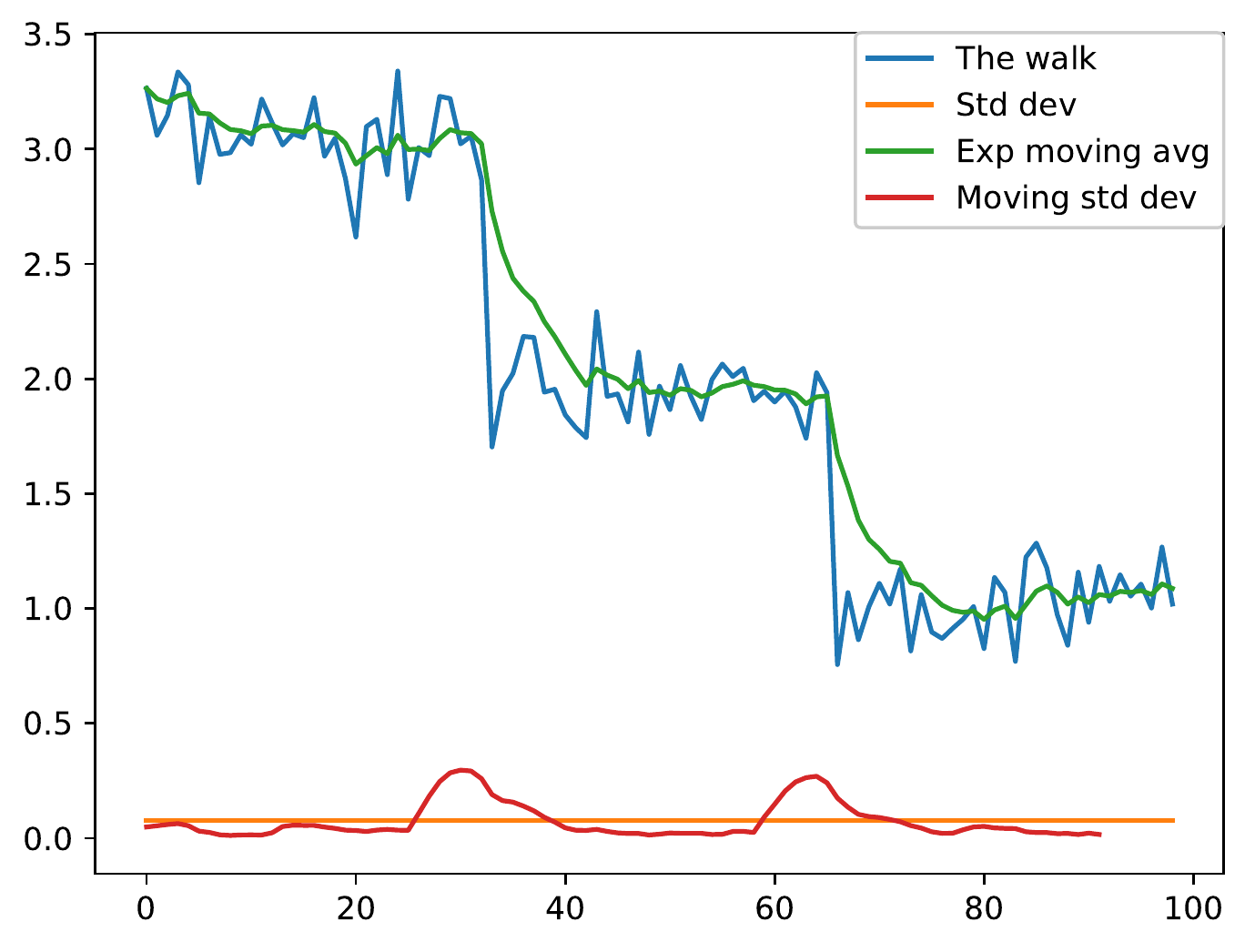}}
  \end{subfloat}\\
  \begin{subfloat}
    [Window of 12\label{fig:example:12}]{\includegraphics[height=1.7in]{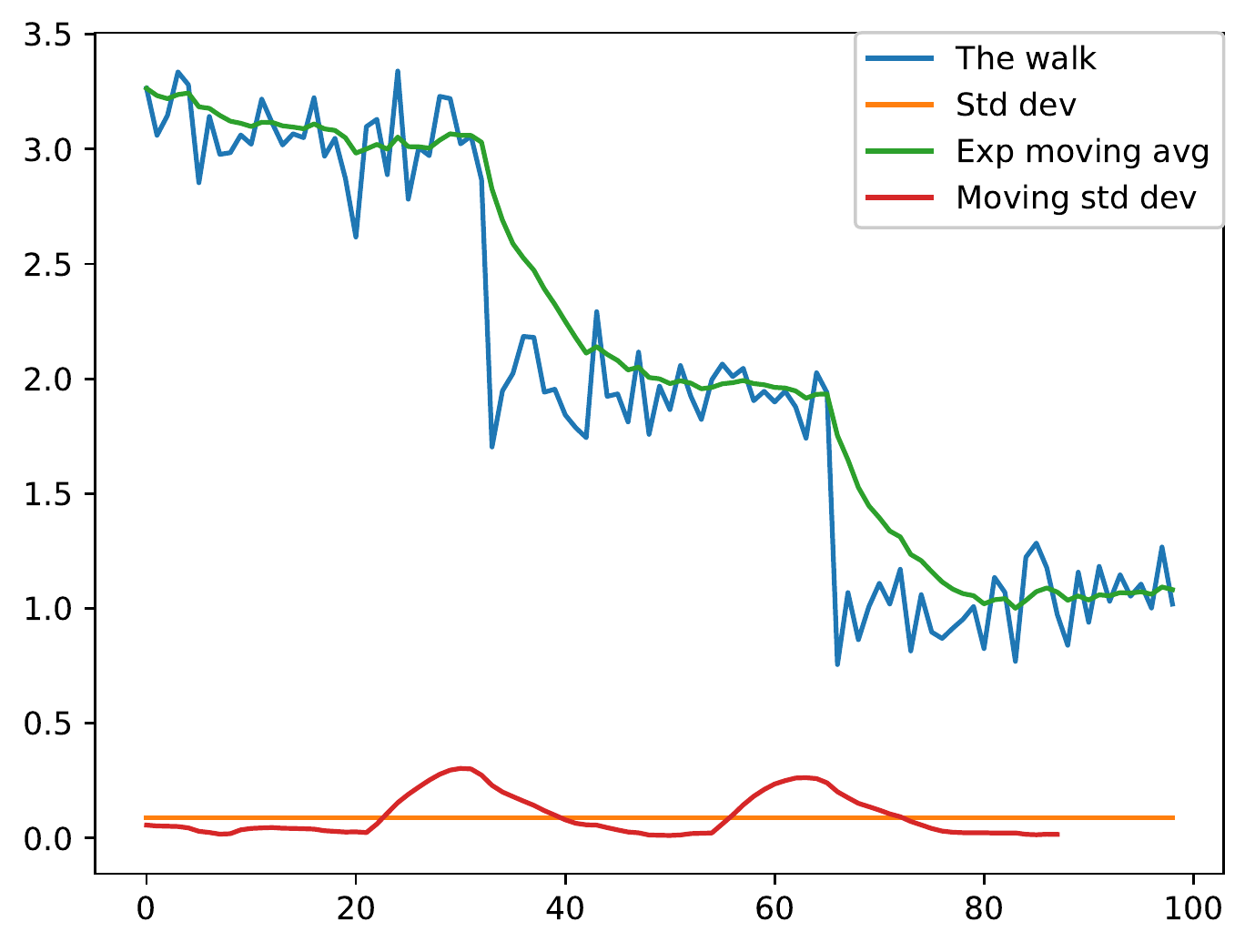}}
  \end{subfloat}
  \begin{subfloat}
    [Window of 20\label{fig:example:20}]{\includegraphics[height=1.7in]{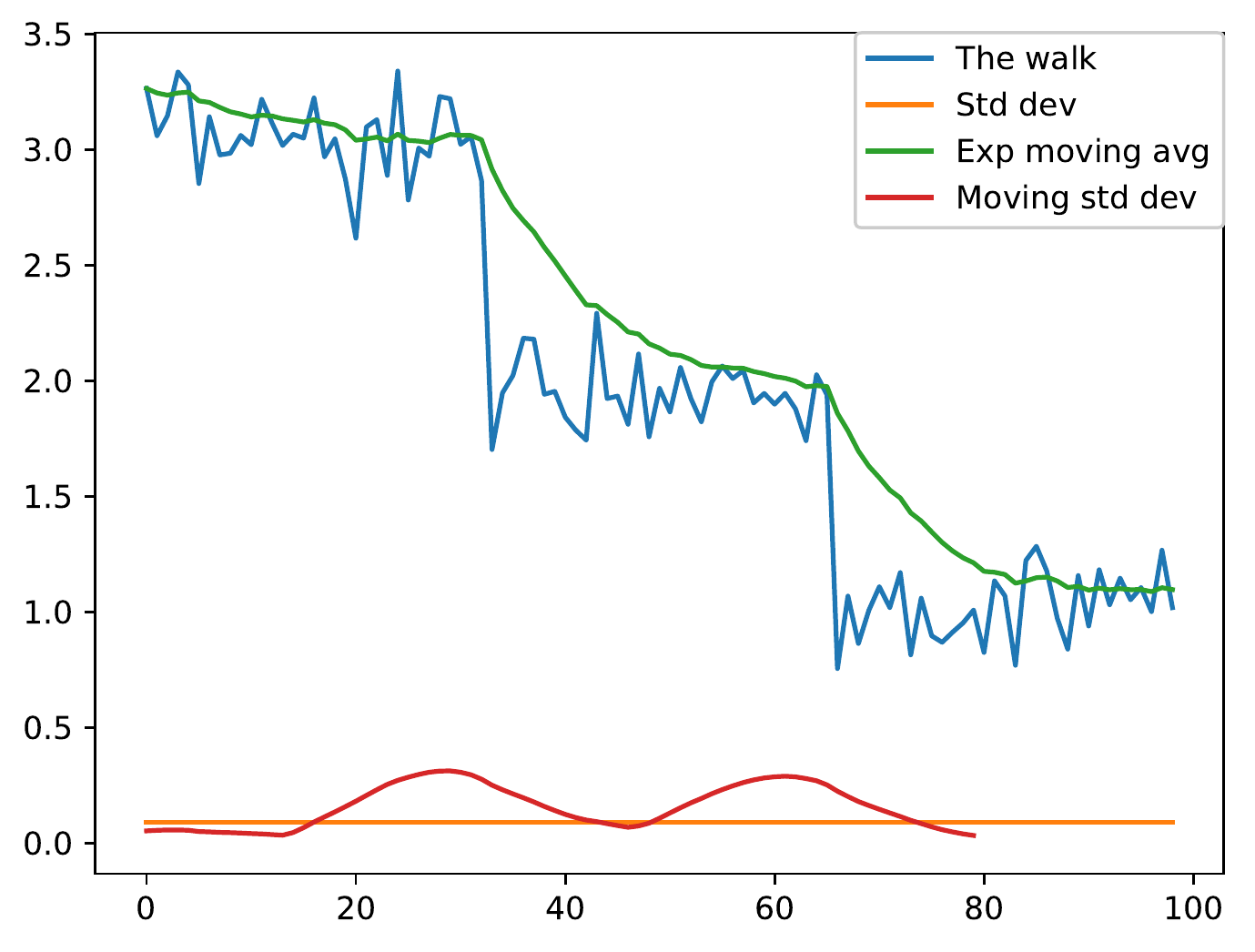}}
  \end{subfloat}
  \caption{Illustration of the proposed technique to estimate the
    number and extent of the basins of
    attraction. Figures~\ref{fig:example:6} to~\ref{fig:example:20}
    show the effect of window size on the sample
    smoothing.}\label{fig:example}
\end{figure}

The proposed approach is illustrated in Figure~\ref{fig:example}. The simulated walk oscillates around three different error values. The moving standard deviation line dips below the all-sample standard deviation threshold three times, which corresponds to the three simulated stagnant areas.

Figure~\ref{fig:example} illustrates that the window size has a significant effect on the attraction basin estimates: too little smoothing (Figure~\ref{fig:example:6}) may cause fluctuations to be
perceived as stationary regions. Excessive smoothing, on the other hand (Figure~\ref{fig:example:20}), may fail to detect all
stationary regions. Therefore, the window size has to be optimised per sample. If the sequence contains oscillations, then too little
smoothing will cause multiple ``spikes'' in the walk to be regarded as areas of stagnation. These short bursts of ``stagnation'' will yield a small average basin length, $l_{stag}$. If the sequence is smoothed excessively, the sample will start to resemble a wave more and more, perceiving flat areas as areas with an incline, which will once again cause the $l_{stag}$ to decrease. Thus, too little as well as too much smoothing will shrink the $l_{stag}$. Therefore, the window size $w$ can be optimised by maximising the $l_{stag}$ value. Table~\ref{table:lstag:example} lists $l_{stag}$ values obtained on the simulated walk shown in Figure~\ref{fig:example} under various values of $w$. Table~\ref{table:lstag:example} shows that $l_{stag}$ reaches its maximum for $w = 8$, and decreases for smaller, as well as larger, values of $w$.

\begin{table}[htb]
  \caption{Effect of window size $w$ on $l_{stag}$}
  \label{table:lstag:example}
  \centering
    \begin{tabular}{|l|c|c|c|c|c|c|c|c|}\hline
      $w$       & 6 & 8 & 10 & 12 & 14 & 16 & 18 & 20 \\\hline
      $l_{stag}$ & 18.75 & {\bf 22.0} & 20.67 & 18.0 & 16.33 & 14.33 & 14.0 & 12.0\\
      \hline
    \end{tabular}
\end{table}  
The window size $w$ can therefore be automatically optimised by calculating $l_{stag}$ over a range of $w$ values, and picking the value of $w$ that yields the highest $l_{stag}$ value. In this study, $w$ is optimised by successively applying $w \in \{6, 8, \dots, 18, 20\}$. Given a window of size $w$, the EWMA value of $\alpha$ is calculated as $\alpha = {2}/({w + 1})$. The $w$ value yielding the largest $l_{stag}$ is subsequently used for the final $l_{stag}$ and $n_{stag}$ estimates.

Thus, two estimates to quantify the basins of attraction are proposed:
\begin{enumerate}
\item The average number of times that stagnation was observed, $n_{stag}$.
\item The average length of the stagnant sequence, $l_{stag}$.
\end{enumerate}
Pseudocode to calculate $n_{stag}$ and $l_{stag}$ is provided
in~\ref{appendix:lstag}.

\section{Experimental Procedure}\label{sec:setup}
The aim of the study was to visually and numerically investigate the
local minima and basins of attraction exhibited by quadratic and
entropic loss functions. This section discusses the experimental
set-up of the study, and is structured as follows:
Section~\ref{sub:nn} describes the NN hyperparameters employed in the experiments; Section~\ref{sub:bench} lists the benchmark problems used; Section~\ref{sub:walks} outlines the sampling algorithm parameters, and the data recorded for each sampled point.

\subsection{Neural network hyperparameters}\label{sub:nn}
All experiments employed feed-forward NNs with a single hidden
layer. The sigmoid activation function was used in the experiments,
given by $f_{NN}(net)=1/(1+e^{-net})$, where $net$ is the sum of
weighted inputs. While the choice of activation function has an effect on the resulting error landscape, the aim of this study was to investigate the difference between quadratic and entropic loss.

\subsection{Benchmark problems}\label{sub:bench}
A selection of well-known classification problems of varied
dimensionality were used in this study. Table~\ref{table:benchmarks}
summarises the NN architecture parameters used for each dataset, as
well as the total dimensionality of the weight space. The specified
sources point to publications from which each dataset and/or NN
architectures were adopted.

\begin{table}[!htb]
\renewcommand{\arraystretch}{1.3}
\caption{Benchmark Problems}
\label{table:benchmarks}
\centering
\begin{tabular}{l|ccc|c|c}
\hline
{\bf Problem} & {In} & {Hidden} & {Out} & {Dim.} & {Source}\\
\hline
XOR & 2 & 2 & 1 & 9& \cite{ref:Hamey:1998} \\
Iris & 4 & 4 & 3 & 35 &\cite{ref:Fisher:1936}  \\
Diabetes& 8 & 8 & 1 & 81 &\cite{ref:Prechelt:1994} \\
Glass& 9 & 9 & 6 & 150 & \cite{ref:Prechelt:1994}  \\
Cancer& 30 & 10 & 1 & 321 & \cite{ref:Prechelt:1994}\\
Heart & 32 & 10 & 1 & 341 & \cite{ref:Prechelt:1994}\\
MNIST & 784 & 10 & 10 & 7960 & \cite{ref:LeCun:2010}\\
\hline
\end{tabular}
\end{table}

The properties of each dataset are briefly discussed below:
\begin{enumerate}
\item{\bf XOR:} XOR (exclusive-or) is a simple, but linearly non-separable problem that can be solved by a feedforward NN with at least two hidden neurons. As such, XOR is often used to analyse basic properties of artificial neural networks. The dataset consists of 4 binary patterns. 
\item{\bf Iris:} The famous Iris flower data set \cite{ref:Fisher:1936} contains 50 specimens
from each of the three species of iris flowers, i.e. {\it Iris Setosa}, {\it Iris Versicolor}, and {\it Iris Virginica}. There are 150 patterns in the dataset.
\item{\bf Diabetes:} 
The diabetes dataset \cite{ref:Prechelt:1994} captures personal data
of 768 Pima Indian patients, classified as diabetes positive or
diabetes negative.
\item{\bf Glass:}
The glass dataset \cite{ref:Prechelt:1994} captures chemical
components of glass shards. Each glass shard belongs to one of six
classes: float processed or non-float processed building windows,
vehicle windows, containers, tableware, or head lamps. There are 214
patterns in the dataset.
\item{\bf Cancer:}
The breast cancer Wisconsin (diagnostic) dataset~\cite{ref:Prechelt:1994} consists of 699 patterns, each containing tumor descriptors, and a binary classification into benign or malignant.
\item{\bf Heart:}
The heart disease prediction dataset~\cite{ref:Prechelt:1994} contains
920 patterns, each describing various patient descriptors.
\item{\bf MNIST:}
The MNIST dataset of handwritten digits~\cite{ref:LeCun:2010} contains
70,000 examples of grey scale handwritten digits from $0$ to $9$. For
the purpose of this study, the 2-dimensional input is treated as a
1-dimensional vector.
\end{enumerate}
Input values for all problems except XOR have been standardised by
subtracting the mean per input dimension, and scaling every input
variable to unit variance. All outputs were binary encoded for
problems with two output classes, and one-hot binary encoded for
problems with more than two output classes.

\subsection{Sampling parameters}\label{sub:walks}
For the purpose of sampling the areas of low error, a progressive
gradient walk, discussed in Section~\ref{sec:fla}, was used as the
sampling mechanism. To allow for adequate coverage of the search
space, the number of independent walks was set to be one order of
magnitude higher than the dimensionality of the problem, i.e. for a
problem of $d$ dimensions, $10\times d$ independent progressive
gradient walks were performed. The walks were not restricted by search space bounds, however, two different initialisation ranges were considered, namely $[-1,1]$ and $[-10,10]$. The smaller range is typically used for NN weight initialisation. The larger range is likely to contain high fitness solutions~\cite{ref:Bosman:2016}. Since the granularity of the walk, i.e. the average step size, has a bearing on the resulting FLA metrics~\cite{ref:Malan:2009}, two granularity settings were used throughout the experiments: micro, where the maximum step size was set to 1\% of the initialisation range, and macro, where the maximum step size was set to 10\% of the initialisation range. Micro walks performed 1000 steps each, and macro walks performed 100 steps each.

For all problems except the XOR problem, the dataset was split into
80\% training and 20\% test subsets. The training set was used to
calculate the direction of the gradient, as well as the error of the
current point on the walk. The test set was used to evaluate the
generalisation ability of each point in the walk. To calculate the
training and the generalisation errors, the entire train/test subsets were used for all problems except MNIST. For MNIST, random batches of 100 patterns were sampled from the respective training and test sets.

In order to identify stationary points discovered by the gradient walks, the magnitude of the gradient vector was recorded for each step together with the loss value. Additionally, the eigenvalues of the Hessian matrix were calculated for each step, and used to classify each step as convex, concave, saddle, or singular.

\section{Empirical Results}\label{sec:results}
This section presents the analysis of apparent local minima and the
corresponding basins of attraction as captured by the progressive
gradient walks. The results obtained for each problem are discussed
separately.

\subsection{XOR}
\begin{figure}
\begin{center}
  \begin{subfloat}[{SSE, micro, $[-1,1]$ }\label{fig:xor:b1:mse:micro}]{    \includegraphics[width=0.8\textwidth]{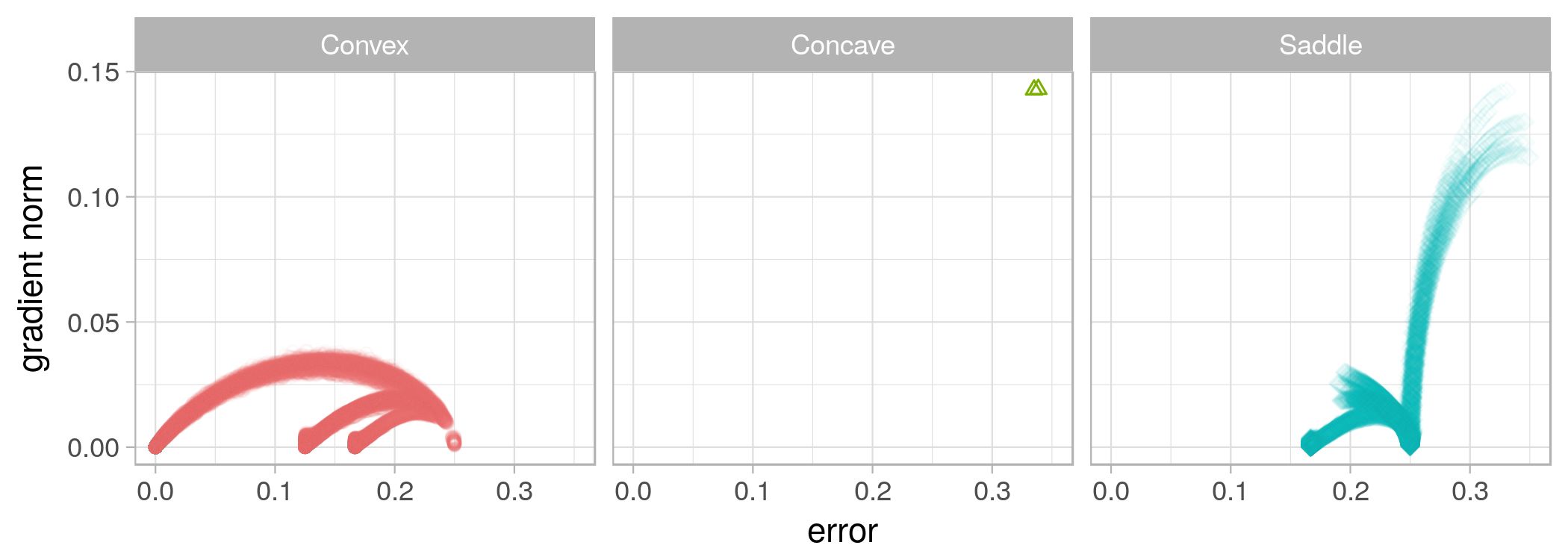}}
  \end{subfloat}
  \begin{subfloat}[{CE, micro, $[-1,1]$ }\label{fig:xor:b1:ce:micro}]{    \includegraphics[width=0.8\textwidth]{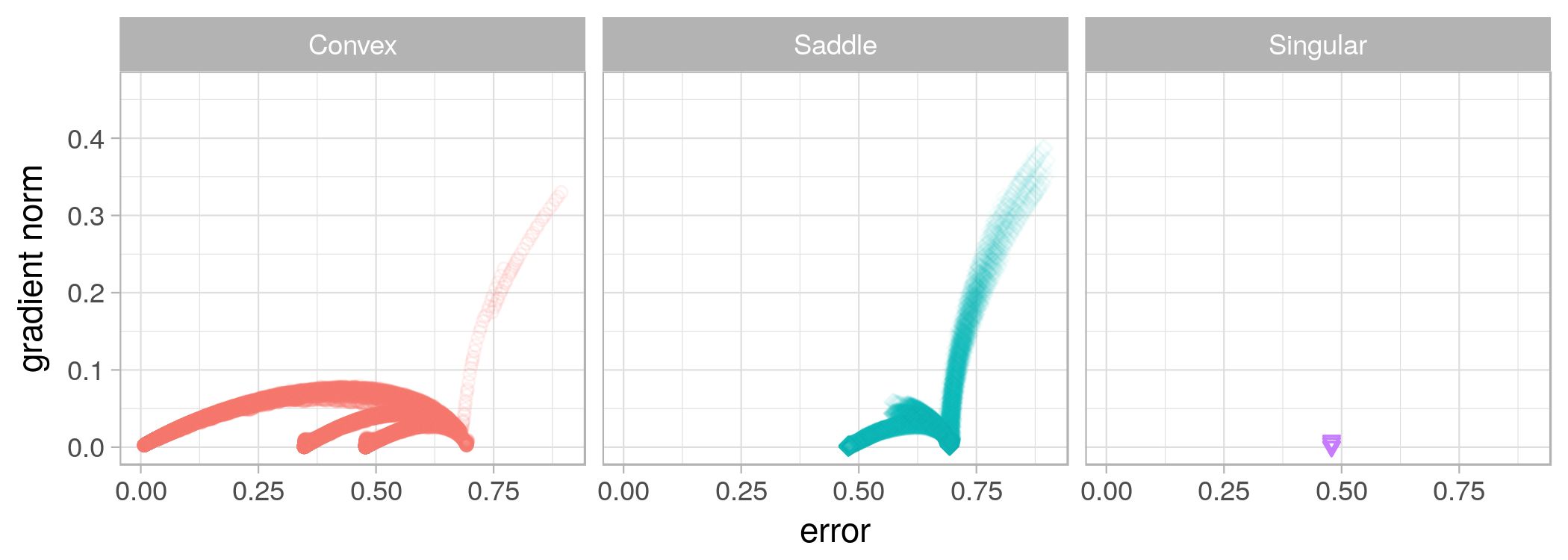}}
  \end{subfloat}
  \begin{subfloat}[{SSE, macro, $[-1,1]$ }\label{fig:xor:b1:mse:macro}]{    \includegraphics[width=0.8\textwidth]{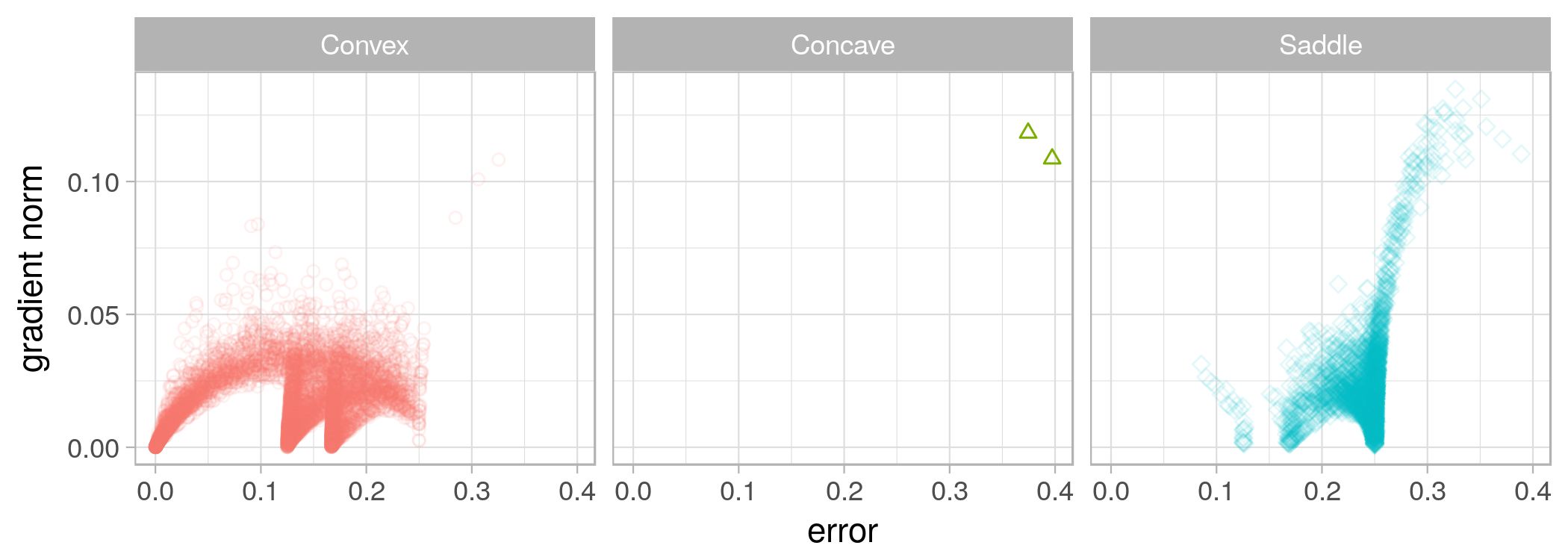}}
  \end{subfloat}
  \begin{subfloat}[{CE, macro, $[-1,1]$ }\label{fig:xor:b1:ce:macro}]{    \includegraphics[width=0.8\textwidth]{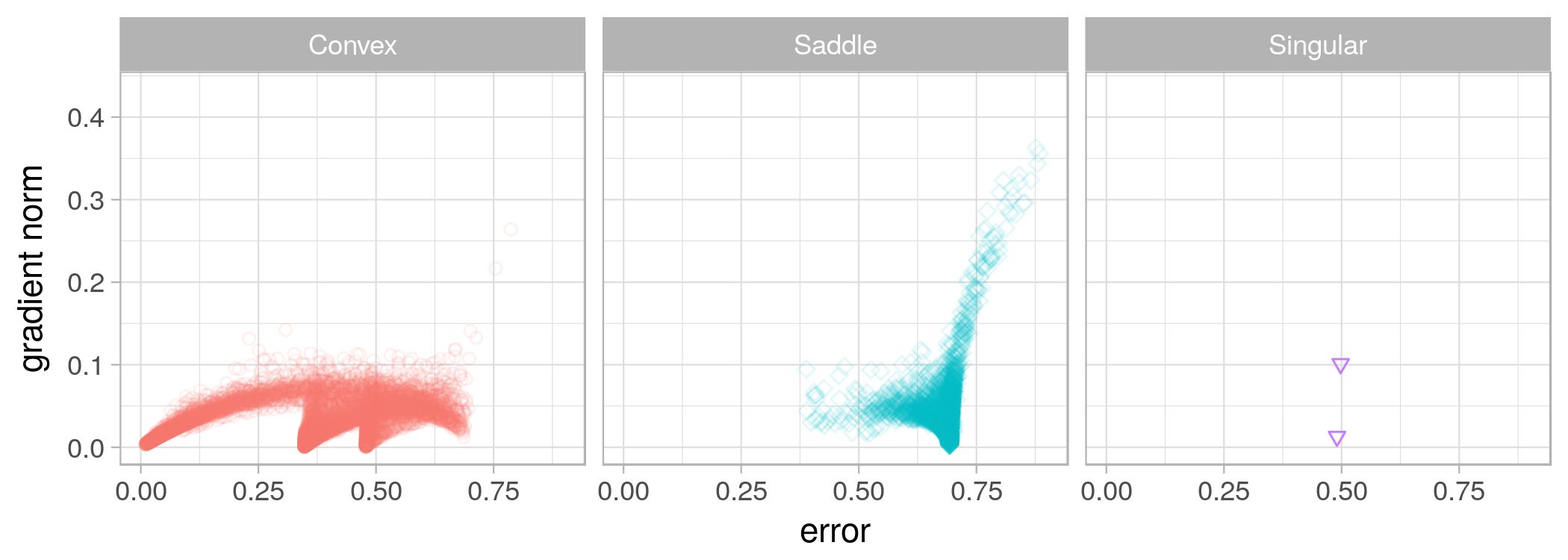}}
  \end{subfloat}
  \caption{L-g clouds for the gradient walks initialised in the $[-1,1]$ range for the XOR problem.}\label{fig:xor:b1}
\end{center}
\end{figure}

\begin{figure}
\begin{center}
  \begin{subfloat}[{SSE, micro, $[-10,10]$ }\label{fig:xor:b10:mse:micro}]{    \includegraphics[width=0.8\textwidth]{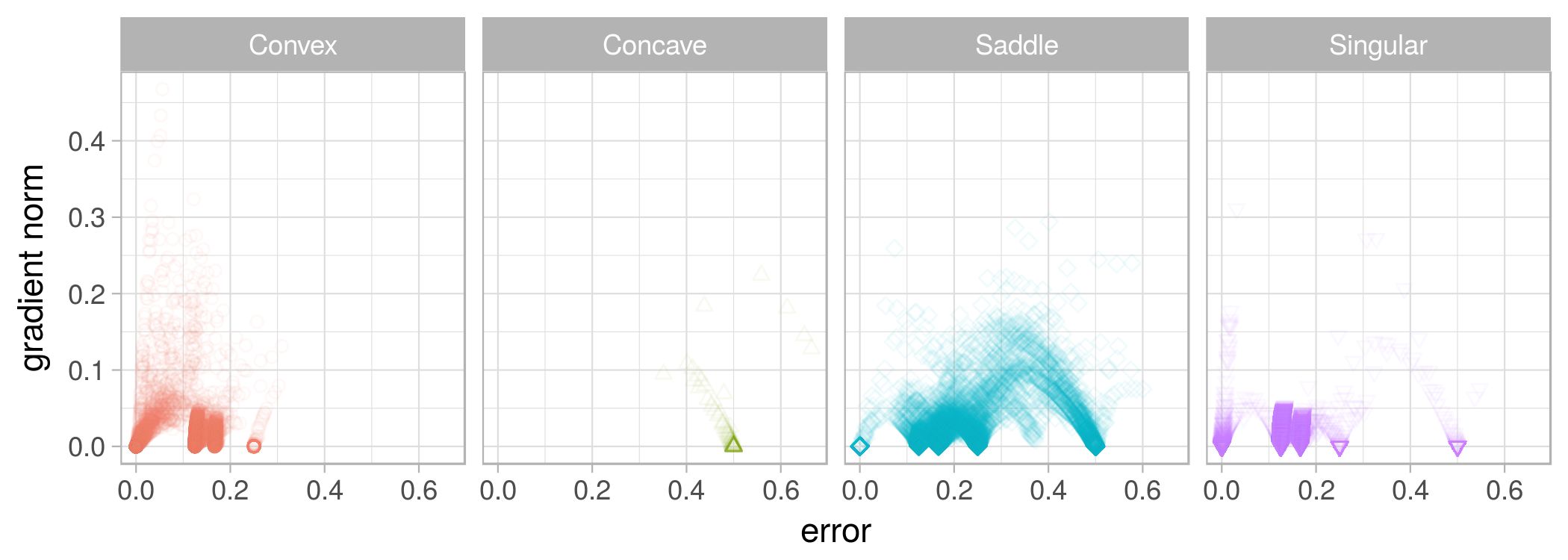}}
  \end{subfloat}
  \begin{subfloat}[{CE, micro, $[-10,10]$ }\label{fig:xor:b10:ce:micro}]{    \includegraphics[width=0.8\textwidth]{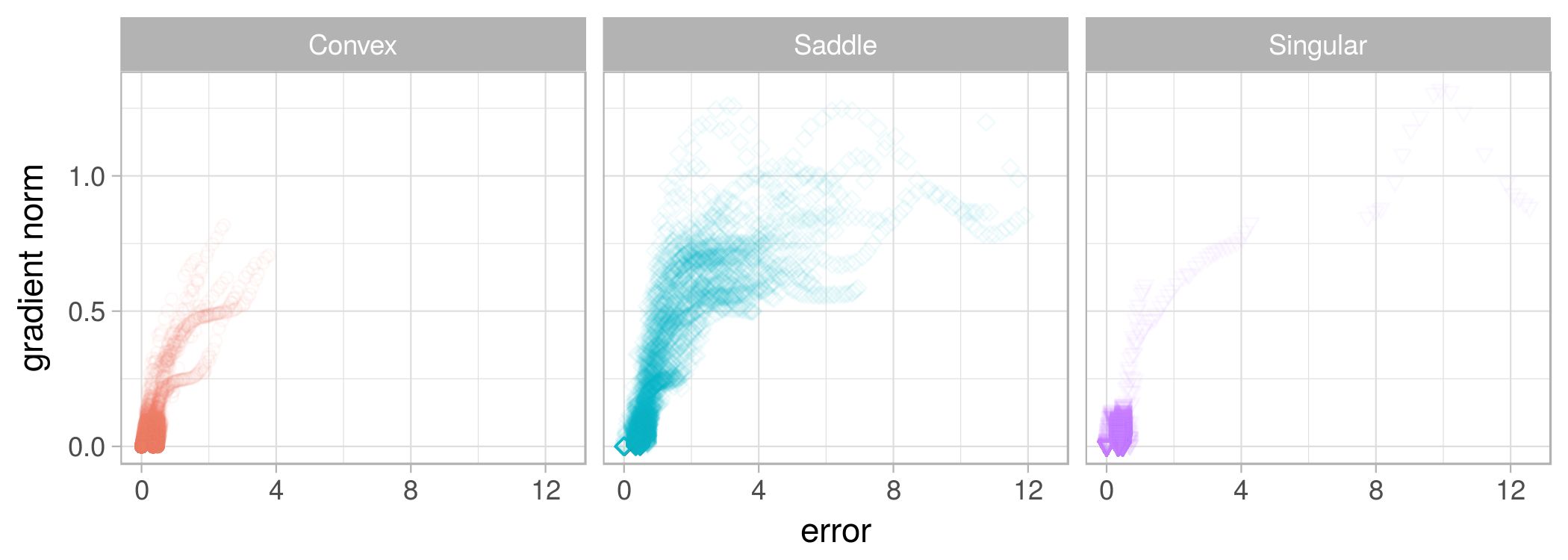}}
  \end{subfloat}
  \begin{subfloat}[{SSE, macro, $[-10,10]$ }\label{fig:xor:b10:mse:macro}]{    \includegraphics[width=0.8\textwidth]{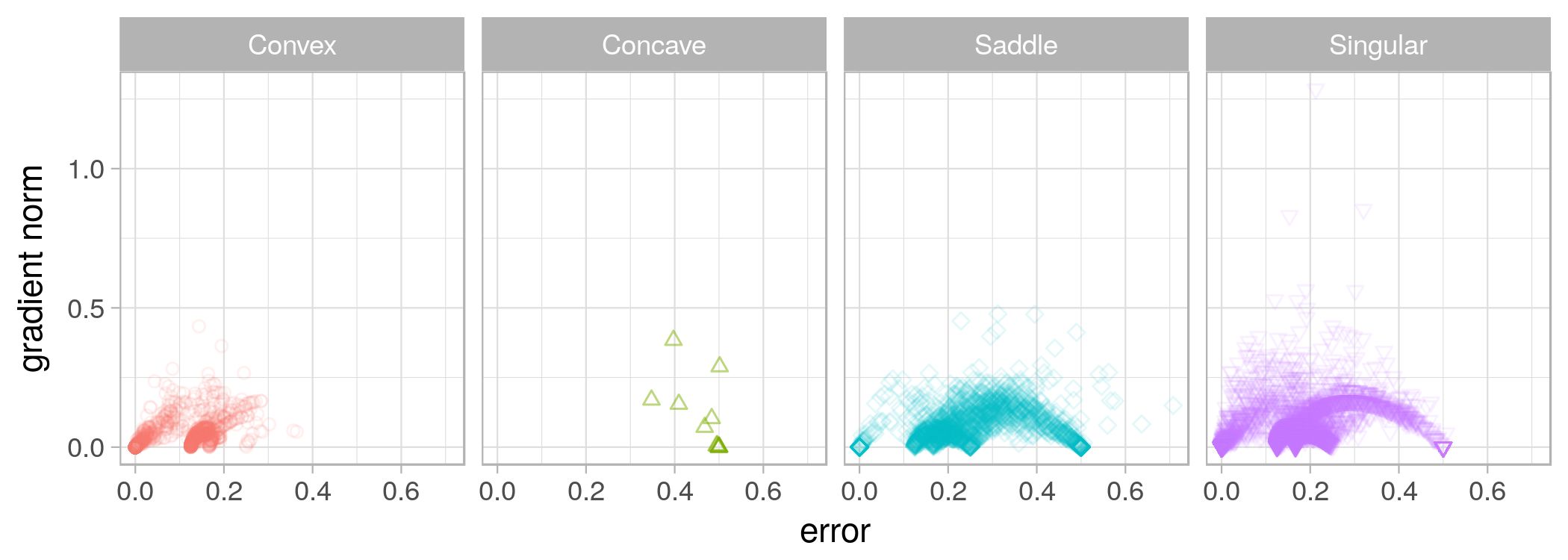}}
  \end{subfloat}
  \begin{subfloat}[{CE, macro, $[-10,10]$ }\label{fig:xor:b10:ce:macro}]{    \includegraphics[width=0.8\textwidth]{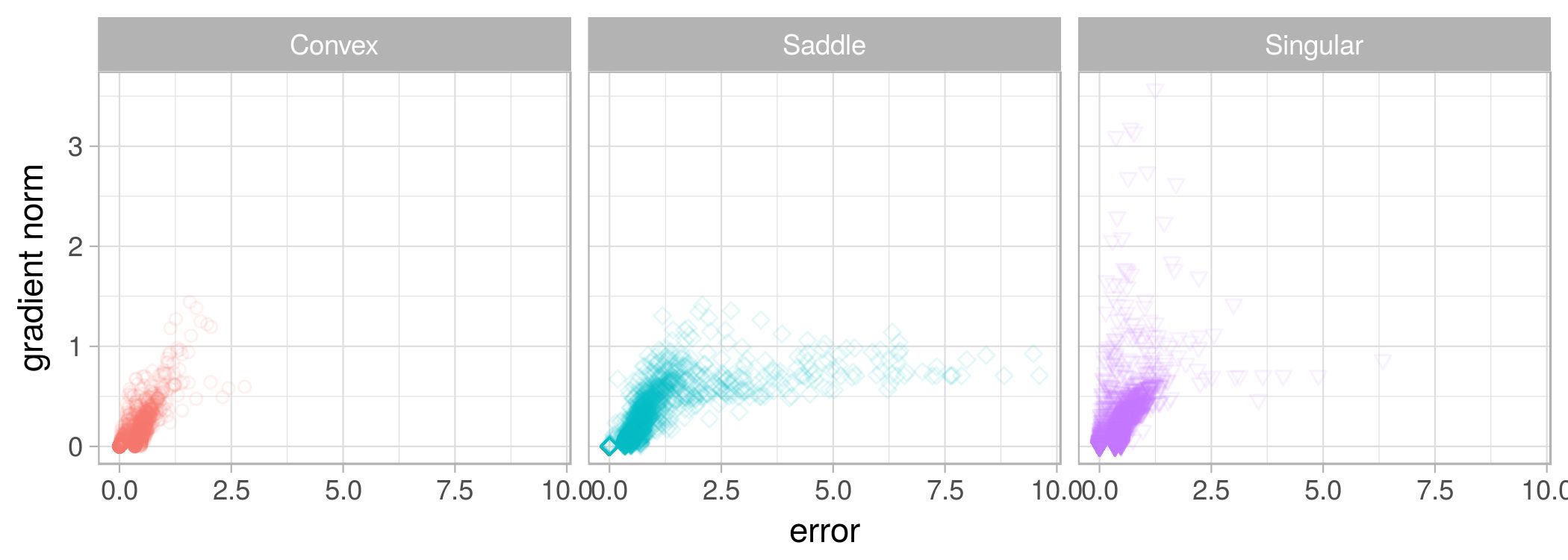}}
  \end{subfloat}
  \caption{L-g clouds for the gradient walks initialised in the $[-10,10]$ range for the XOR problem.}\label{fig:xor:b10}
\end{center}
\end{figure}

Figures~\ref{fig:xor:b1} and~\ref{fig:xor:b10} show the l-g clouds obtained for the XOR problem, separated into panes according to the curvature.

The first observation that can immediately be made from Figure~\ref{fig:xor:b1} is that both SSE and CE yielded exactly four stationary points on the walks initialised in the $[-1,1]$ range. Furthermore, these four points were classified as convex according to the Hessian eigenvalues, indicating that the points can be classified as local minima rather than saddle points. A transition from saddle curvature to convex curvature was observed for both SSE and CE. Points further away from a global optimum were classified as exhibiting saddle curvature. Two stationary points furthest away from a global optimum were sometimes classified as saddles, indicating that both saddles and local minima of equal loss value were discovered. Under the macro setting (larger steps), a few singular points have been sampled in the same apparent basin, indicating that the area was flat (no curvature) in some dimensions.

However, the global minima discovered by the gradient walks initialised in the $[-1,1]$ range appeared perfectly convex. The area surrounding the global minima, as well as the two adjacent local minima, also exhibited convexity. Thus, the XOR problem definitely exhibits convex local minima.

Another interesting observation can be made by observing the trajectories connecting the apparent local minima: It is evident from Figure~\ref{fig:xor:b1} that most high loss, high gradient points first descended to the local minimum furthest away from the global minimum, and from thereon proceeded to one of the three better minima. The three convex minima, however, were not connected by trajectories. In other words, once the gradient walk descended into one of the basins, escape from the basin became unlikely, given the limited step size. To further support this claim, $n_{stag}$ and $l_{stag}$ values calculated for the various XOR gradient walks are reported in Table~\ref{table:xor:lstag}. According to Table~\ref{table:xor:lstag}, the average number of basins visited by the $[-1,1]$ micro-step walks was $1.88889$ for SSE, and $2.04444$ for CE. Thus, the walks visited two or fewer basins. The $n_{stag}$ values are even smaller for macro-step walks initialised in the same range, i.e. $1.33333$ for SSE, and $1.35556$ for CE. Figures~\ref{fig:xor:b1:mse:macro} and~\ref{fig:xor:b1:ce:macro} illustrate that larger step sizes allowed some of the walk trajectories to skip the poor loss area, while the smaller steps consistently became stuck, and proceeded directly to one of the better minima. Small $n_{stag}$ values indicate that transition between adjacent minima was still unlikely for the given step size.

\begin{table}
  \caption{Basin of attraction estimates calculated for the XOR problem. Standard deviation shown in parenthesis.}
  \label{table:xor:lstag}
  \begin{center}
    \begin{tabular}{l|cc|cc}
      & \multicolumn{2}{|c|}{SSE} & \multicolumn{2}{|c}{CE} \\
      \hline
       &  $n_{stag}$    & $l_{stag}$ & $n_{stag}$ & $l_{stag}$ \\
      \hline
      $[-1,1]$, micro &      1.88889 & 367.04444 & 2.04444 & 313.32130 \\
      &        (0.31427) & (134.84453) & (0.44500) & (148.75671) \\\hline
      $[-1,1]$, macro &      1.33333 &  37.14815 & 1.35556 &  30.35000 \\
      & (0.49441) &  (16.68220) & (0.50136) &  (13.32863) \\\hline
      $[-10,10]$, micro&     1.63333 & 684.77778 & 1.16667 & 870.87222 \\
      & (0.72188) & (263.72374) & (0.37268) & (180.17149) \\\hline
      $[-10,10]$, macro&     1.10000 &  57.98889 & 1.03333 &  74.79444 \\
      & (0.39581) &  (24.91864) & (0.23333) &  (20.49253) \\
      \hline
    \end{tabular}
  \end{center}
\end{table}

CE and SSE thus exhibited very similar properties when sampled with $[-1,1]$ gradient walks. The same number of local minima was observed, and the basins of attraction exhibited similar behaviour in terms of basin-to-basin transitions. According to Figure~\ref{fig:xor:b1}, CE exhibited stronger gradients. This corresponds to the theoretical predictions made in~\cite{ref:Solla:1988}. A comparison of Figures~\ref{fig:xor:b1:mse:macro} and~\ref{fig:xor:b1:ce:macro} shows that SSE exhibited more non-convex behaviour around the apparent local minima, which indicates that SSE would be harder to search for an optimisation algorithm than CE.

Figure~\ref{fig:xor:b10} shows the l-g clouds obtained for gradient walks initialised in the $[-10,10]$ range. Figures~\ref{fig:xor:b10:mse:micro} and~\ref{fig:xor:b10:mse:macro} indicate that initialisation in a wider range caused the gradient walks to discover more stationary points on the SSE loss surface: Instead of four points of zero gradient, six can be seen in the figures. Out of these six, only four exhibited convexity. Even the points that exhibited convexity were surrounded by points with saddle curvature or no curvature. Such overlap between convex and non-convex structure indicates that the surface around the minima was not smooth. Overlap of convexity and non-convexity can also indicate that multiple minima of the same loss value exist that exhibit different landscape curvature properties.

\begin{figure}
\begin{center}
  \begin{subfloat}[{CE, micro, $[-10,10]$, error $< 1$ }\label{fig:xor:b10:zoom:a}]{    \includegraphics[width=0.8\textwidth]{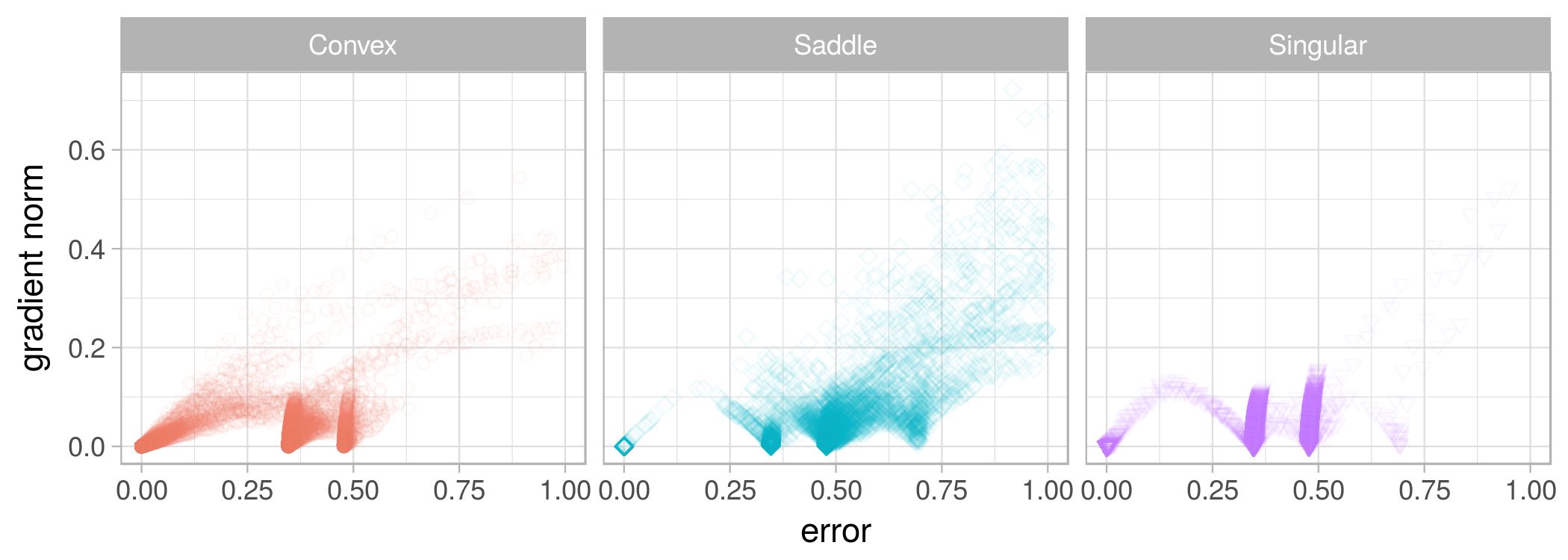}}
  \end{subfloat}
  \begin{subfloat}[{CE, macro, $[-10,10]$, error $< 1$  }]{    \includegraphics[width=0.8\textwidth]{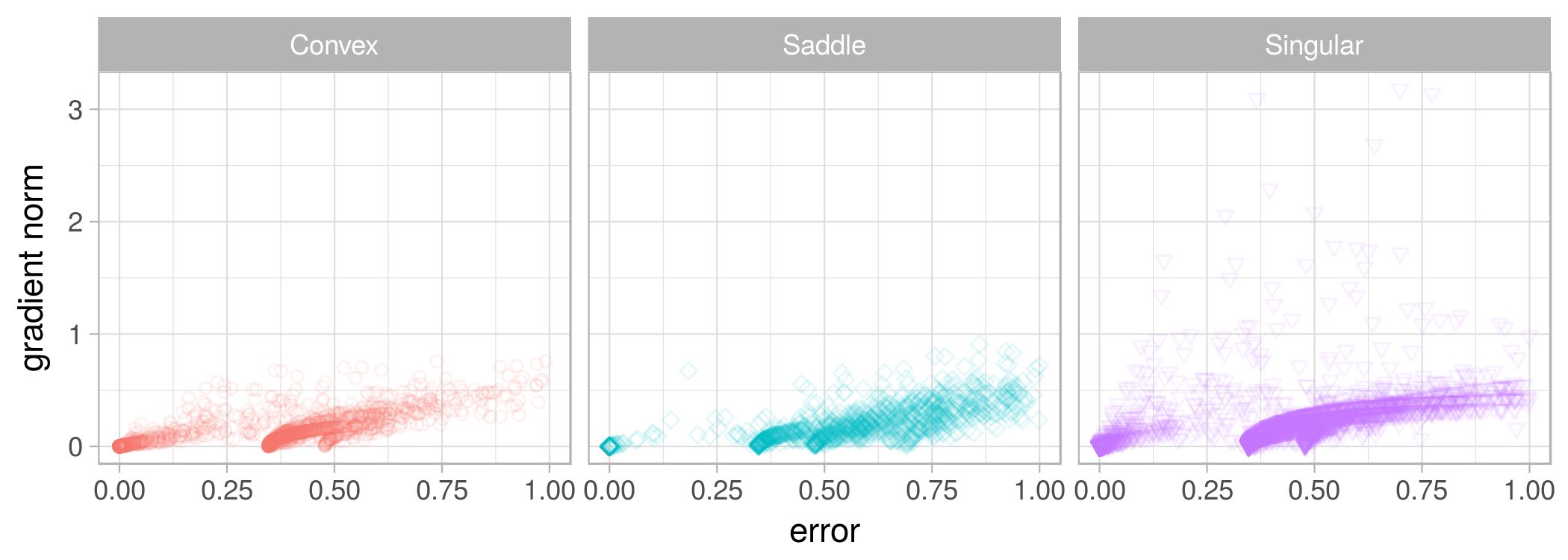}}
  \end{subfloat}
  \caption{L-g clouds for the CE values less than $1$, sampled by the gradient walks initialised in the $[-10,10]$ range for the XOR     problem.}\label{fig:xor:b10:zoom}
\end{center}
\end{figure}

Figures~\ref{fig:xor:b10:ce:micro} and~\ref{fig:xor:b10:ce:macro} show that the loss surface of CE exhibited noticeably different properties when probed in a larger range. While non-convex curvature remained prevalent, CE, as opposed to SSE, did not exhibit additional stationary points. Instead, points of high loss exhibited high fitness, leading the gradient walks towards the same basins as discovered with the $[-1,1]$ walks. Figure~\ref{fig:xor:b10:zoom} displays only those points of the gradient walks that yielded a CE loss value less than 1. Four stationary points can be observed, only three of which exhibited convexity. Thus, CE exhibited fewer local minima than SSE. This observation corresponds with the theoretical predictions made in~\cite{ref:Solla:1988}.

Once again, the convex minima observed in Figures~\ref{fig:xor:b10} and~\ref{fig:xor:b10:zoom} were disconnected from one another. No convex trajectory has been captured that visited all the stationary points present. In fact, according to Figure~\ref{fig:xor:b10:zoom:a}, the only transition between the global optima and the adjacent local optima corresponded to the indefinite Hessians. Thus, to make a transition from one convex minimum to another one, the algorithm had to traverse a flat area with little to no convexity. With reference to Table~\ref{table:xor:lstag}, the $n_{stag}$ values were smaller for the $[-10,10]$ initialisation range, and the $l_{stag}$ values were larger than those yielded by the $[-1,1]$ walks. Thus, the walks were more likely to stagnate once, and to remain in the stagnated state for the entire walk.

A comparison of Figures~\ref{fig:xor:b10:mse:micro} and~\ref{fig:xor:b10:ce:micro} shows that CE demonstrated a smoother, more consistent relationship between the gradient and the loss values than SSE. Together with evidently fewer stationary points, this property makes CE an easier loss surface to search. 

Figures~\ref{fig:xor:b10:mse:macro} and~\ref{fig:xor:b10:ce:macro} indicate that gradient walks with a macro step size, initialised in a larger area, still managed to find the global optima for both SSE and CE, but on fewer occasions than the micro walks. A large portion of the points yielded indefinite Hessians, indicating flatness. This is to be expected, as the loss surfaces of NNs with sigmoidal activation functions are known to exhibit increasing hidden neuron saturation with an increased distance from the origin~\cite{ref:vanAardt:2017}.

\subsection{Iris}
The Iris classification problem is one the most trivial and most commonly used real-world classification datasets. The benefit of studying the Iris problem compared to the XOR problem is that the Iris dataset is large enough to be split into the training and testing subsets. The training subset can then be used to sample the loss surface, and the testing set can be used to evaluate the discovered
minima and stationary points for their ability to generalise. For the rest of the paper, the training set loss values are referred to as $E_t$, and the test set loss values are referred to as $E_g$.

\begin{figure}
\begin{center}
  \begin{subfloat}[{SSE, micro, $[-1,1]$ }]{    \includegraphics[width=0.8\textwidth]{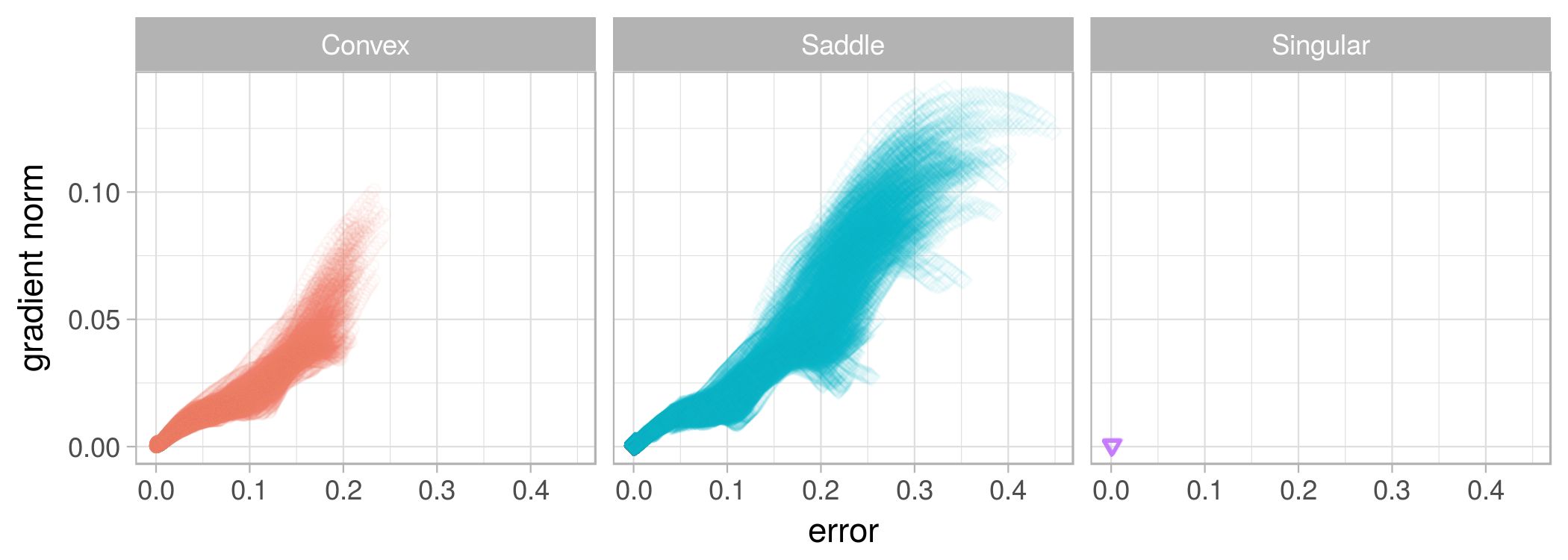}}
  \end{subfloat}
  \begin{subfloat}[{CE, micro, $[-1,1]$ }]{    \includegraphics[width=0.8\textwidth]{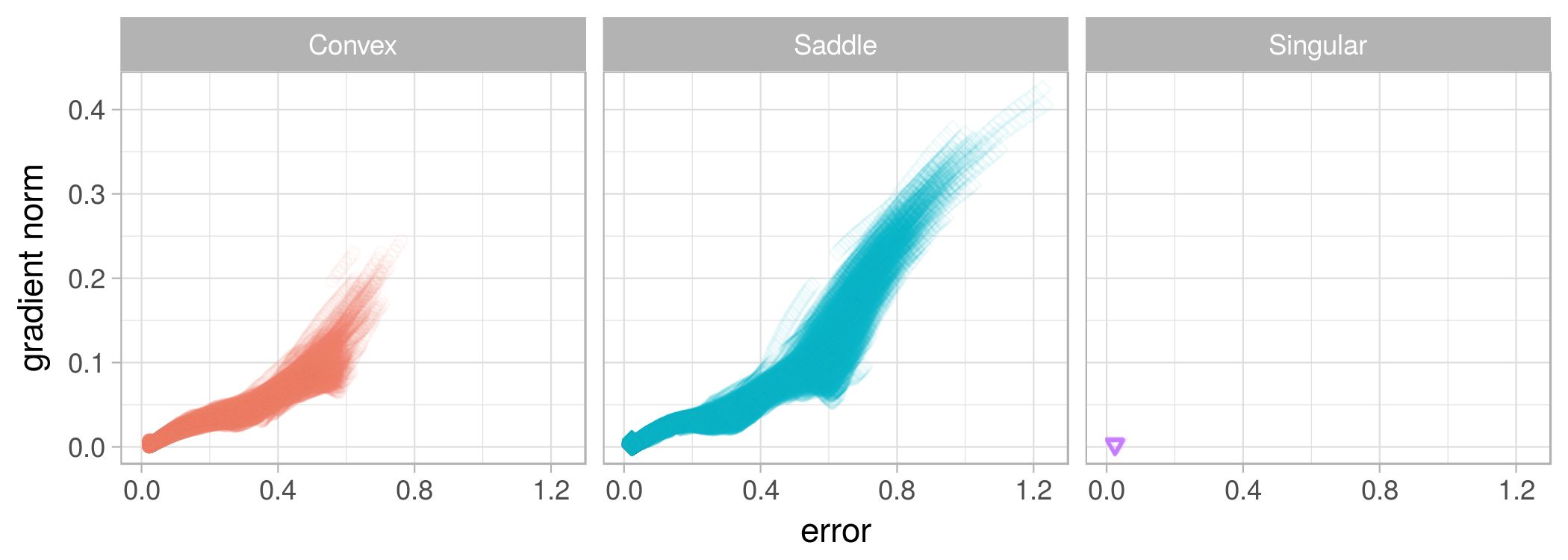}}
  \end{subfloat}
  \begin{subfloat}[{SSE, macro, $[-1,1]$ }]{    \includegraphics[width=0.8\textwidth]{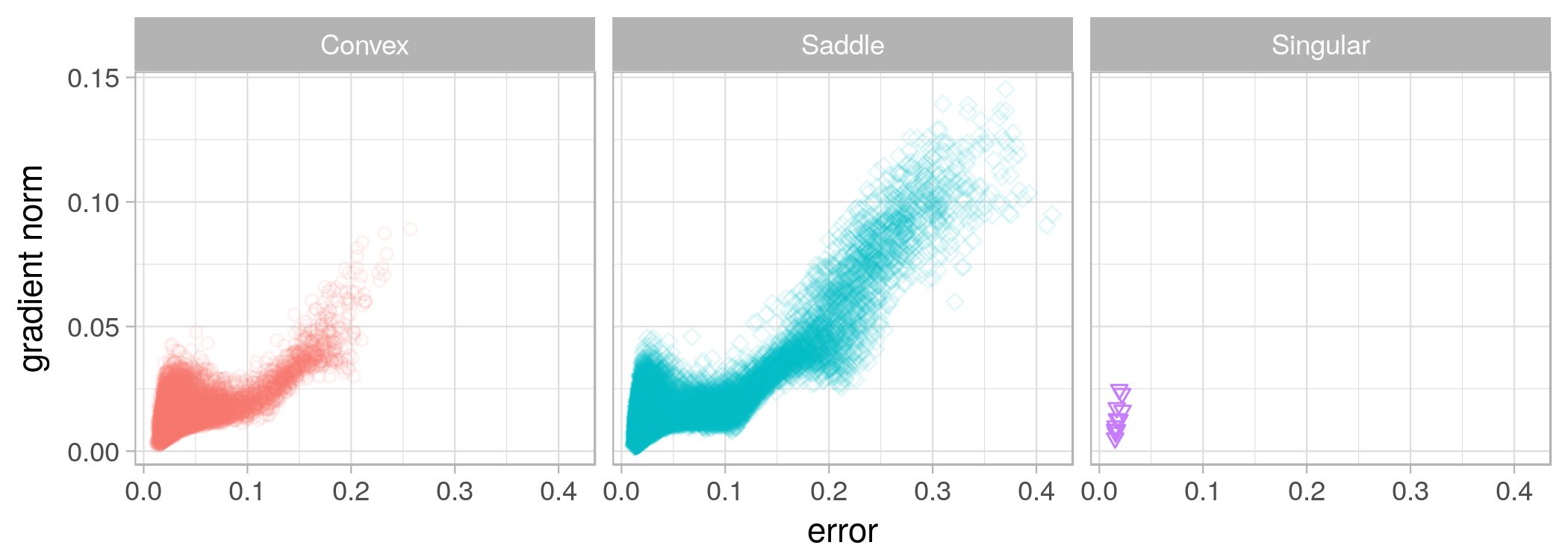}}
  \end{subfloat}
  \begin{subfloat}[{CE, macro, $[-1,1]$ }]{    \includegraphics[width=0.8\textwidth]{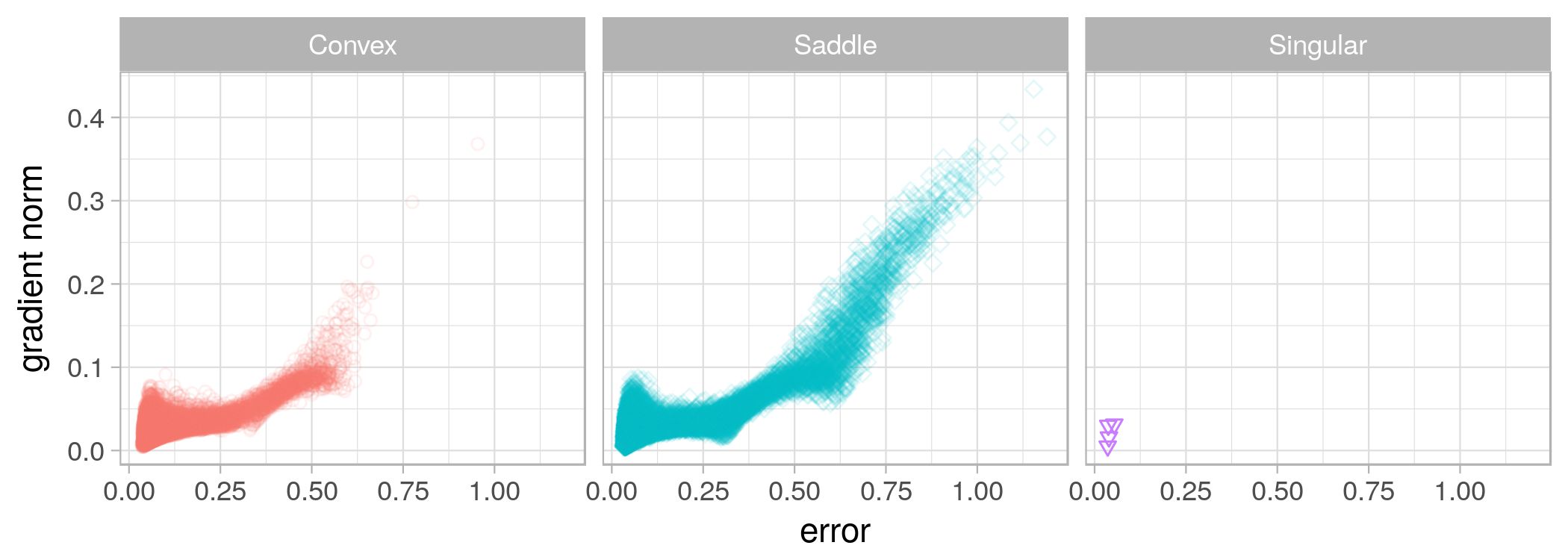}}
  \end{subfloat}
  \caption{L-g clouds for the gradient walks initialised in the $[-1,1]$ range for the Iris problem.}\label{fig:iris:b1}
\end{center}
\end{figure}

\begin{figure}
\begin{center}
  \begin{subfloat}[{SSE, micro, $[-10,10]$ }\label{fig:iris:b10:micro:mse}]{    \includegraphics[width=0.8\textwidth]{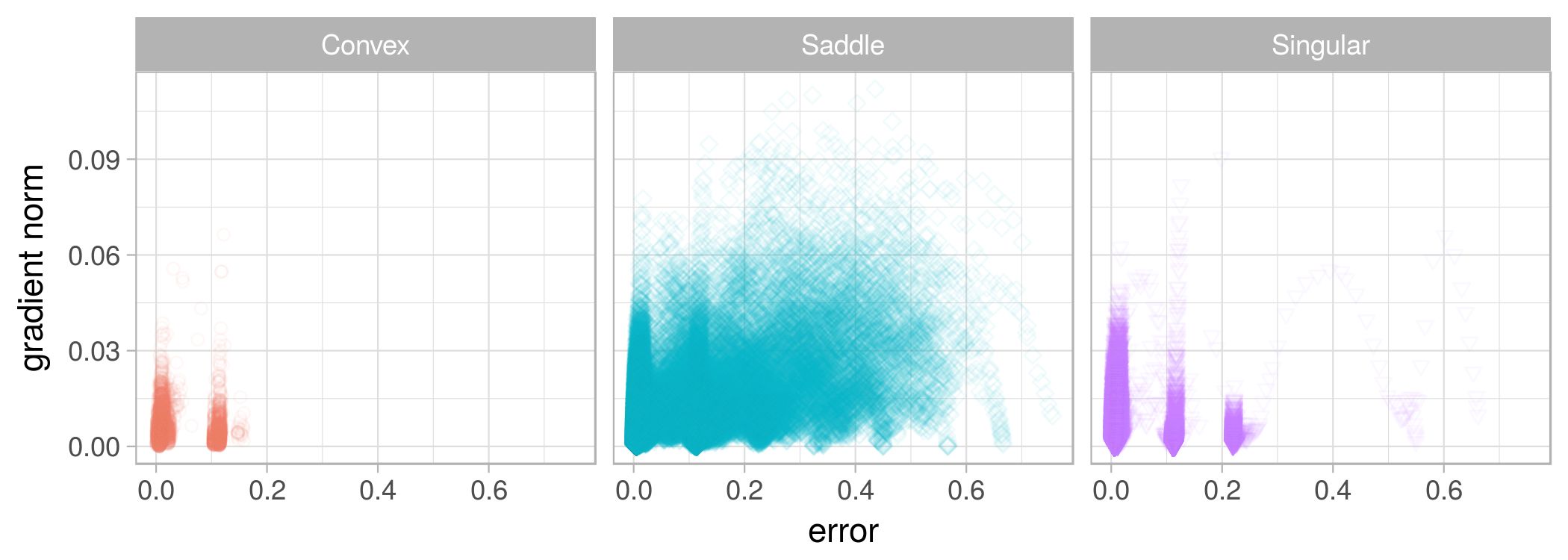}}
  \end{subfloat}
  \begin{subfloat}[{CE, micro, $[-10,10]$ }]{    \includegraphics[width=0.8\textwidth]{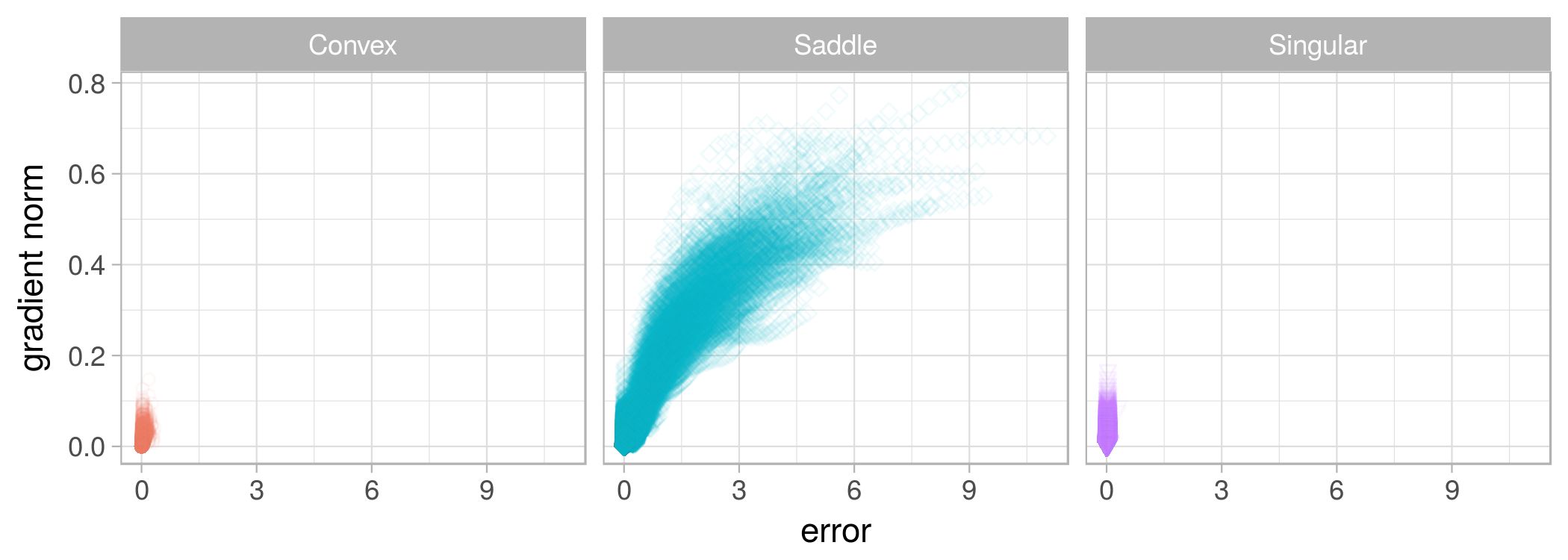}}
  \end{subfloat}
  \begin{subfloat}[{SSE, macro, $[-10,10]$ }\label{fig:iris:b10:macro:mse}]{    \includegraphics[width=0.8\textwidth]{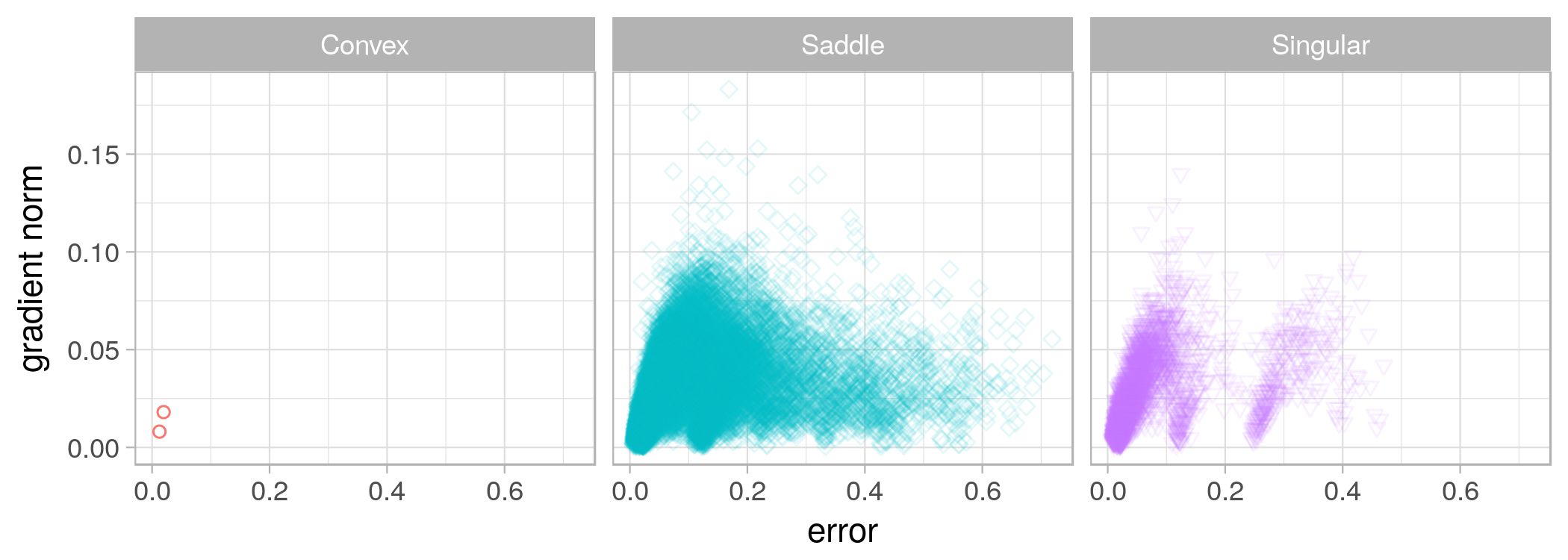}}
  \end{subfloat}
  \begin{subfloat}[{CE, macro, $[-10,10]$ }\label{fig:iris:b10:macro:ce}]{    \includegraphics[width=0.8\textwidth]{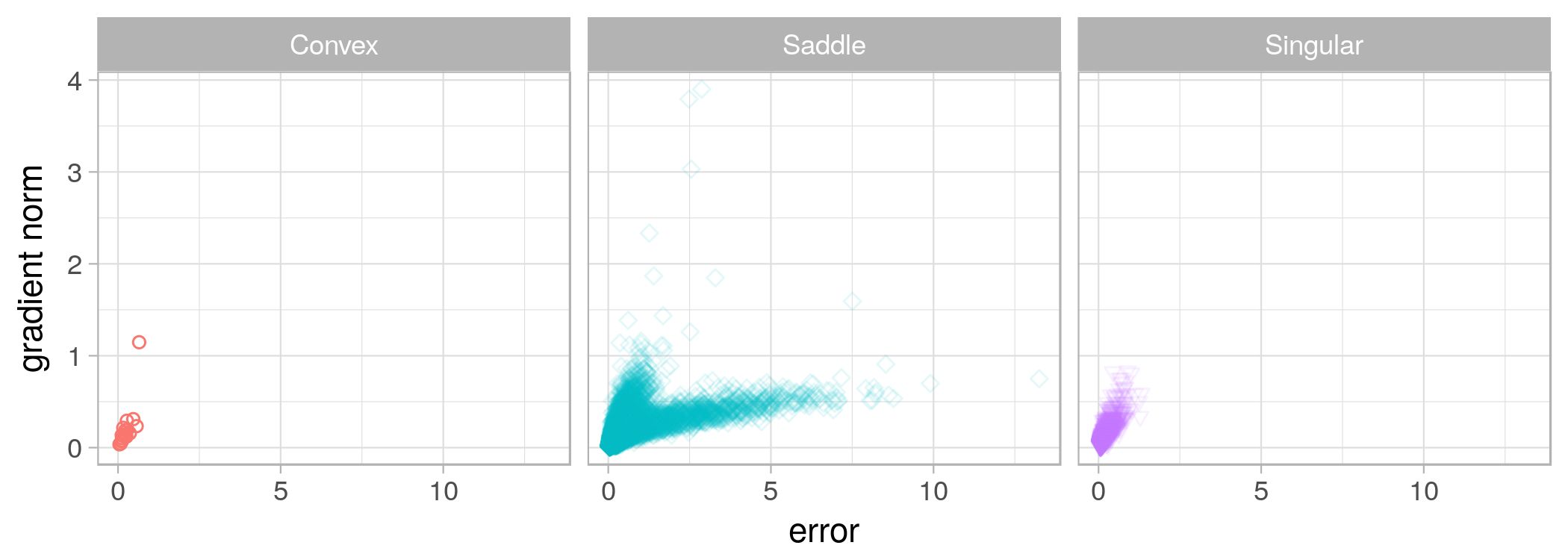}}
  \end{subfloat}
  \caption{L-g clouds for the gradient walks initialised in the $[-10,10]$ range for the Iris problem.}\label{fig:iris:b10}
\end{center}
\end{figure}

Figures~\ref{fig:iris:b1} and~\ref{fig:iris:b10} show the l-g clouds obtained for the Iris problem. According to Figure~\ref{fig:iris:b1}, only one attractor with zero gradient has been discovered on both the SSE and CE loss surfaces by gradient walks initialised in the $[-1,1]$ range. Two more attractors of non-zero gradient can also be observed, however, these attractors do not constitute local minima. Transition from non-convex space to convex space was still present, but was less distinct than for XOR. Points around the global minima exhibited convex as well as saddle behaviour, and saddle behaviour was prevalent. Both the SSE and CE surfaces exhibited flatness (indicated by the singular Hessians) around the global optima. This corresponds to theoretical claims that the loss surface around the global minima is flat \cite{ref:Auer:1996}. However, the flatness was not prevalent. 

A comparison of the micro and macro steps in Figure~\ref{fig:iris:b1} indicates that the macro steps discovered the same landscape characteristics as the micro steps. In the macro setting, a wider range of gradient values around the global minima was discovered. This can be explained by the fact that NN loss surfaces are known to contain ravines and valleys~\cite{ref:Gallagher:2000}, and optima is typically found at the bottom of such structures. The macro step size may have caused the gradient walks to oscillate and to sample points on the sides of the valley where the global minima was discovered.

To further analyse the landscape properties sampled by the gradient walks, the $n_{stag}$ and $l_{stag}$ values have been calculated for the $E_t$ values sampled by the gradient walks, as well as for the corresponding $E_g$ values. Table~\ref{table:iris:t:lstag} lists the $E_t$ and $E_g$ values obtained. According to Table~\ref{table:iris:t:lstag}, both SSE and CE yielded an average
$n_{stag}$ very close or equal to 1 for all gradient walks initialised in the $[-1,1]$ range. Thus, a single basin of attraction was discovered by each individual walk. This correlates well with the results shown in Figure~\ref{fig:iris:b1}. For the macro setting in the $[-1,1]$ range, both SSE and CE produced an $n_{stag}$ average of 1, with a standard deviation of zero. This observation indicates that the macro steps in the $[-1,1]$ range were sufficient to prevent stagnation in suboptimal areas, yet convergence in an attraction basin still took place. The generalisation error exhibited similar behaviour, as shown in Table~\ref{table:iris:t:lstag}. The presence of a single global attractor makes the loss surface associated with the Iris problem
trivial to search using a gradient-based method.

\begin{table}
  \caption{Basin of attraction estimates calculated for the Iris problem on the $E_t$ and $E_g$ walks. Standard deviation shown in parenthesis.}
  \label{table:iris:t:lstag}
  \begin{center}
    \begin{tabular}{l|cc|cc}
      & \multicolumn{2}{|c|}{SSE} & \multicolumn{2}{|c}{CE} \\
      \hline
      {\bf $E_t$ } &  $n_{stag}$    & $l_{stag}$ & $n_{stag}$ & $l_{stag}$ \\
      \hline
      $[-1,1]$,  &  1.00857 & 848.82048 & 1.00571 & 820.20429 \\
      micro      & (0.11922) &  (52.82897) & (0.07538) &  (54.28522) \\\hline
      $[-1,1]$,  &  1.00000 &  76.46571 & 1.00000 &  73.40857 \\
      macro& (0.00000) &   (4.03382) & (0.00000) &   (4.72216) \\\hline
      $[-10,10]$,& 1.28571 & 796.07000 & 1.00000 & 953.26286 \\
       micro& (0.48234) & (212.33922) & (0.00000) &  (10.64382) \\\hline
      $[-10,10]$, & 1.02571 &  73.77000 & 1.00571 &  84.55857 \\
      macro& (0.15828) &  (13.50309) & (0.07538) &  (6.27217) \\\hline\hline
      {\bf $E_g$} &  $n_{stag}$    & $l_{stag}$ & $n_{stag}$ & $l_{stag}$ \\\hline
$[-1,1]$,  &   1.10571 & 820.07167 & 1.02286 & 818.33905 \\
micro& (0.38206) & (143.30209) & (0.18375) &  (77.64190) \\\hline
$[-1,1]$,  &   1.00000 &  74.69429 & 1.00286 &  67.71143 \\
macro& (0.00000) &  (4.52494) & (0.05338) &   (7.26987) \\\hline
$[-10,10]$, &  1.36000 & 770.39048 & 1.12286 & 917.17541 \\
micro& (0.57231) & (229.39260) & (0.75160) & (138.19499) \\\hline
$[-10,10]$, &  1.03143 &  75.41714 & 1.01429 &  83.85857 \\
macro& (0.17447) &  (13.86152) & (0.11867) &   (8.78615) \\\hline
\end{tabular}
  \end{center}
\end{table}

Figure~\ref{fig:iris:b10} shows the l-g clouds obtained for the gradient walks initialised in the $[-10,10]$ interval. According to Figures~\ref{fig:iris:b10:micro:mse} and~\ref{fig:iris:b10:macro:mse}, multiple stationary points were discovered on the SSE loss surface. Two of the discovered stationary points, including the global minima, have exhibited convexity. Thus, there is at least one local minimum on the SSE loss surface associated with the Iris problem. Additionally, the discovered stationary points were disjoint in the convex and singular (flat) space. The saddle space was more connected; however, the $n_{stag}$ values presented in Table~\ref{table:iris:t:lstag} indicate that the gradient walks did not generally become stuck more than twice. Thus, the multiple stationary points discovered were not trivial to escape from. 

\begin{figure}
  \begin{center}
  \begin{subfloat}[{CE, micro, $[-10,10]$, $E_t < 1$ }\label{fig:iris:b10:zoom:mic}]{    \includegraphics[width=0.8\textwidth]{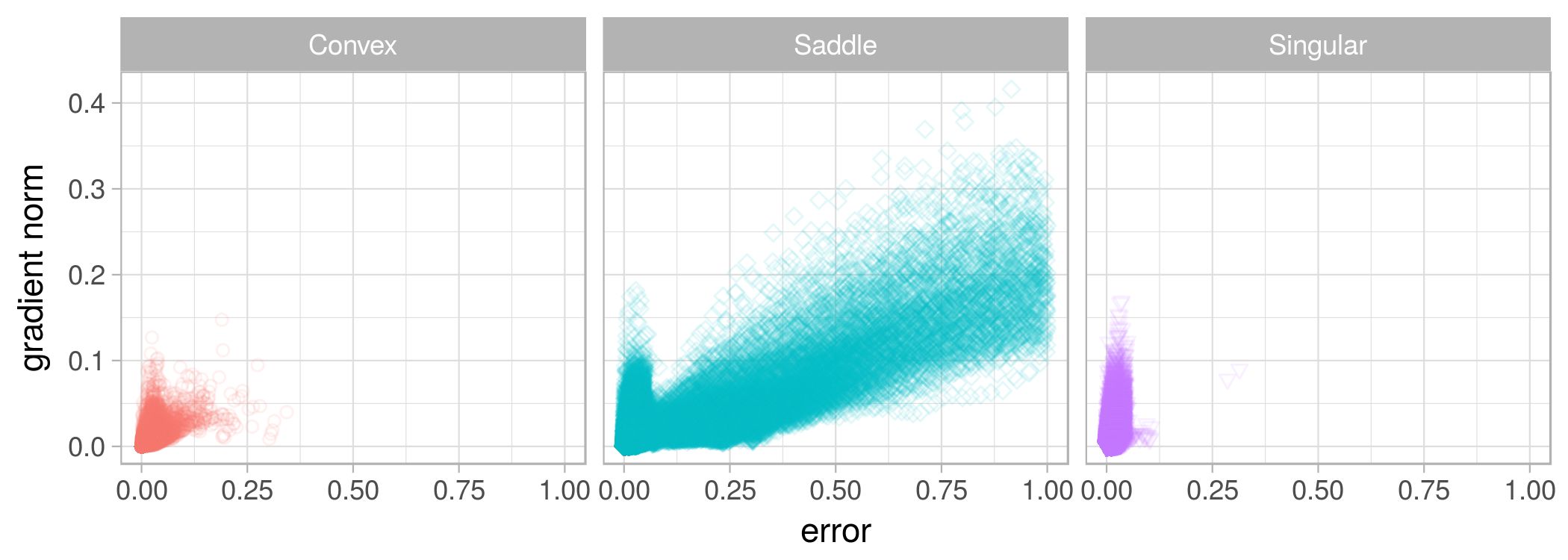}}
  \end{subfloat}
  \begin{subfloat}[{CE, macro, $[-10,10]$, $E_t < 1$  }]{    \includegraphics[width=0.8\textwidth]{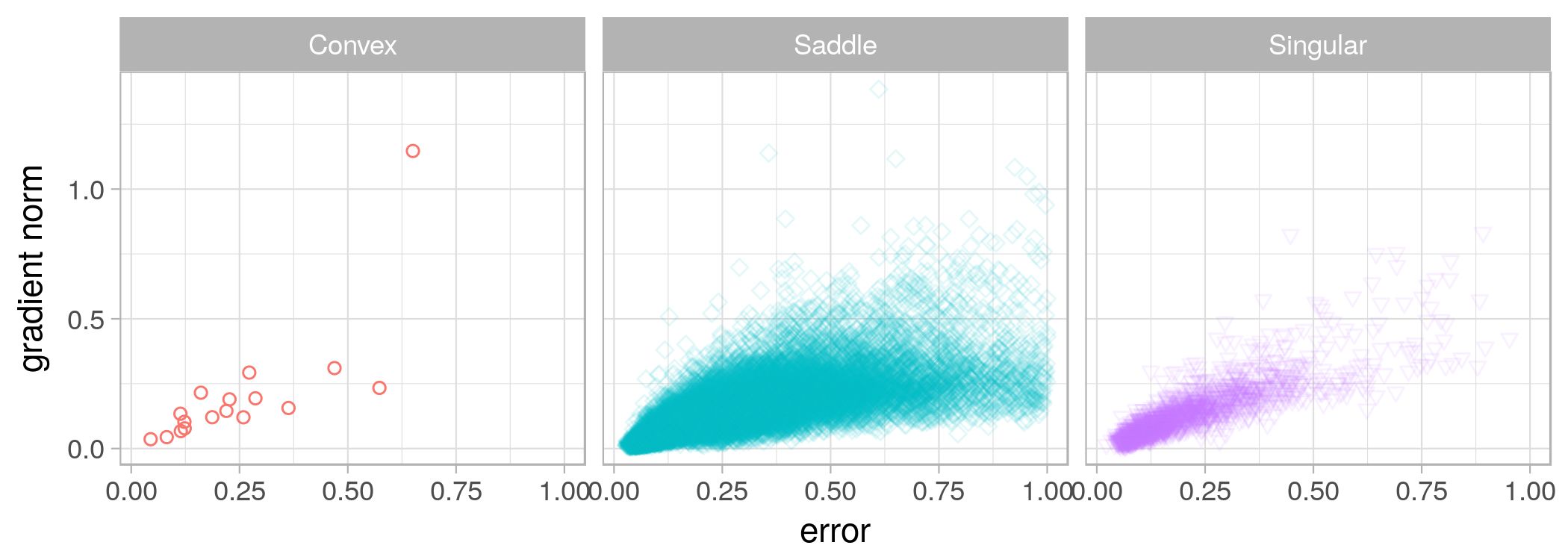}}
  \end{subfloat}
  \caption{L-g clouds for the CE loss values less than $1$ sampled by the gradient walks initialised in the $[-10,10]$ range for the Iris
    problem.}\label{fig:iris:b10:zoom}
  \end{center}
\end{figure}

CE, on the other hand, exhibited only one attractor at the global minima, as illustrated in Figure~\ref{fig:iris:b10}. Figure~\ref{fig:iris:b10:zoom} shows only those points of the gradient walks that yielded a CE loss value less than 1. Even though all points belong to the same global attraction basin, two distinct clusters can be observed in Figure~\ref{fig:iris:b10:zoom:mic}: points that lie in the low error region, and exhibit higher gradients, and points that lie in the higher error region, and exhibit lower gradients. The same tendency can be observed in Figure~\ref{fig:iris:b10:macro:ce}. These observations indicate that the gradient walks have explored wide (higher error, lower gradient) as well as narrow (higher gradient, lower error) valleys, which the NN error landscapes are known to exhibit~\cite{ref:Chaudhari:2017}.

\begin{figure}
  \begin{center}
  \begin{subfloat}[{SSE, micro, $[-1,1]$, $E_t< 0.05$ }\label{fig:iris:gen:micro:mse}]{    \includegraphics[width=0.4\textwidth]{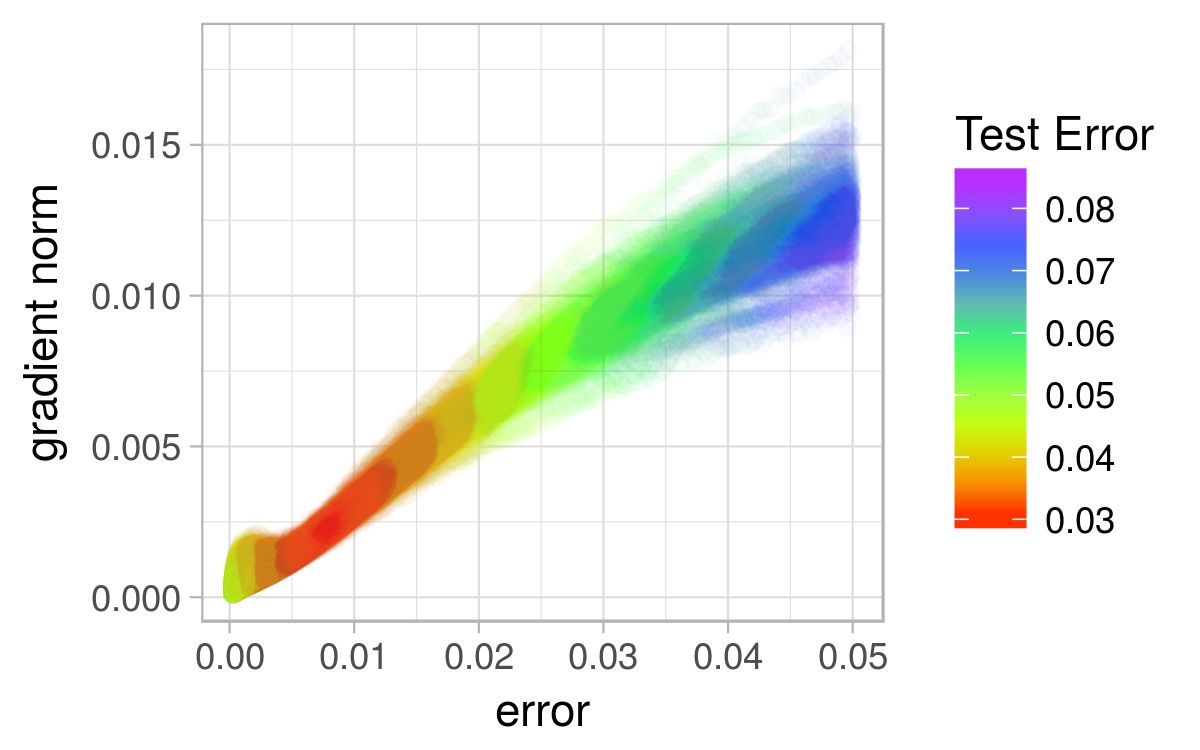}}
  \end{subfloat}
  \begin{subfloat}[{CE, micro, $[-1,1]$, $E_t< 0.05$  }\label{fig:iris:gen:micro:ce}]{    \includegraphics[width=0.4\textwidth]{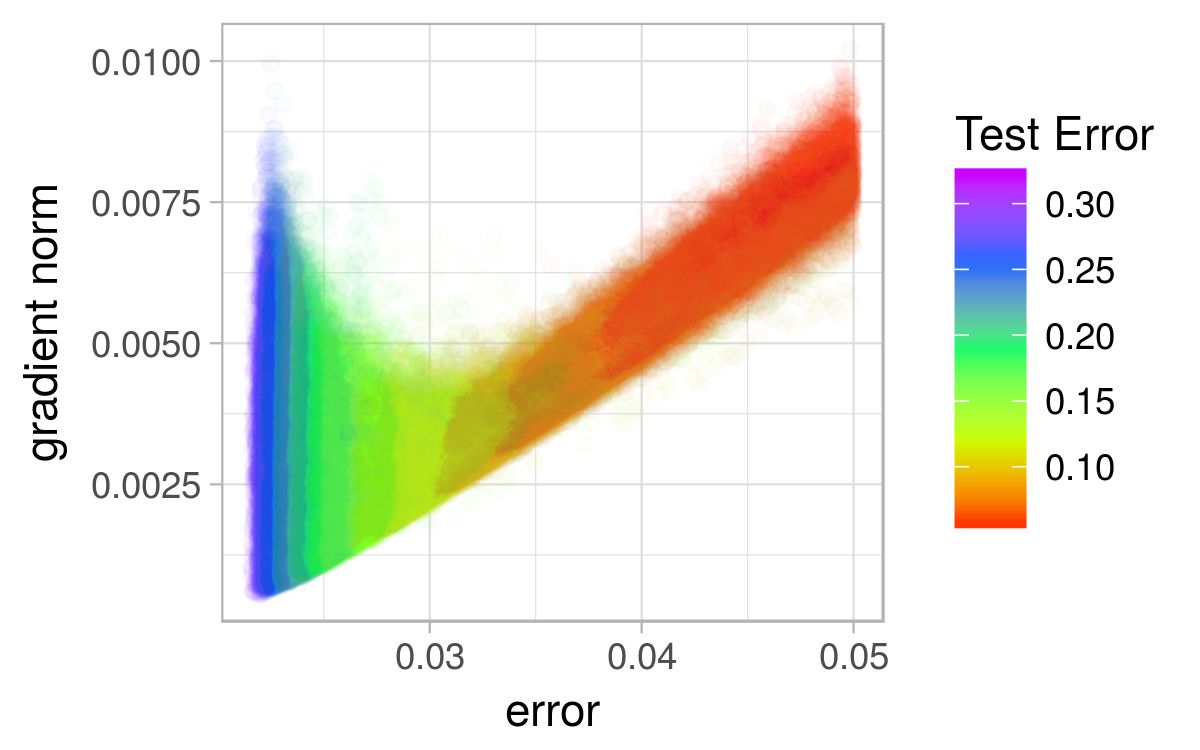}}
  \end{subfloat}
  \begin{subfloat}[{SSE, micro, $[-10,10]$, $E_t< 0.05$ }\label{fig:iris:gen:mse}]{    \includegraphics[width=0.4\textwidth]{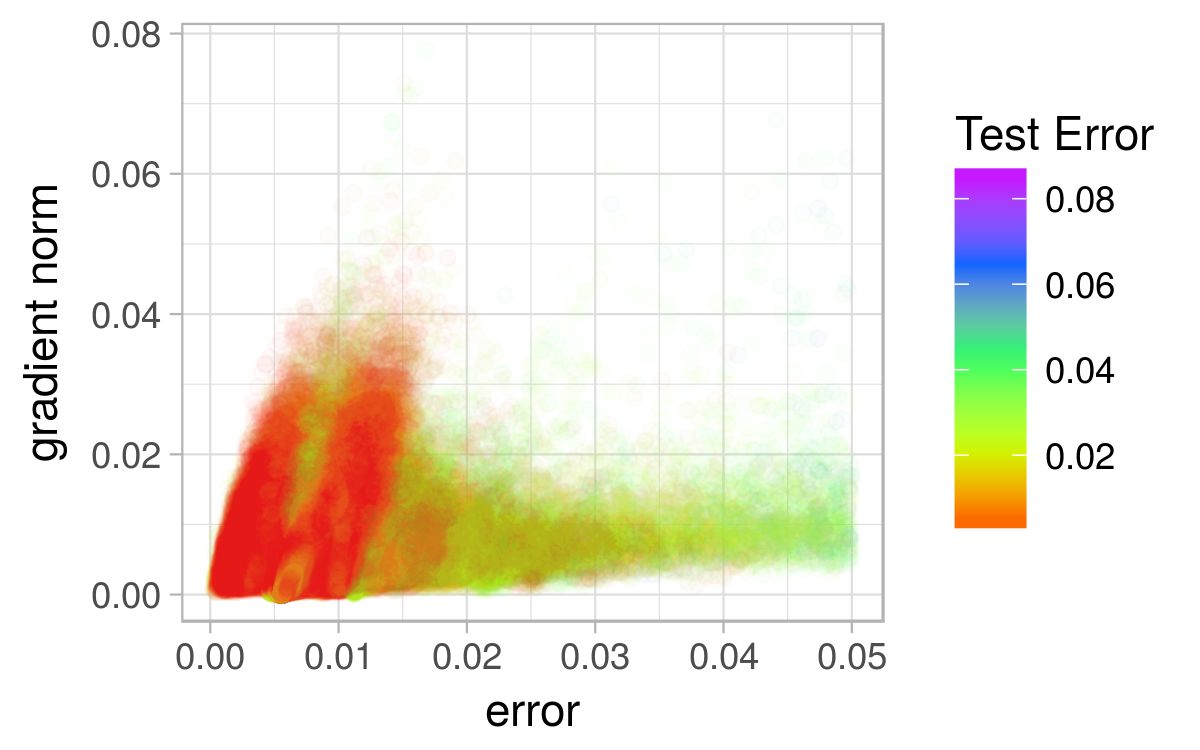}}
  \end{subfloat}
  \begin{subfloat}[{CE, micro, $[-10,10]$, $E_t< 0.05$ }\label{fig:iris:gen:ce}]{    \includegraphics[width=0.4\textwidth]{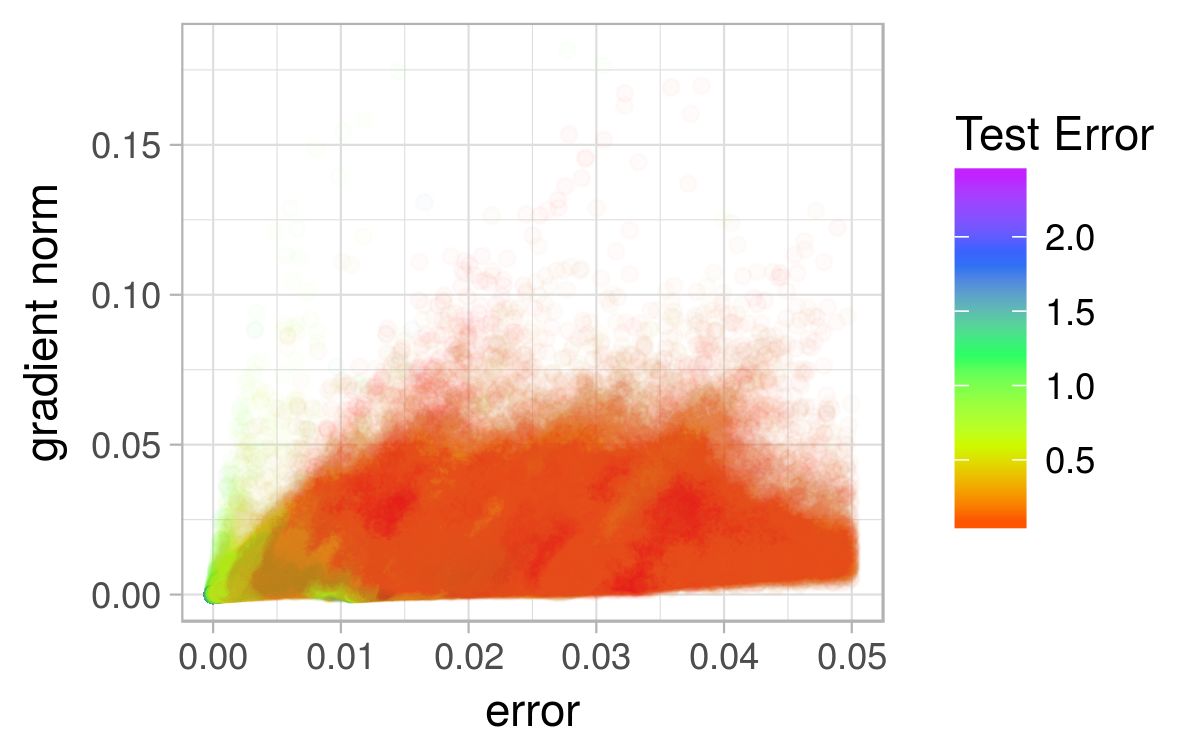}}
  \end{subfloat}
  \caption{L-g clouds colourised according to the corresponding $E_g$ values for the Iris problem.}\label{fig:iris:gen}
  \end{center}
\end{figure}

The CE loss surface once again exhibited fewer local minima than SSE. However, the quality of the minima should also be evaluated in terms of the generalisation capabilities, before any final conclusions can be drawn. Figure~\ref{fig:iris:gen} shows the l-g clouds colourised according to the corresponding $E_g$ values. It is evident from Figure~\ref{fig:iris:gen} that CE yielded poor generalisation performance in the area of the global minima: all $E_g$ values reported were an order of magnitude larger than the corresponding $E_t$ values. This observation is to be expected: achieving 100\% accuracy on the training may easily lead to overfitting. SSE also exhibited overfitting at the global minima, but not as strongly as CE. CE exhibited stronger gradients around the global optima, which can cause gradient-based methods to overfit more easily on CE than on SSE.

\ref{appendix:class} lists all classification errors obtained by the
gradient walks on the various problems. Table~\ref{table:iris:class}
indicates for the Iris problem that SSE has indeed yielded better
generalisation in most scenarios.

Thus, CE exhibited better global structure than SSE on the Iris problem, and was more searchable from the gradient descent perspective. However, stronger gradients around the global optima indicate that CE exhibited sharper minima, causing stronger overfitting on the CE loss surface.

\subsection{Diabetes}
\begin{figure}
\begin{center}
  \begin{subfloat}[{SSE, micro, $[-1,1]$ }\label{fig:diab:mse:micro:b1}]{    \includegraphics[width=0.6\textwidth]{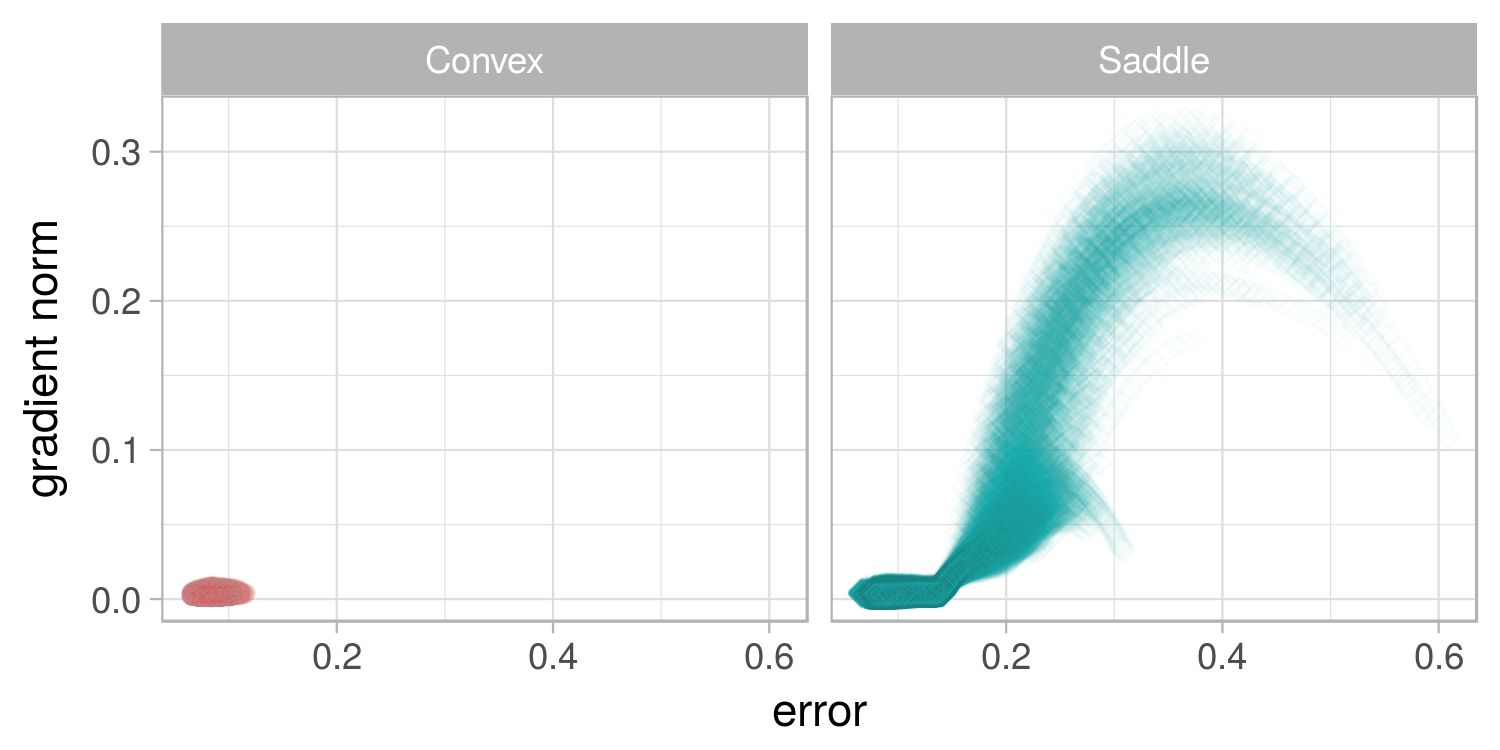}}
  \end{subfloat}
  \begin{subfloat}[{CE, micro, $[-1,1]$ }]{    \includegraphics[width=0.6\textwidth]{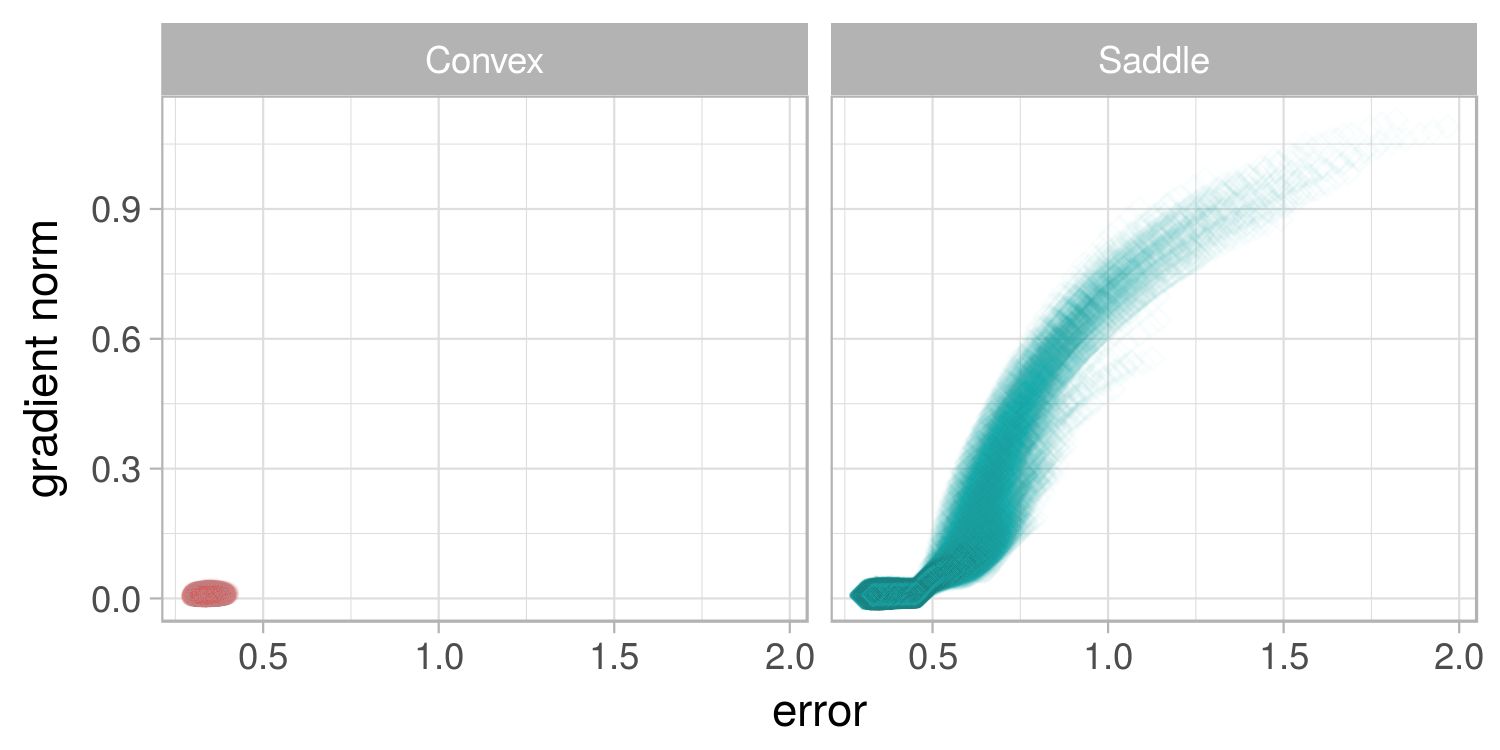}}
  \end{subfloat}
  \begin{subfloat}[{SSE, macro, $[-1,1]$ }\label{fig:diab:mse:macro:b1}]{    \includegraphics[width=0.6\textwidth]{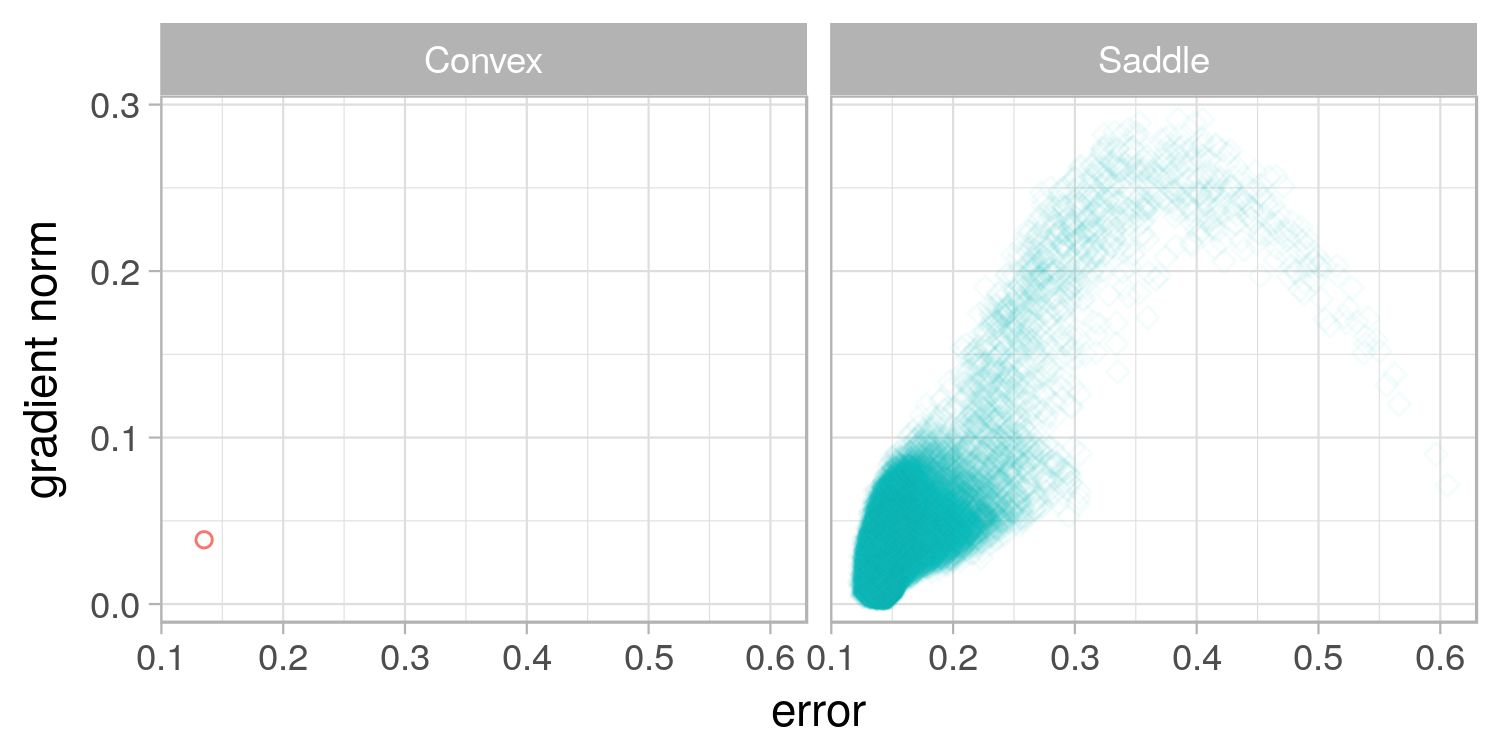}}
  \end{subfloat}
  \begin{subfloat}[{CE, macro, $[-1,1]$ }]{    \includegraphics[width=0.6\textwidth]{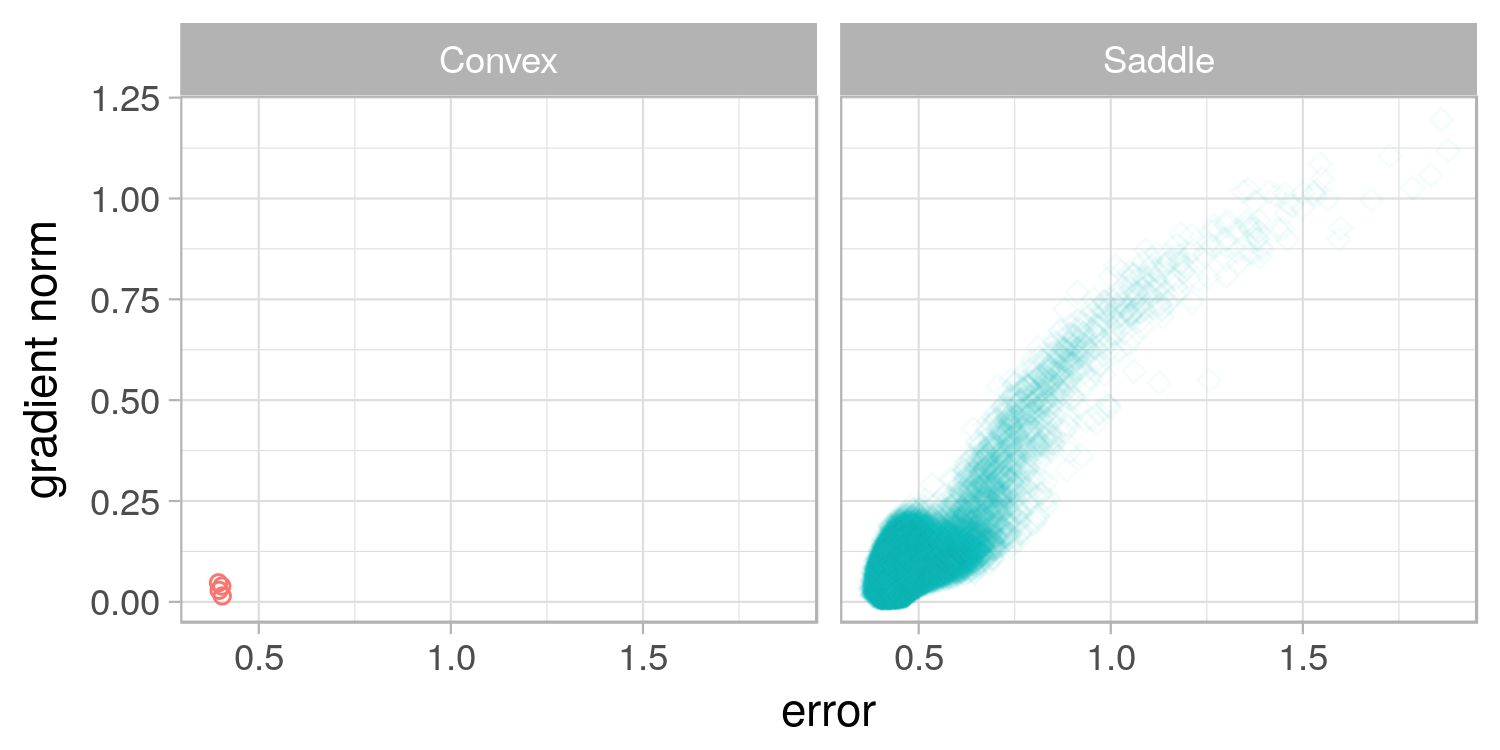}}
  \end{subfloat}
  \caption{L-g clouds for the gradient walks initialised in the $[-1,1]$ range for the Diabetes problem.}\label{fig:diab:b1}
\end{center}
\end{figure}

\begin{figure}
\begin{center}
  \begin{subfloat}[{SSE, micro, $[-10,10]$ }]{    \includegraphics[width=0.6\textwidth]{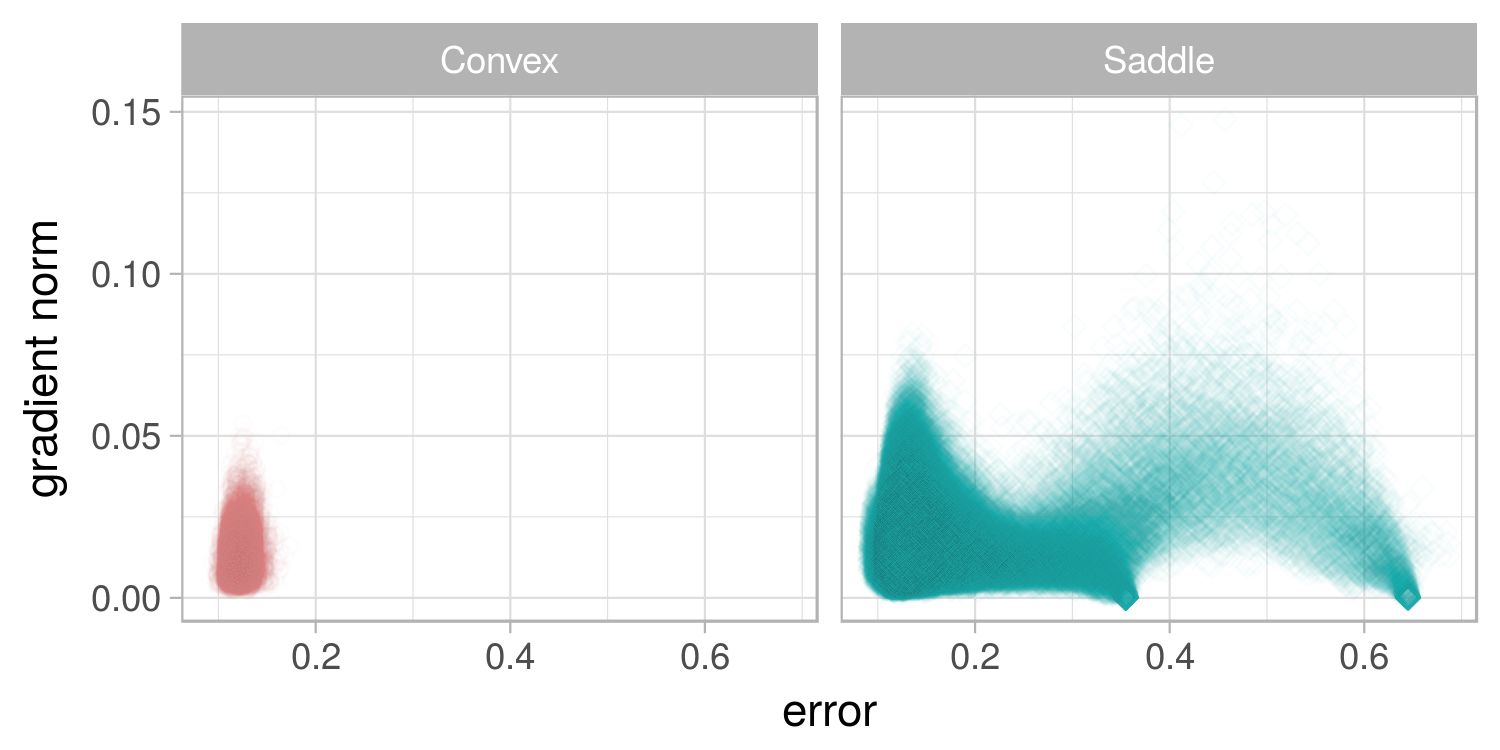}}
  \end{subfloat}
  \begin{subfloat}[{CE, micro, $[-10,10]$ }]{    \includegraphics[width=0.6\textwidth]{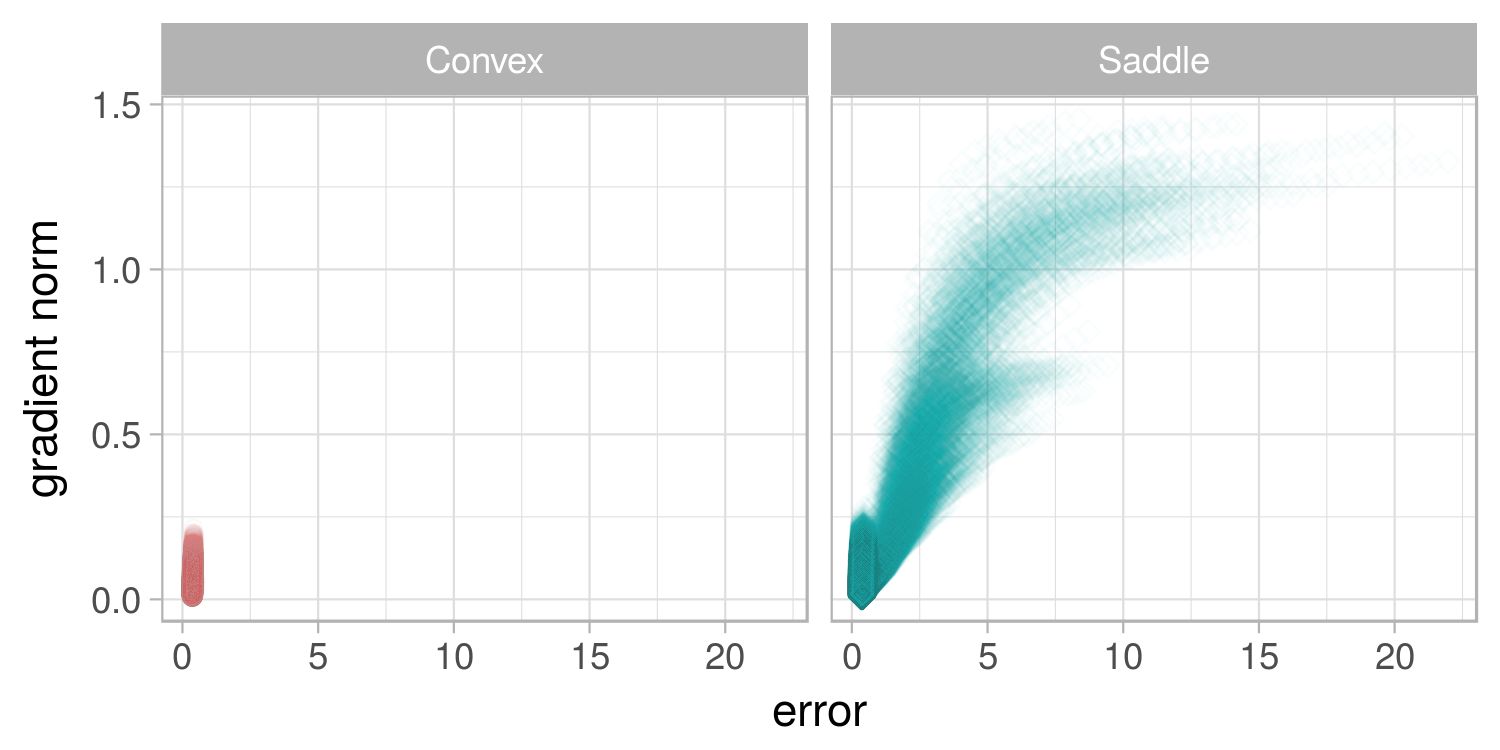}}
  \end{subfloat}
  \begin{subfloat}[{SSE, macro, $[-10,10]$ }]{    \includegraphics[width=0.6\textwidth]{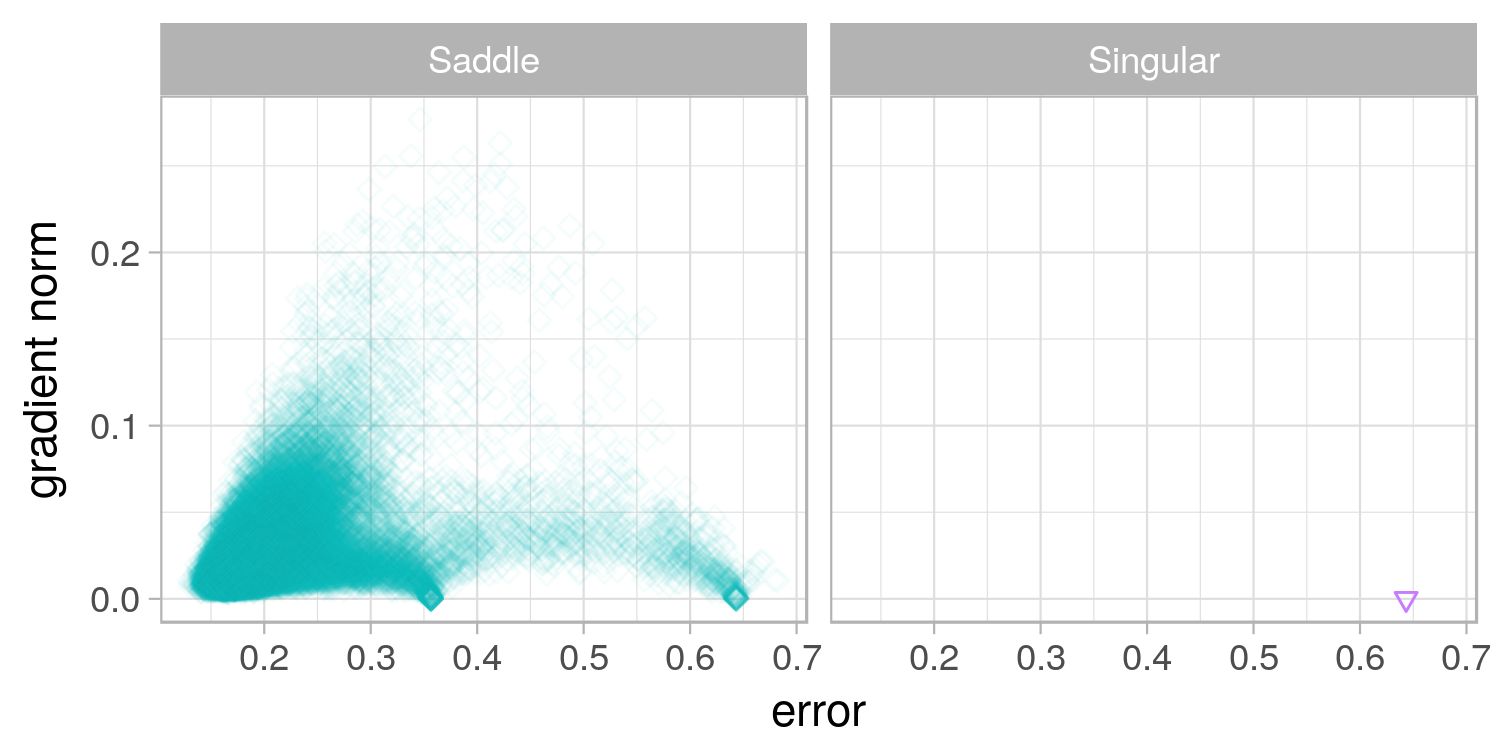}}
  \end{subfloat}
  \begin{subfloat}[{CE, macro, $[-10,10]$ }]{    \includegraphics[width=0.6\textwidth]{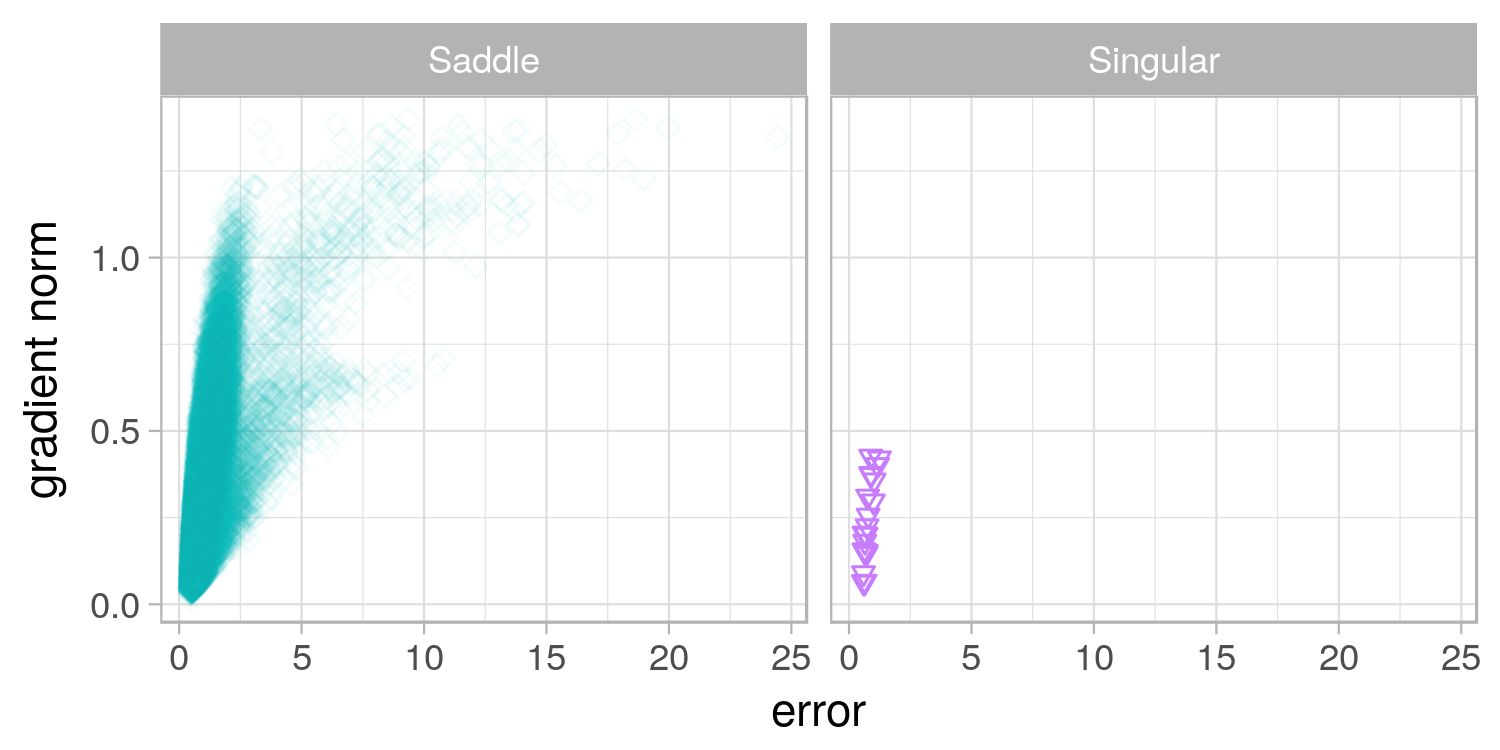}}
  \end{subfloat}
  \caption{L-g clouds for the gradient walks initialised in the $[-1,1]$ range for the Diabetes problem.}\label{fig:diab:b10}
\end{center}
\end{figure}

Figures~\ref{fig:diab:b1} and~\ref{fig:diab:b10} show the l-g clouds obtained for the Diabetes problem. According to Figure~\ref{fig:diab:b1}, both SSE and CE exhibited a single attractor of near-zero gradient, and that attractor constituted a wide area of low gradients around the loss of zero. Both SSE and CE exhibited convexity around zero loss, especially when sampled with micro steps. The majority of the sampled points, however, were once again classified as a saddle according to their Hessians. This corresponds well with the observations made by Dauphin et al. \cite{ref:Dauphin:2014}, where the prevalence of saddle points in non-convex optimisation was studied.

An arch-like curve can be observed in Figures~\ref{fig:diab:mse:micro:b1} and~\ref{fig:diab:mse:macro:b1}, indicating that higher errors were associated with weaker gradients on the SSE loss surface. A transition to the area of higher fitness was associated with a gradient signal that became stronger for some time, and then began to weaken again as a global optimum was approached. The CE l-g clouds in Figure~\ref{fig:diab:b1} indicate that the CE loss surface did not have the tendency to exhibit weaker gradients for higher errors, which makes CE favourable from the gradient descent perspective. This corresponds well with the theoretical properties of both loss functions, which indicate that SSE is expected to exhibit weaker gradients for higher errors, as opposed to CE~\cite{ref:Golik:2013}.

Figure~\ref{fig:diab:b10} shows the l-g clouds obtained for the points sampled by the gradient walks initialised in the $[-10,10]$ interval. SSE loss once again exhibited multiple near-zero gradient attractors (three), and CE loss exhibited only one attractor. The majority of the points sampled by larger steps in a larger area had a saddle curvature. The convex attractors sampled by the $[-10,10]$ walks exhibited more variation in gradient than the corresponding attractors discovered by the $[-1,1]$ walks. This observation can be attributed to the valley structure of the optima: Larger steps induced oscillations around the walls of the valley. Macro $[-10,10]$ walks did not discover any convexity, and yielded a few points of singular (flat) curvature. This behaviour is likely due to the hidden neuron saturation: unconstrained macro gradient walks initialised in a larger range are more likely to explore search space areas that are further away from the origin. Larger weights increase the magnitude of net input signals that the hidden neuron receive, and exceedingly large net input signals cause hidden unit saturation.

The $n_{stag}$ and $l_{stag}$ values reported in Table~\ref{table:diab:t:lstag} indicate that most walks have discovered a single attractor only, which correlates well with Figures~\ref{fig:diab:b1} and~\ref{fig:diab:b10}, and also indicates that the two suboptimal attractors discovered on the SSE loss surface were not easy to escape from. Table~\ref{table:diab:t:lstag} also shows that the generalisation performance of the points discovered on the SSE loss surface was somewhat volatile when sampled using micro walks. Micro walks took smaller steps, and thus were more likely to exploit a particular attractor, causing overfitting.

\begin{table}
  \caption{Basin of attraction estimates calculated for the Diabetes problem on the $E_t$ and $E_g$ walks. Standard deviation shown in parenthesis.}
  \label{table:diab:t:lstag}
  \begin{center}
    \begin{tabular}{l|cc|cc}
      & \multicolumn{2}{|c|}{SSE} & \multicolumn{2}{|c}{CE} \\
      \hline
      {\bf $E_t$}  &  $n_{stag}$    & $l_{stag}$ & $n_{stag}$ & $l_{stag}$ \\
      \hline
      $[-1,1]$,  &  1.00123 & 938.66728 & 1.00000 & 935.27160 \\
      micro      & (0.03511) &  (22.51936) & (0.00000) &  (12.49791) \\\hline
      $[-1,1]$,  &  1.00000 &  85.19012 & 1.00000 &  84.84691 \\
      macro& (0.00000) &   (1.54389) & (0.00000) &   (1.97922) \\\hline
      $[-10,10]$,&  1.09259 & 905.05504 & 1.00000 & 962.31235 \\
       micro& (0.37525) & (138.96827) & (0.00000) &   (5.28613) \\\hline
      $[-10,10]$, & 1.03580 &  77.06975 & 1.02716 &  78.23086 \\
      macro& (0.18580) &  (13.75685) & (0.18393) &  (16.45928) \\\hline\hline
      {\bf $E_g$} &  $n_{stag}$    & $l_{stag}$ & $n_{stag}$ & $l_{stag}$ \\\hline
      $[-1,1]$,  &   1.51852 & 794.78363 & 1.04938 & 925.83735 \\
      micro      & (1.21727) & (270.89365) & (0.31822) & (100.01013) \\\hline
      $[-1,1]$,  &  1.00494 &  85.80988 & 1.00123 &  85.16543 \\
      macro& (0.07010) &   (4.91134) & (0.03511) &   (2.61851) \\\hline
      $[-10,10]$,& 2.76420 & 703.52152 & 1.00617 & 958.96852 \\
       micro& (3.96060) & (343.41982) & (0.07832) &  (39.60395) \\\hline
      $[-10,10]$, & 1.08148 &  53.88477 & 1.08272 &  70.88848 \\
      macro& (0.52189) &  (33.59152) & (0.35041) &  (24.84558) \\\hline
\end{tabular}
  \end{center}
\end{table}

\begin{figure}
\begin{center}
  \begin{subfloat}[{SSE, micro, $[-1,1]$, $E_t< 0.2$ }\label{fig:diab:gen:micro:mse}]{    \includegraphics[width=0.4\textwidth]{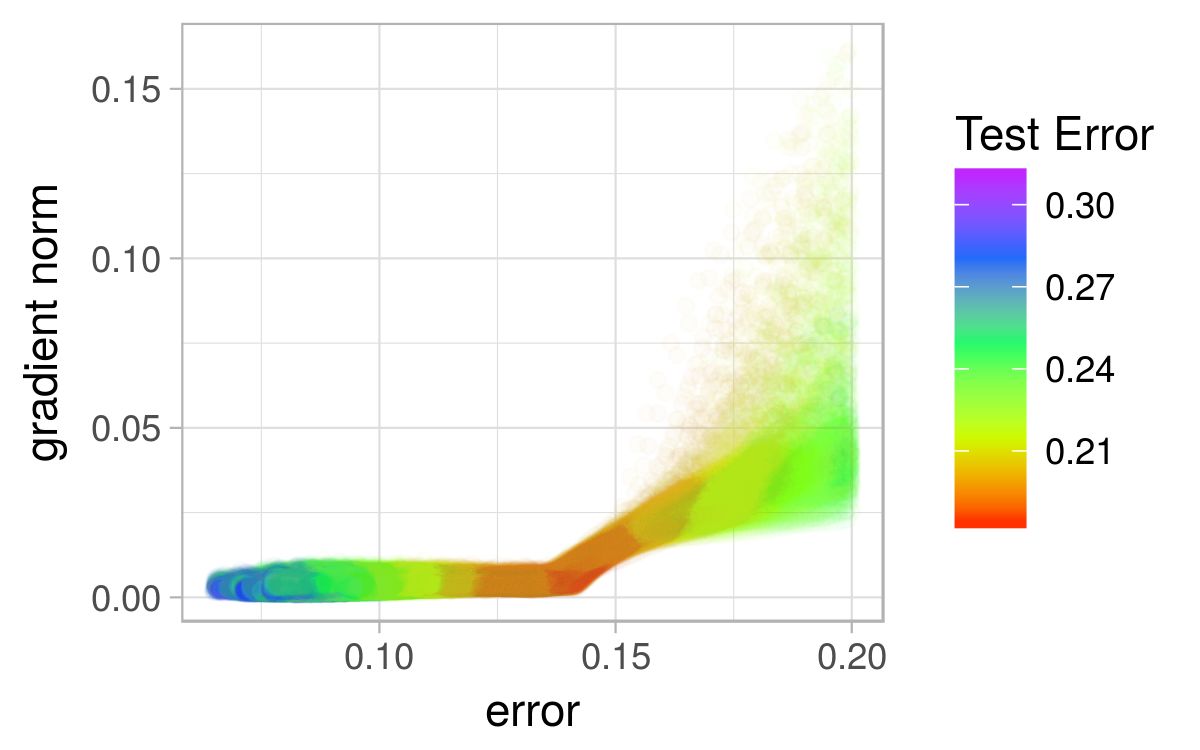}}
  \end{subfloat}
  \begin{subfloat}[{CE, micro, $[-1,1]$, $E_t< 0.5$  }\label{fig:diab:gen:micro:ce}]{    \includegraphics[width=0.4\textwidth]{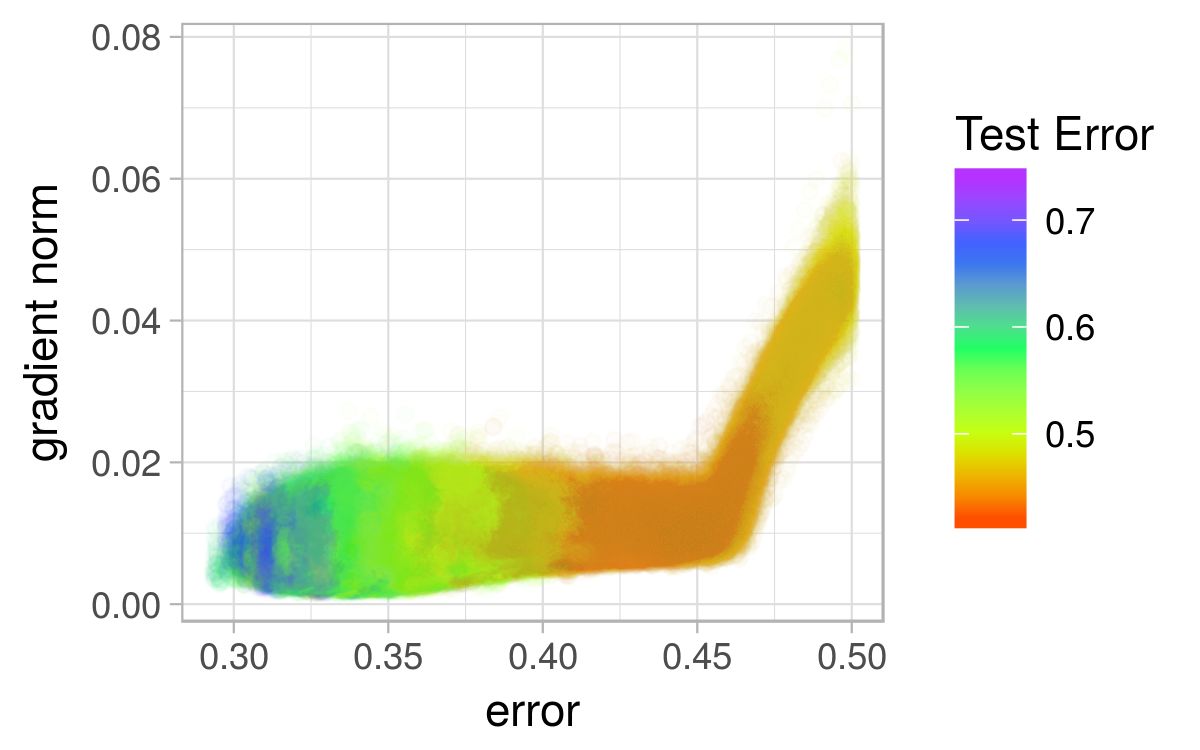}}
  \end{subfloat}
  \begin{subfloat}[{SSE, micro, $[-10,10]$, $E_t< 0.2$ }\label{fig:diab:gen:macro:mse}]{    \includegraphics[width=0.4\textwidth]{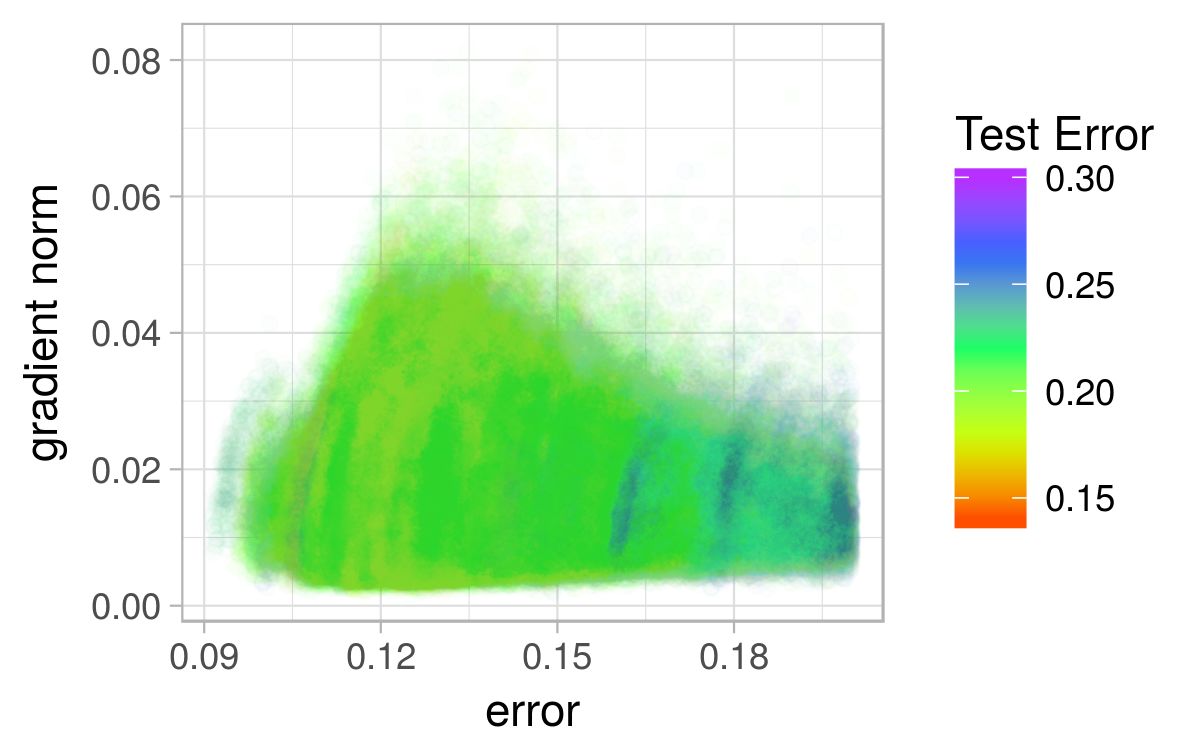}}
  \end{subfloat}
  \begin{subfloat}[{CE, micro, $[-10,10]$, $E_t< 1$  }\label{fig:diab:gen:macro:ce}]{    \includegraphics[width=0.4\textwidth]{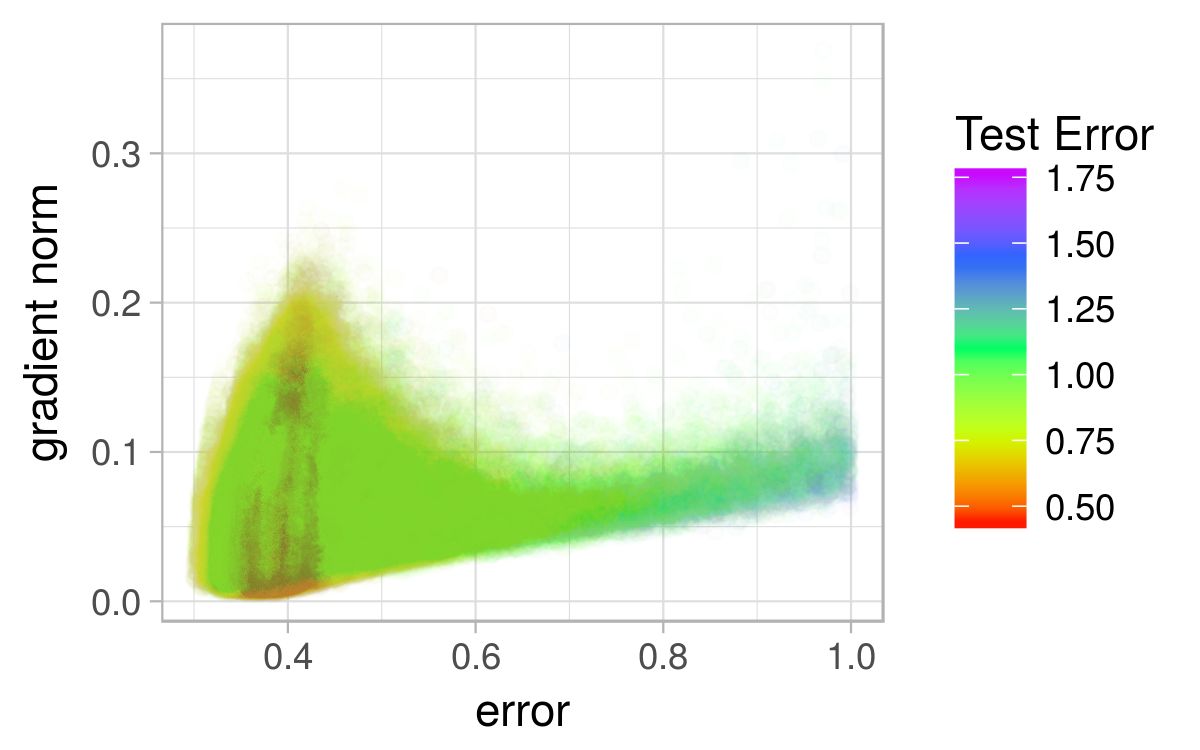}}
  \end{subfloat}
  \caption{L-g clouds colourised according to the corresponding $E_g$ values for the Diabetes problem.}\label{fig:diab:gen}
\end{center}
\end{figure}

Figure~\ref{fig:diab:gen} shows a close-up depiction of the convex attractors, colourised according to their generalisation performance. Both SSE and CE exhibited deteriorating generalisation performance as the walks sampled points closer to the zero loss, which is to be expected. For micro $[-1,1]$ walks, both SSE and CE exhibited a sudden drop in gradient magnitudes, and the points of low gradient with the highest error exhibited the best generalisation performance. As previously noted by Choromanska et al.~\cite{ref:Choromanska:2015}, finding the global minimum may be unnecessary, as the global minimum is likely to overfit the problem. Figures~\ref{fig:diab:gen:macro:mse} and~\ref{fig:diab:gen:macro:ce} indicate for the $[-10,10]$ walks that points around the global minima have exhibited various degrees of generalisation performance, with a significant overlap between good and poor generalisation. This indicates that the discovered minima had the same training error values, but different test error values. The Diabetes problem is known to contain noisy data, and noise is a common cause of overfitting. Table~\ref{table:diab:class} lists the classification errors obtained for the Diabetes problem, and shows that CE loss yielded better generalisation when sampled with the $[-1,1]$ walks, and SSE generalised better when larger step sizes were used.

\subsection{Glass}
\begin{figure}
\begin{center}
  \begin{subfloat}[{SSE, micro, $[-1,1]$ }\label{fig:glass:mse:micro:b1}]{    \includegraphics[height=0.19\textheight]{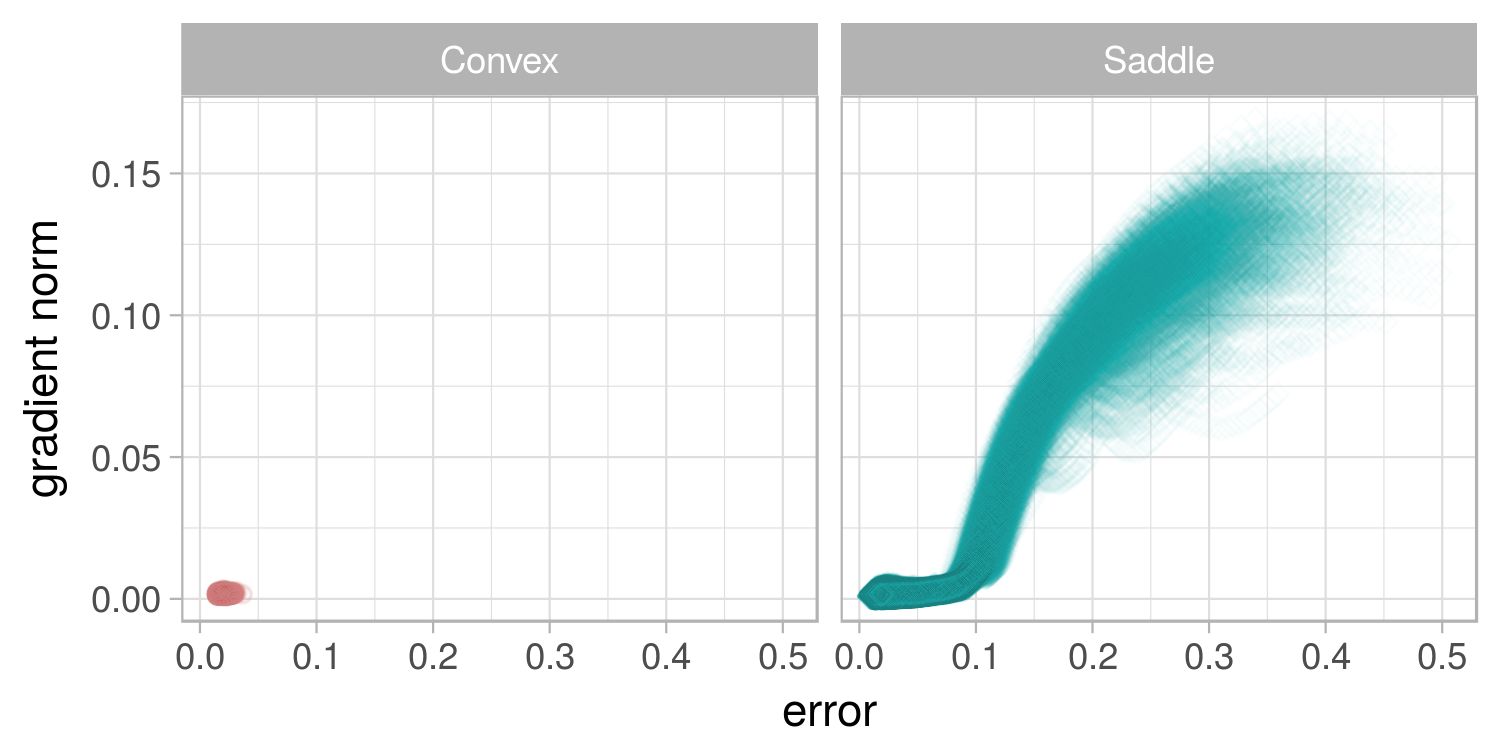}}
  \end{subfloat}
  \begin{subfloat}[{CE, micro, $[-1,1]$ }]{    \includegraphics[height=0.19\textheight]{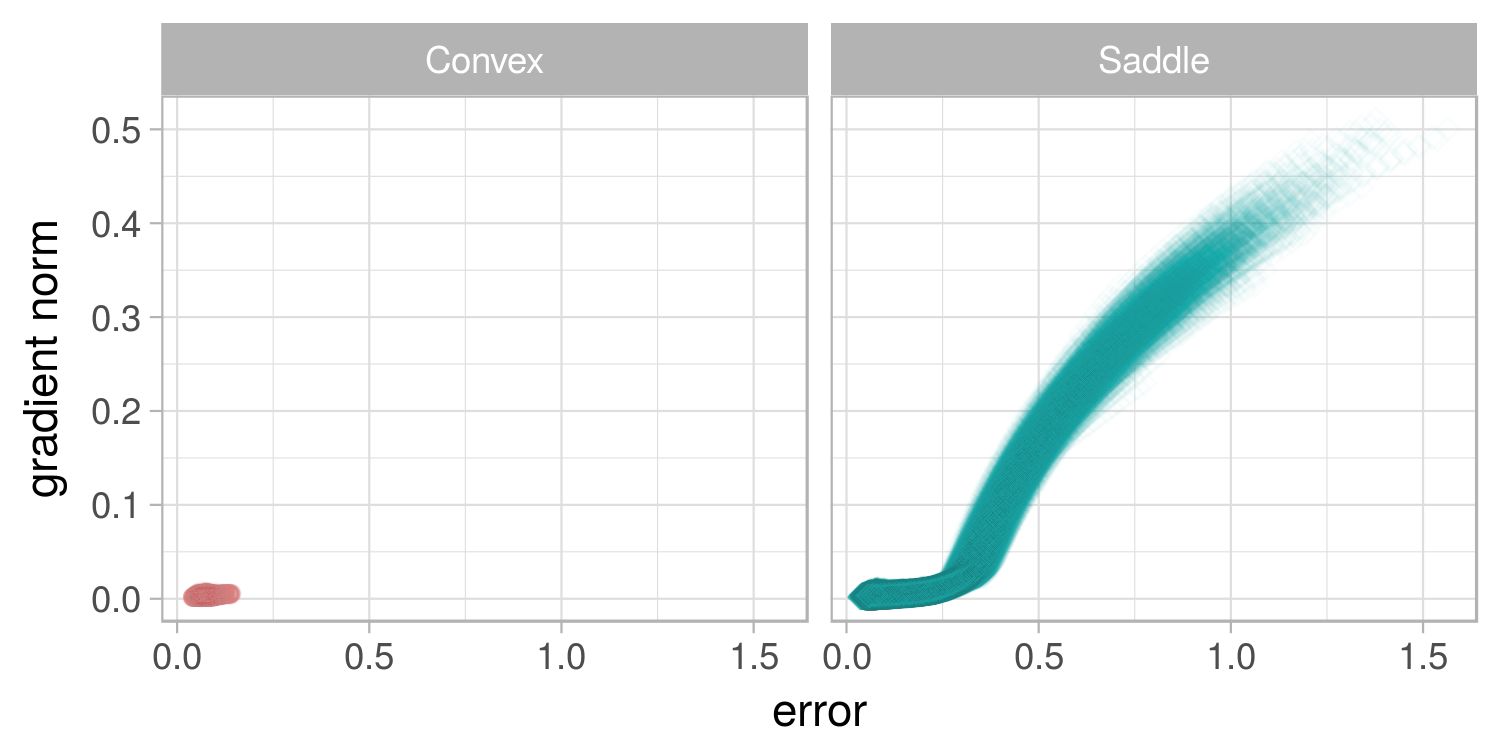}}
  \end{subfloat}\\
  \begin{subfloat}[{SSE, macro, $[-1,1]$ }\label{fig:glass:mse:macro:b1}]{    \includegraphics[height=0.19\textheight]{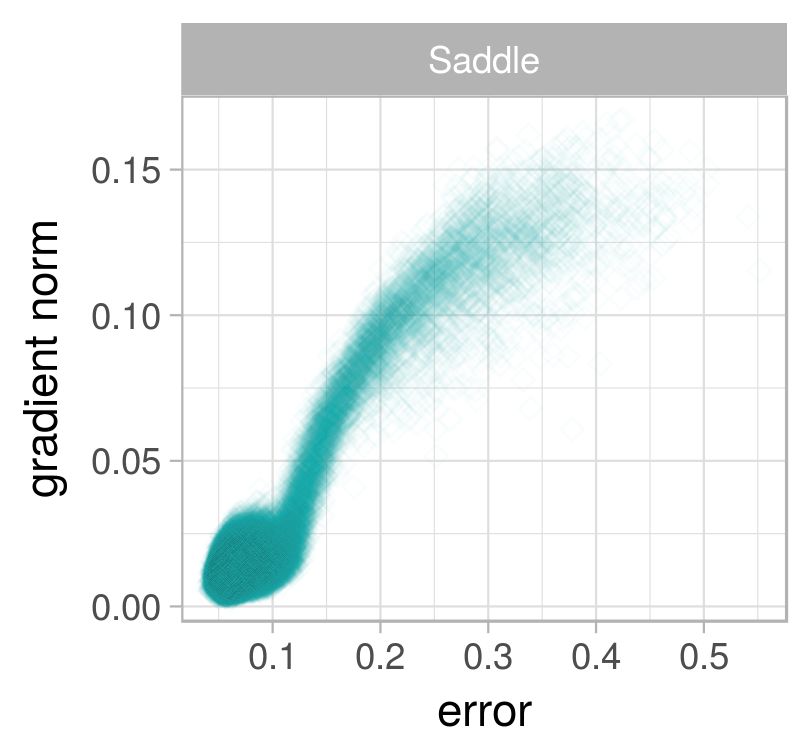}}
  \end{subfloat}
  \begin{subfloat}[{CE, macro, $[-1,1]$ }]{    \includegraphics[height=0.19\textheight]{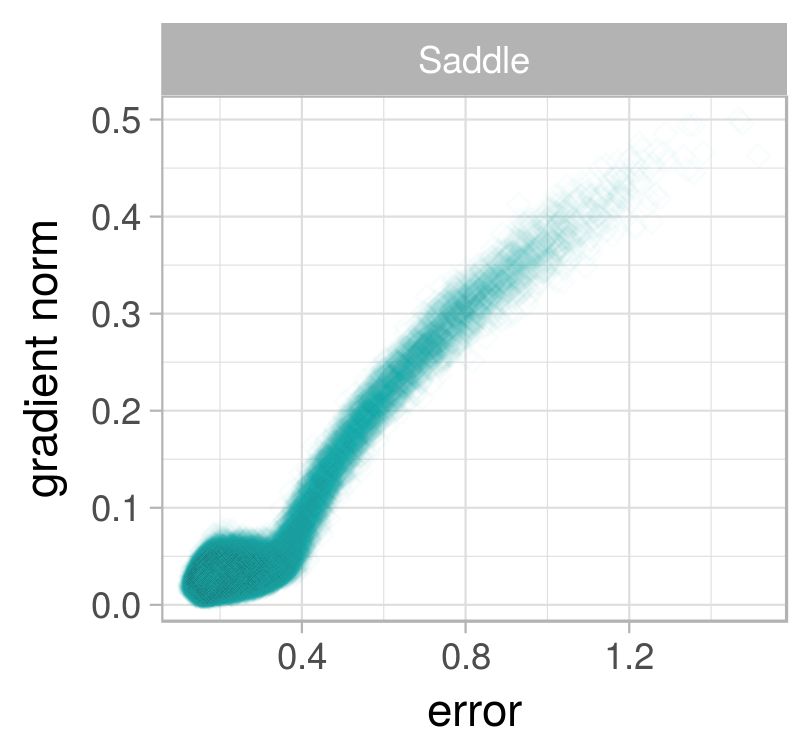}}
  \end{subfloat}
  \caption{L-g clouds for the gradient walks initialised in the $[-1,1]$ range for the Glass problem.}\label{fig:glass:b1}
\end{center}
\end{figure}

\begin{figure}
\begin{center}
  \begin{subfloat}[{SSE, micro, $[-10,10]$ }]{    \includegraphics[height=0.19\textheight]{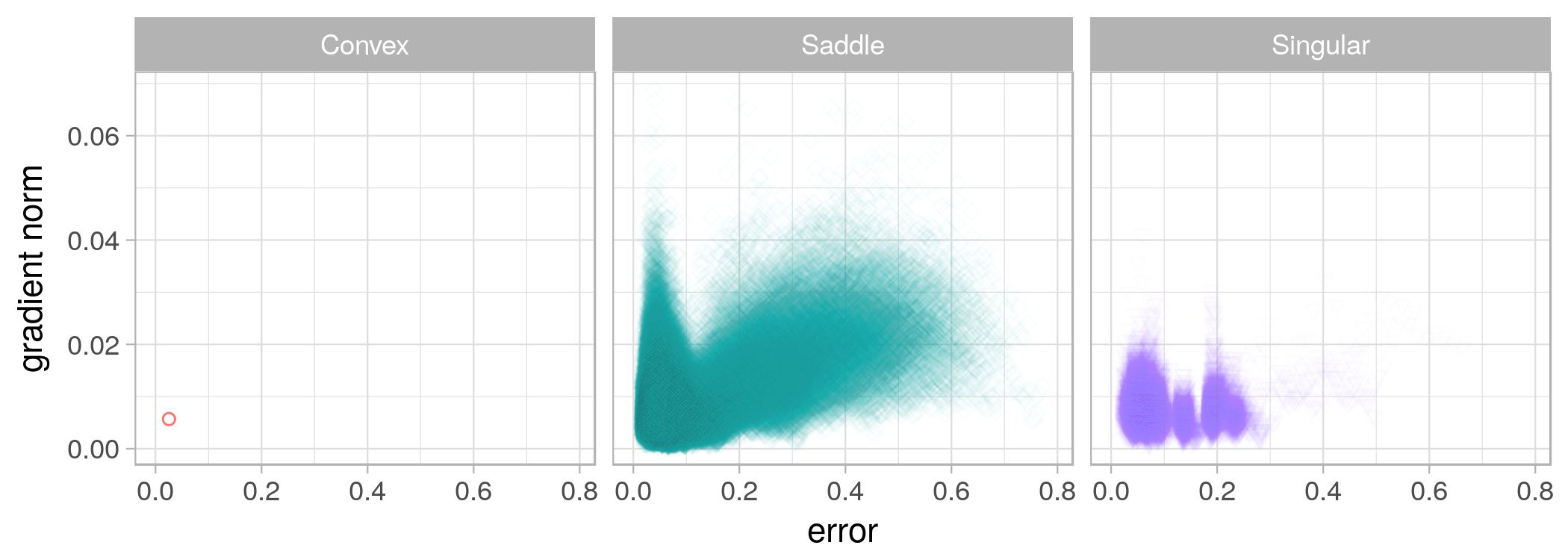}}
  \end{subfloat}
  \begin{subfloat}[{CE, micro, $[-10,10]$ }]{    \includegraphics[height=0.19\textheight]{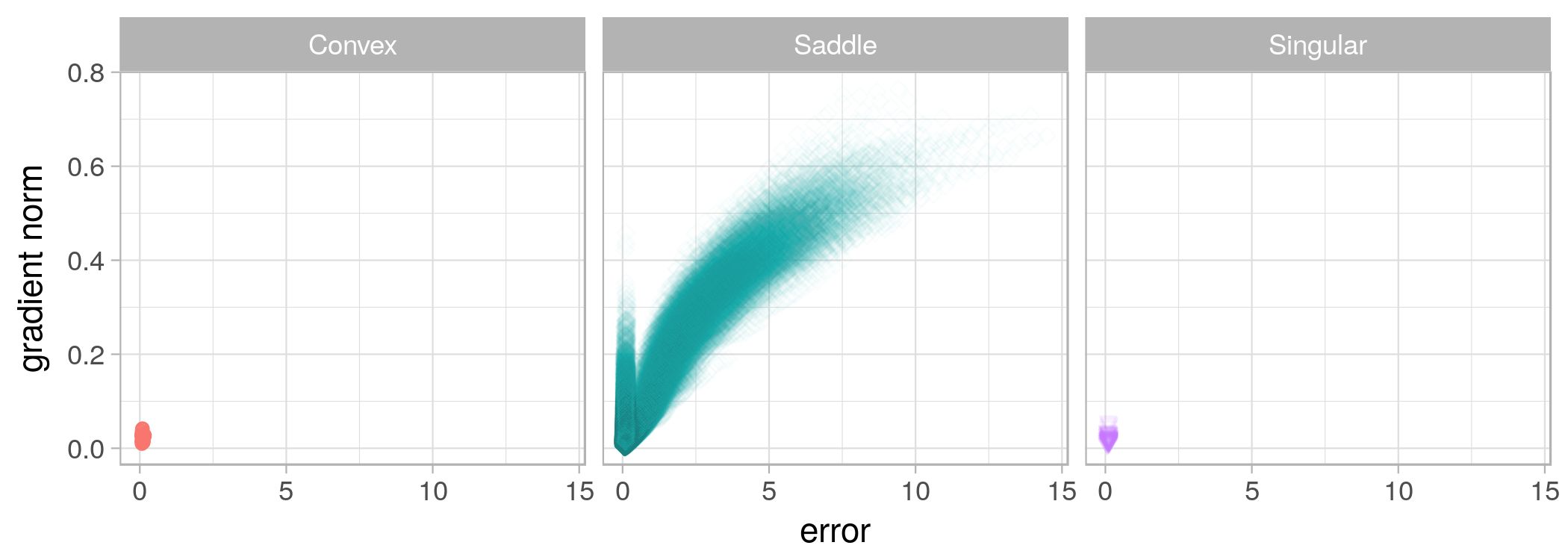}}
  \end{subfloat}
  \begin{subfloat}[{SSE, macro, $[-10,10]$ }]{    \includegraphics[height=0.19\textheight]{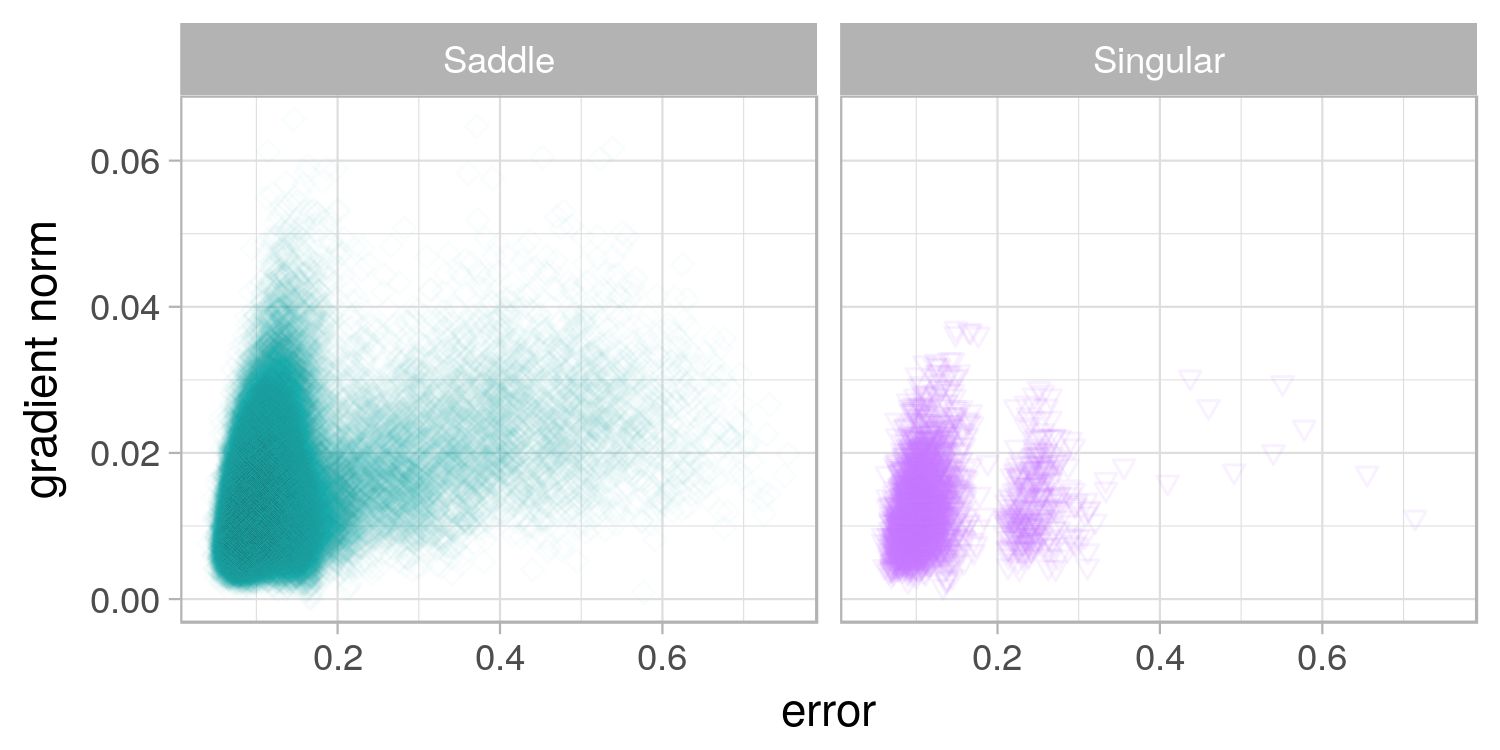}}
  \end{subfloat}
  \begin{subfloat}[{CE, macro, $[-10,10]$ }]{    \includegraphics[height=0.19\textheight]{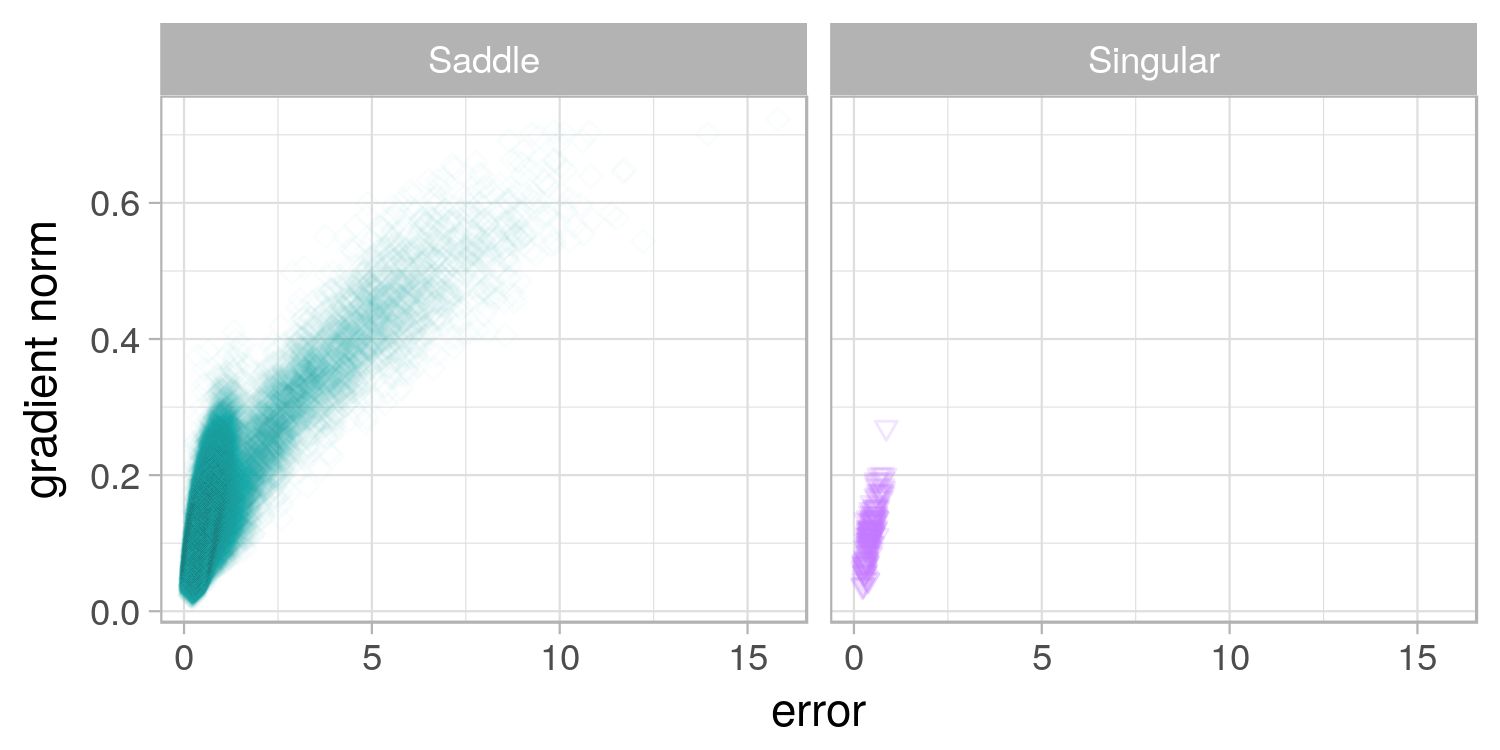}}
  \end{subfloat}
  \caption{L-g clouds for the gradient walks initialised in the $[-1,1]$ range for the Glass problem.}\label{fig:glass:b10}
\end{center}
\end{figure}

Figures~\ref{fig:glass:b1} and~\ref{fig:glass:b10} show the l-g clouds obtained for the Glass problem. According to Figure~\ref{fig:glass:b1}, convexity was found around the global minima only, and only by the micro walks initialised in the $[-1,1]$ range. Macro walks in the same range have discovered exclusively saddle curvature points. This observation once again confirms that the search space for both SSE and CE is dominated by saddle curvature points. Convexity could only be discovered by the smallest steps tested, indicating that the convex area was sharp, and could easily be ``overstepped'' by a larger step size.

From Figure~\ref{fig:glass:b1}, the attractor dynamics exhibited by CE and SSE were quite similar: both losses yielded a general near-linear decline in gradient associated with a decline in error. Once the error became low enough, the gradients flattened, and a further decrease in error towards zero was performed with near-zero gradients. Both CE and SSE exhibited a single major attractor around the global minima, indicating that all near-stationary points discovered by the walks had a similar error value. The macro steps discovered higher gradients around zero error than the micro steps, but the separation into flat and non-flat areas was still evident. This behaviour is likely to be caused by the gradient walks descending to the bottom of a valley first, and then travelling down the bottom of the valley towards a global minimum.

Table~\ref{table:glass:lstag} reports the $n_{stag}$ and $l_{stag}$ values obtained by the various walks on the glass problem. All walks consistently discovered only one attractor. The $l_{stag}$ values indicate that the attractor was found within the first 10\% to 20\% of the steps, and from thereon the walks proceeded to explore the discovered attractor. Thus, all walks quickly descended into a valley, and then travelled at the bottom of the valley for the majority of the steps. It was clearly quite easy to find a valley, and the error values at the bottom of all discovered valleys were rather similar. No inter-valley transition was observed.

\begin{table}
	\caption{Basin of attraction estimates calculated for the Glass problem. Standard deviation shown in parenthesis.}
	\label{table:glass:lstag}
	\begin{center}
		\begin{tabular}{l|cc|cc}
			& \multicolumn{2}{|c|}{SSE} & \multicolumn{2}{|c}{CE} \\
			\hline
			{\bf $E_t$}  &  $n_{stag}$    & $l_{stag}$ & $n_{stag}$ & $l_{stag}$ \\
			\hline
			$[-1,1]$,  & 1.00000 & 947.12067 & 1.00000 & 939.70667 \\
			micro      &(0.00000) &   (7.81525) & (0.00000) &   (7.50773) \\
			$[-1,1]$,  &1.00000 &  86.13867 & 1.00000 &  85.04333 \\
			macro&  (0.00000) &   (0.77595) & (0.00000) &   (1.03672) \\
			$[-10,10]$,&1.04133 & 927.42956 & 1.00000 & 961.23867 \\
			micro& (0.20238) &  (96.74308) & (0.00000) &   (5.18617) \\
			$[-10,10]$, &1.00400 &  85.29156 & 1.00133 &  87.17956 \\
			macro& (0.08155) &   (4.99437) & (0.05162) &   (2.28182) \\
			\hline\hline
			{\bf $E_g$}  &  $n_{stag}$    & $l_{stag}$ & $n_{stag}$ & $l_{stag}$ \\
			\hline
			$[-1,1]$,  & 1.00000 & 951.66867 & 1.00800 & 941.96300 \\
			micro      &(0.00000) &   (8.18235) & (0.10925) &  (42.59343) \\
			$[-1,1]$,  & 1.00000 &  86.67000 & 1.00000 &  86.01267 \\
			macro&  (0.00000) &   (0.57715) & (0.00000) &   (0.71683) \\
			$[-10,10]$,&1.11400 & 902.05250 & 1.02733 & 950.75153 \\
			micro& (0.44084) & (148.80139) & (0.26568) &  (73.87091) \\
			$[-10,10]$, &1.00533 &  85.01433 & 1.00400 &  86.62400 \\
			macro&(0.07283) &   (6.10341) & (0.06312) &   (4.83090) \\
			\hline
		\end{tabular}
	\end{center}
\end{table}

The corresponding $n_{stag}$ and $l_{stag}$ values obtained for $E_g$ indicate that $E_g$ also yielded a single attractor per walk. Standard deviations of $n_{stag}$ and $l_{stag}$ are higher for $E_g$ than for $E_t$, indicating that a steady decrease in $E_t$ was not always associated with a steady decrease in $E_g$. 

\begin{figure}
	\begin{center}
		\begin{subfloat}[{SSE, micro, $[-1,1]$, $E_t< 0.15$ }\label{fig:glass:gen:micro:mse}]{    \includegraphics[width=0.4\textwidth]{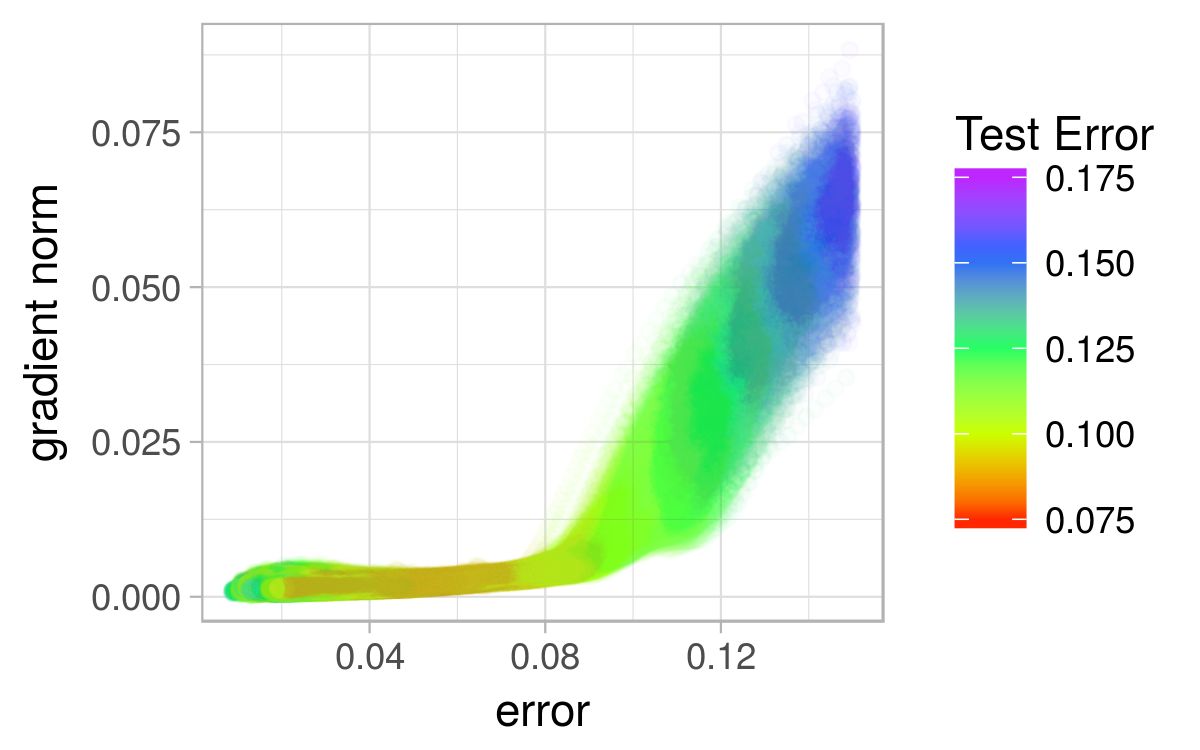}}
		\end{subfloat}
		\begin{subfloat}[{CE, micro, $[-1,1]$, $E_t< 0.5$  }\label{fig:glass:gen:micro:ce}]{    \includegraphics[width=0.4\textwidth]{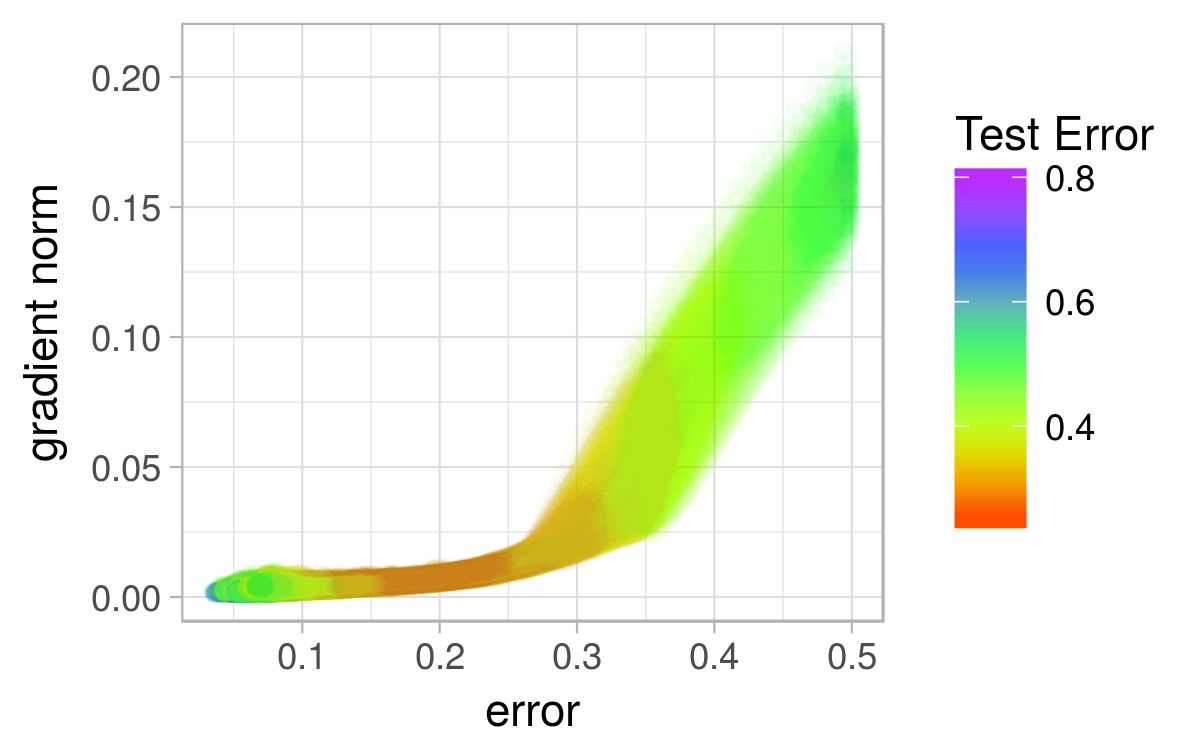}}
		\end{subfloat}
		\begin{subfloat}[{SSE, micro, $[-10,10]$, $E_t< 0.2$ }\label{fig:glass:gen:macro:mse}]{    \includegraphics[width=0.4\textwidth]{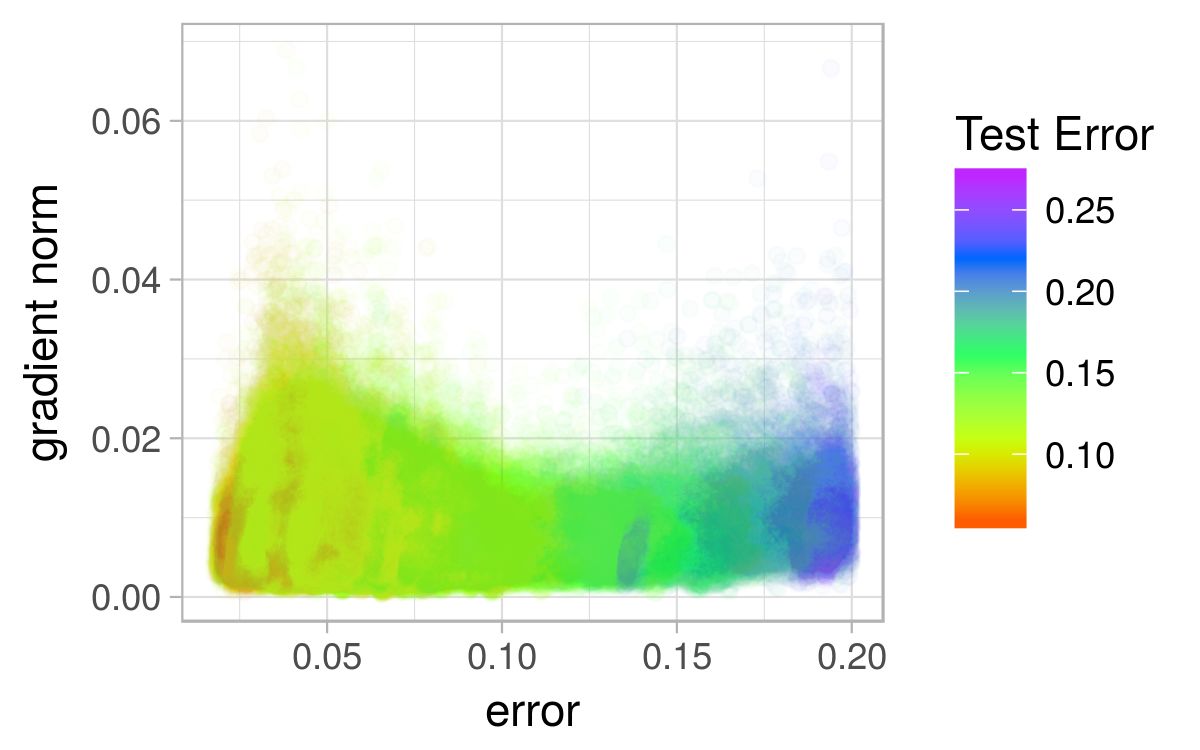}}
		\end{subfloat}
		\begin{subfloat}[{CE, micro, $[-10,10]$, $E_t< 1$  }\label{fig:glass:gen:macro:ce}]{    \includegraphics[width=0.4\textwidth]{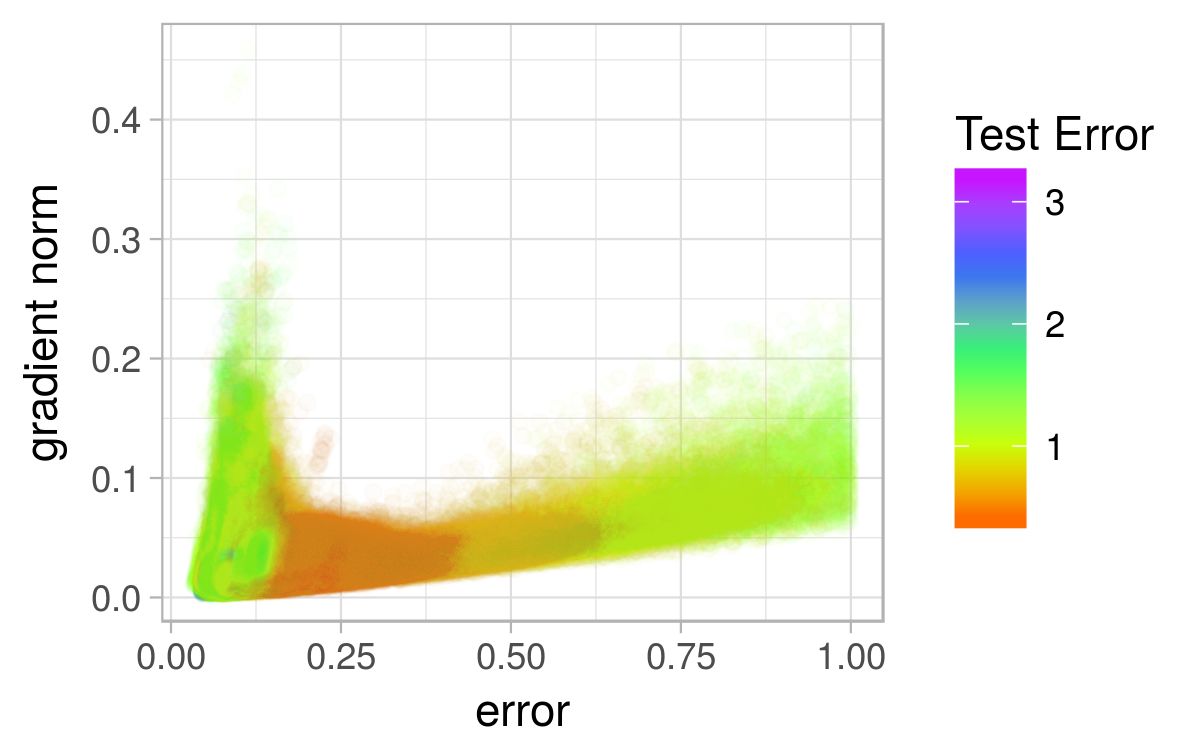}}
		\end{subfloat}
		\caption{L-g clouds colourised according to the corresponding $E_g$ values for the Glass problem.}\label{fig:glass:gen}
	\end{center}
\end{figure}

To further study the generalisation behaviour of the two loss functions, Figure~\ref{fig:glass:gen} shows the l-g clouds of the attractors discovered for CE and SSE, colourised according to the corresponding $E_g$ values. According to Figures~\ref{fig:glass:gen:micro:mse} and~\ref{fig:glass:gen:micro:ce}, both SSE and CE exhibited a decrease in $E_g$ associated with an initial decrease in $E_t$, but for both loss functions $E_g$ increased as $E_t$ approached zero. Thus, once an algorithm has descended into a valley, further exploitation of the valley becomes unnecessary, as far as the generalisation performance is concerned.

Figure~\ref{fig:glass:b10} shows the l-g clouds obtained by the micro and macro walks initialised in the $[-10,10]$ range. According to Figure~\ref{fig:glass:b10}, a larger initialisation range yielded indefinite Hessians, indicating that points of little to no curvature were discovered. A larger initialisation range is more likely to yield exploration of areas further away from the origin. Since the NNs in this study employed the sigmoid activation, the observed flatness is attributed to the saturation of the activation signals. Multiple flat attractors were observed for SSE, while CE exhibited a single major attractor. While this single attractor was at the global minima, the sampled points clustered around two ``paths'': lower errors associated with higher gradients, and higher errors associated with lower gradients. This indicates the presence of two structures: narrow as well as wide valleys.

It has been previously observed that wide valleys are likely to yield better generalisation performance~\cite{ref:Xing:2018,ref:Chaudhari:2017}. There was also a counter-argument presented, where a sharp minimum with good generalisation properties was artificially created~\cite{ref:Dinh:2017}. To study the generalisation performance of the sampled points, the l-g clouds obtained for the $[-10,10]$ micro walks, colourised according to the $E_g$ values, are presented in Figures~\ref{fig:glass:gen:macro:mse} and~\ref{fig:glass:gen:macro:ce}. Figure~\ref{fig:glass:gen:macro:ce} confirms that points of large gradient and low error generalised poorly for CE, while points of higher error and lower gradient generalised better. Thus, points of low error exhibited overfitting for CE loss on the glass problem, and the wide valleys have exhibited better generalisation properties. Interestingly, the same did not hold for SSE loss: according to Figure~\ref{fig:glass:gen:macro:mse}, the smallest $E_g$ was observed for the points of the lowest $E_t$. Thus, SSE loss was less prone to overfitting when sampled at the given resolution. Therefore, despite exhibiting more low gradient attractors, SSE exhibits better generalisation properties in some scenarios. The classification error values reported in Table~\ref{table:glass:class} indicate that SSE and CE have in fact performed very similarly, and have both generalised poorly. The glass dataset is rather small, and small datasets lead to overfitting.

\subsection{Cancer}
\begin{figure}
\begin{center}
  \begin{subfloat}[{SSE, micro, $[-1,1]$ }\label{fig:cancer:mse:micro:b1}]{    \includegraphics[height=0.19\textheight]{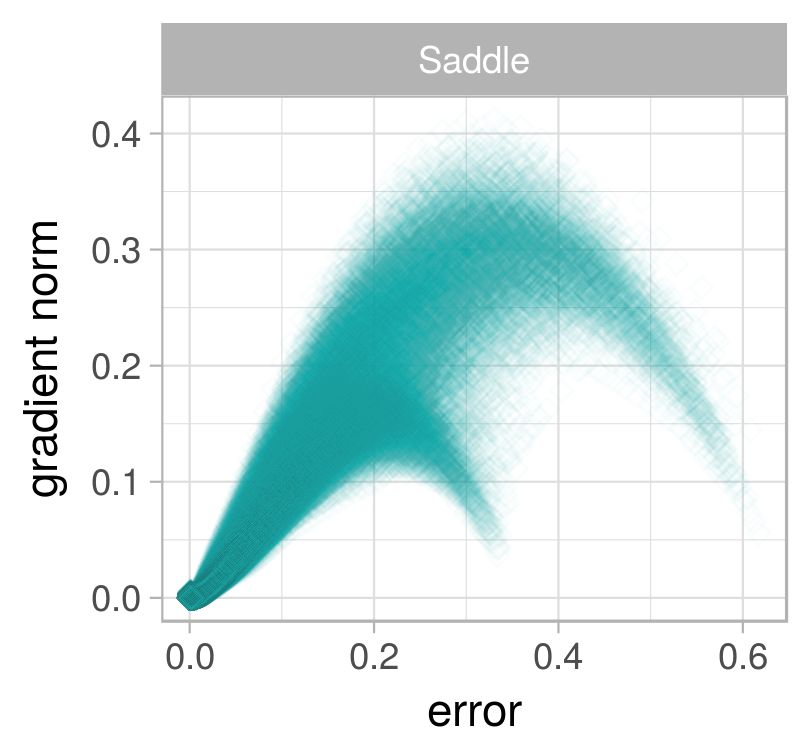}}
  \end{subfloat}
  \begin{subfloat}[{CE, micro, $[-1,1]$ }]{    \includegraphics[height=0.19\textheight]{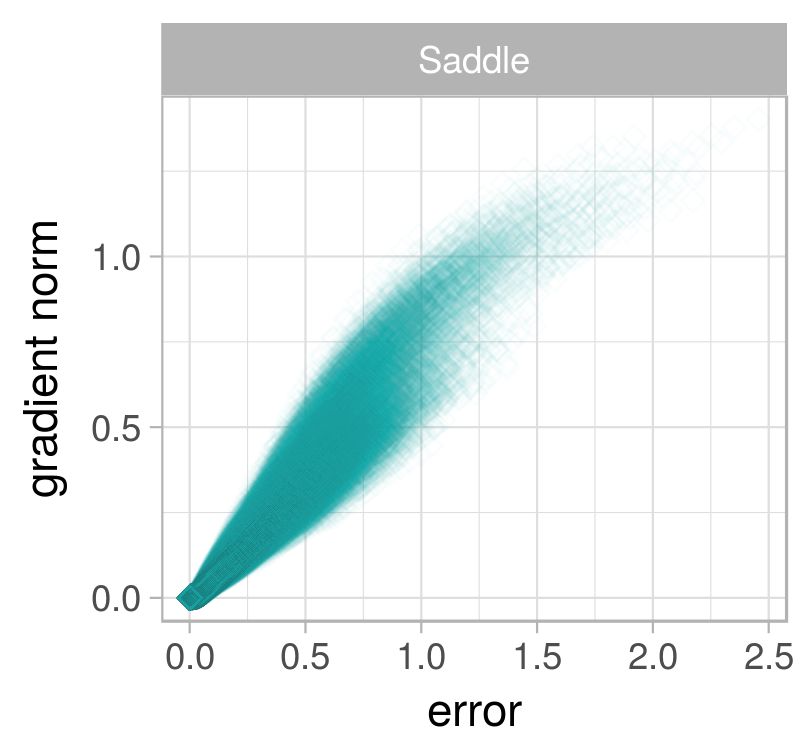}}
  \end{subfloat}\\
  \begin{subfloat}[{SSE, macro, $[-1,1]$ }\label{fig:cancer:mse:macro:b1}]{    \includegraphics[height=0.19\textheight]{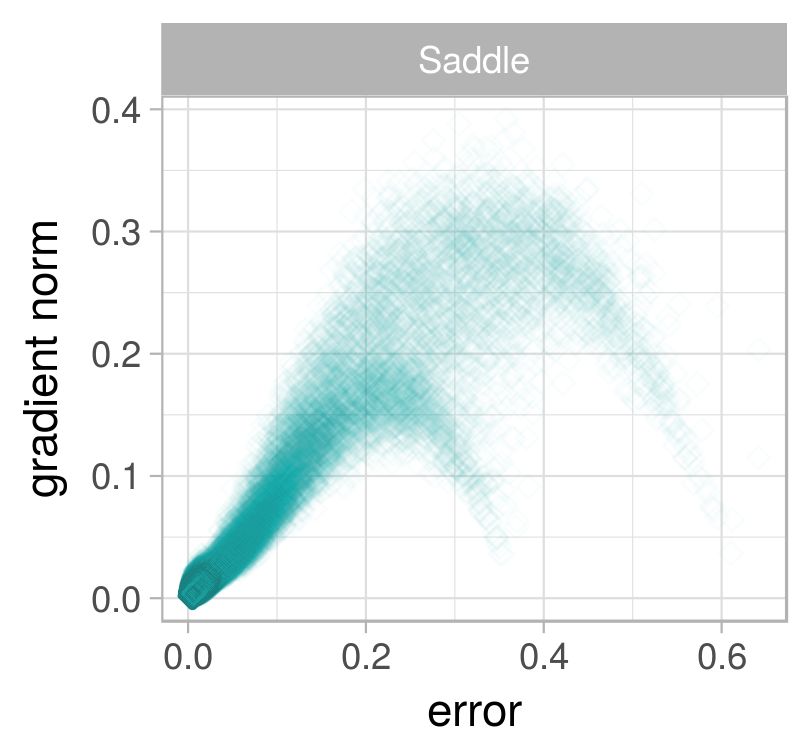}}
  \end{subfloat}
  \begin{subfloat}[{CE, macro, $[-1,1]$ }]{    \includegraphics[height=0.19\textheight]{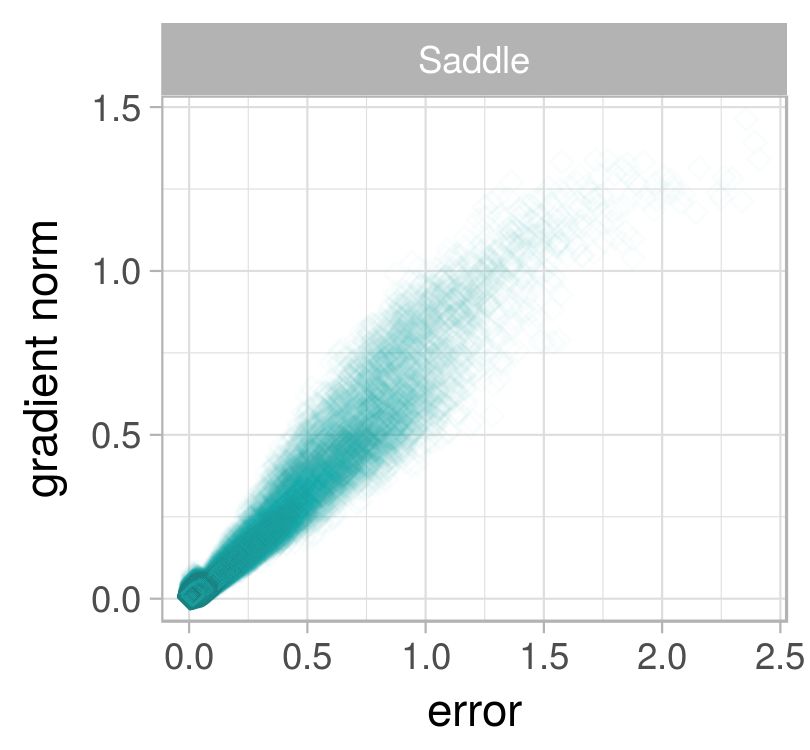}}
  \end{subfloat}
  \caption{L-g clouds for the gradient walks initialised in the $[-1,1]$ range on the Cancer
    problem.}\label{fig:cancer:b1}
\end{center}
\end{figure}

\begin{figure}
\begin{center}
  \begin{subfloat}[{SSE, micro, $[-10,10]$ }]{    \includegraphics[height=0.19\textheight]{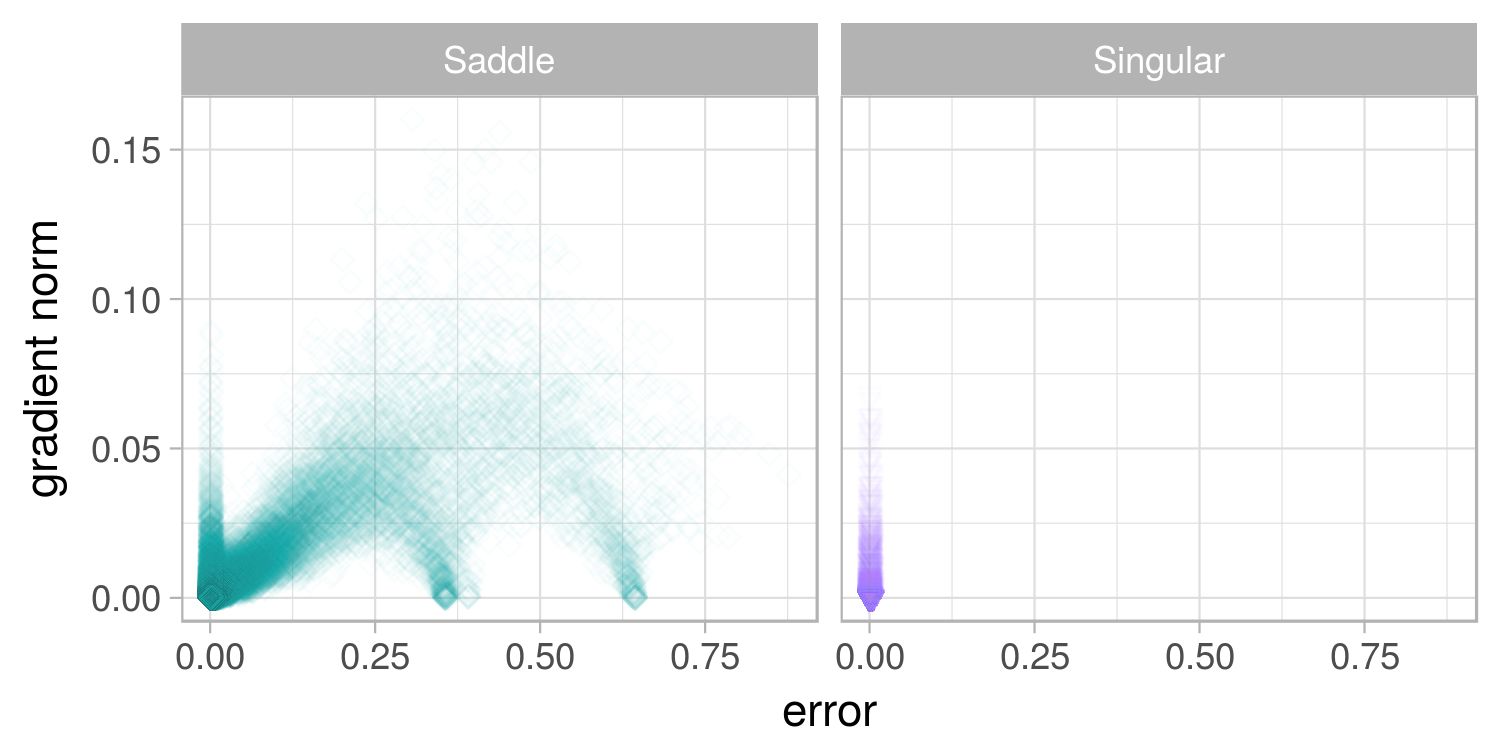}}
  \end{subfloat}
  \begin{subfloat}[{CE, micro, $[-10,10]$ }]{    \includegraphics[height=0.19\textheight]{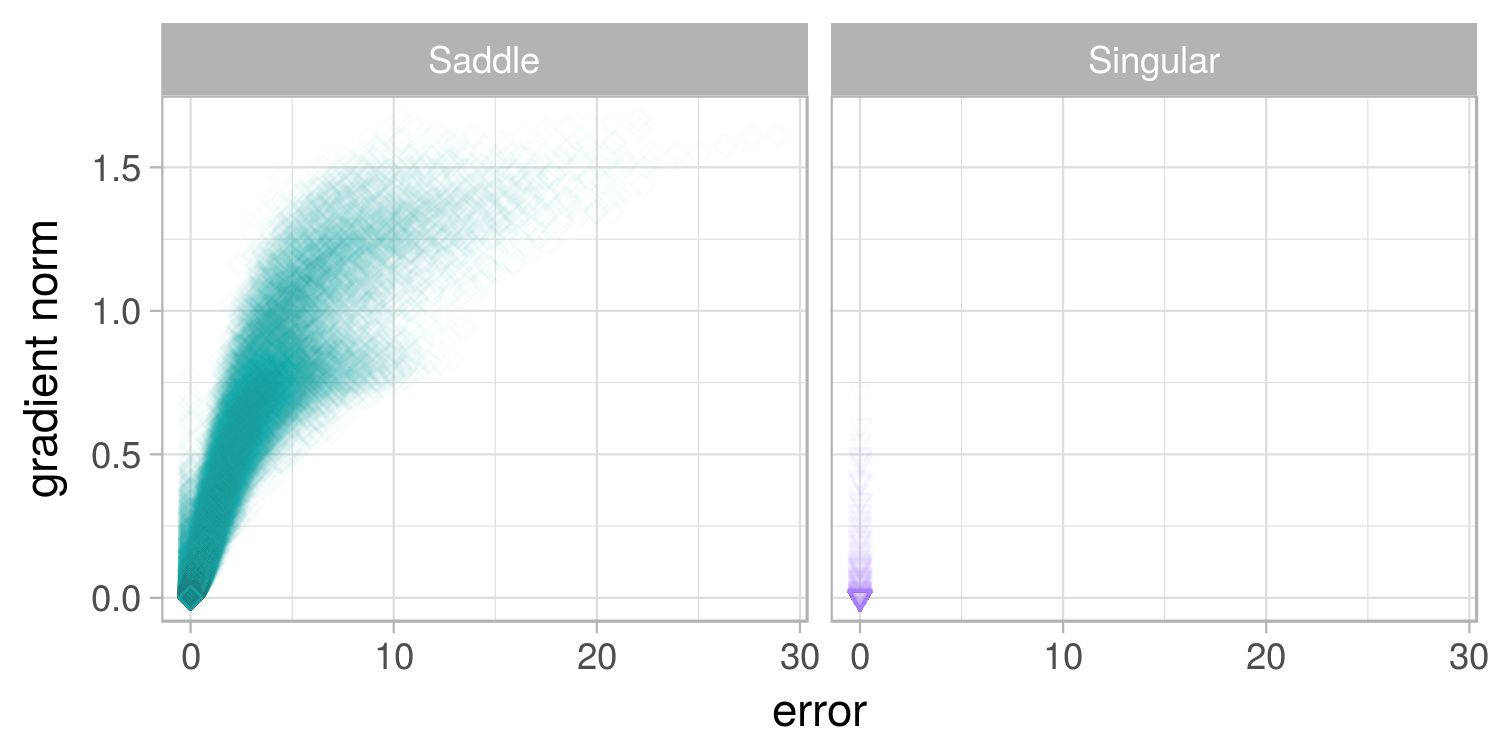}}
  \end{subfloat}
  \begin{subfloat}[{SSE, macro, $[-10,10]$ }]{    \includegraphics[height=0.19\textheight]{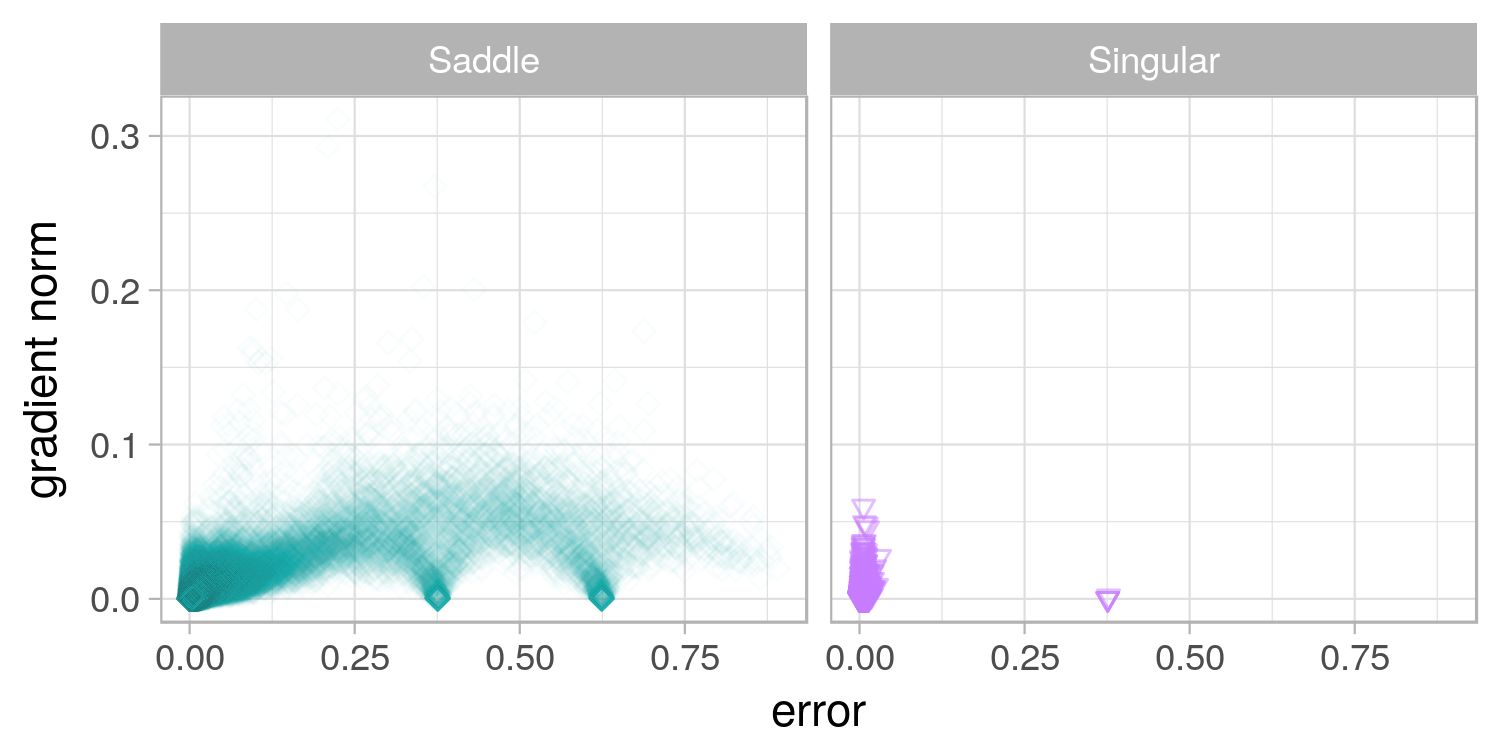}}
  \end{subfloat}
  \begin{subfloat}[{CE, macro, $[-10,10]$ }]{    \includegraphics[height=0.19\textheight]{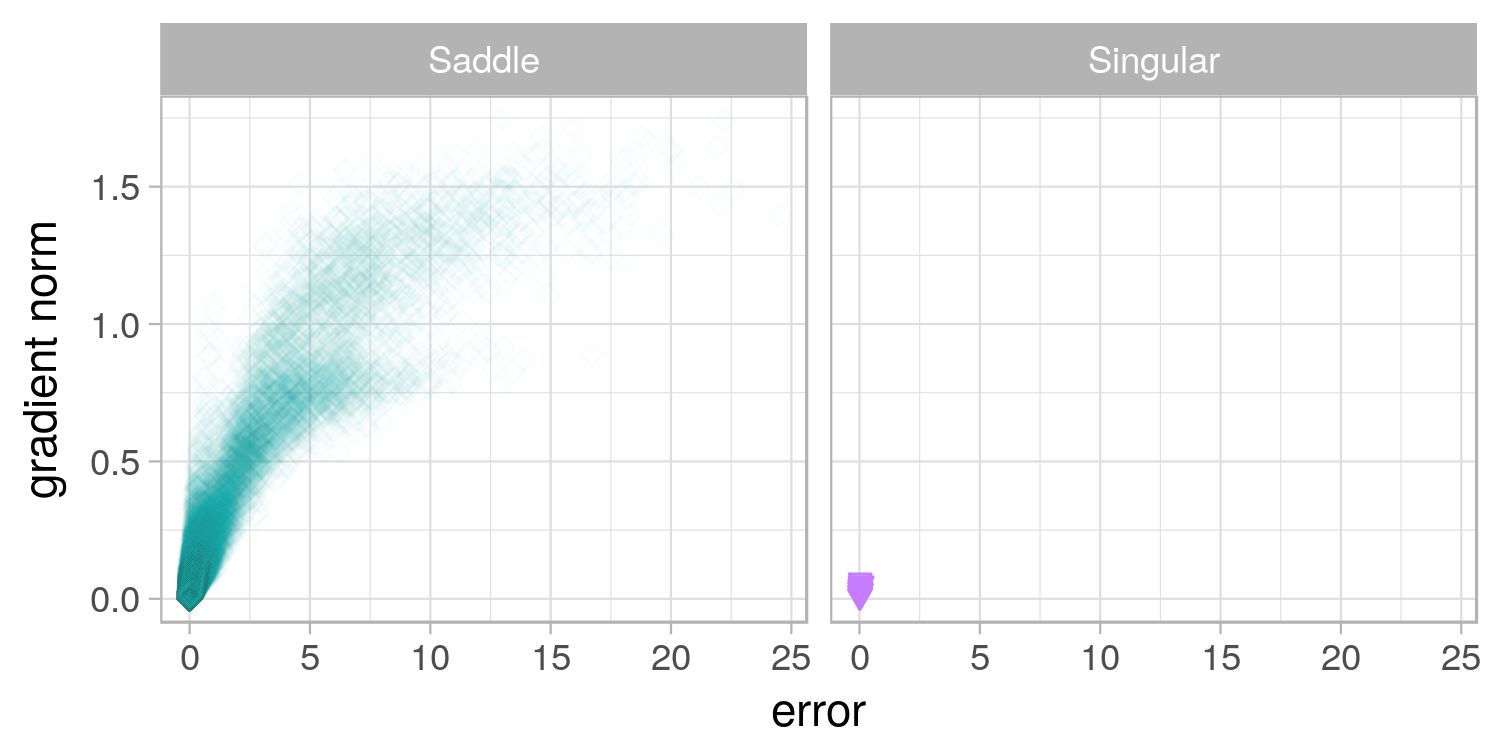}}
  \end{subfloat}
  \caption{L-g clouds for the gradient walks initialised in the $[-1,1]$ range on the Cancer
    problem.}\label{fig:cancer:b10}
\end{center}
\end{figure}

Figures~\ref{fig:cancer:b1} and~\ref{fig:cancer:b10} show the l-g clouds obtained for the Cancer problem. According to Figure~\ref{fig:cancer:b1}, all points sampled by micro and macro walks initialised in the $[-1,1]$ range exhibited saddle curvature. Total dimensionality of the cancer problem is 321, which is noticeably higher than that of the previous problems considered. Saddle curvature is expected to become more and more prevalent as the dimensionality increases~\cite{ref:Dauphin:2014}.

According to Figure~\ref{fig:cancer:b1}, both SSE and CE had one major attractor at the global minima. In addition to the global minima, SSE exhibited two more attractors of low, but non-zero gradient. Trajectories can be observed leading to the global minima from either of the two high error attractors. However, there is no trajectory connecting the attractors to one another. The $n_{stag}$ and $l_{stag}$ values reported in Table~\ref{table:cancer:lstag} confirm
that all walks discovered a single attractor only, thus no transition between the attractors took place.

\begin{table}
  \caption{Basin of attraction estimates calculated for the Cancer problem. Standard deviation shown in parenthesis.}
  \label{table:cancer:lstag}
  \begin{center}
    \begin{tabular}{l|cc|cc}
      & \multicolumn{2}{|c|}{SSE} & \multicolumn{2}{|c}{CE} \\
      \hline
          {\bf $E_t$}  &  $n_{stag}$    & $l_{stag}$ & $n_{stag}$ & $l_{stag}$ \\
          \hline          
          $[-1,1]$,  & 1.00000 & 962.09844 & 1.00000 & 953.89307 \\
          micro      &(0.00000) &   (5.39294) & (0.00000) &   (5.37201) \\
          $[-1,1]$,&1.00000 &  87.77788 & 1.00000 &  87.18816 \\
          macro& (0.00000) &   (0.44464) & (0.00000) &   (0.51044) \\
          $[-10,10]$,& 1.00000 & 972.77778 & 1.00000 & 975.43836 \\
          micro      &(0.00000) &   (7.45025) & (0.00000) &   (3.01133) \\
          $[-10,10]$,& 1.00000 &  87.44517 & 1.00000 &  87.80498 \\
          macro& (0.00000) &   (0.89423) & (0.00000) &   (1.12240) \\
          \hline\hline
              {\bf $E_g$}  &  $n_{stag}$    & $l_{stag}$ & $n_{stag}$ & $l_{stag}$ \\
              \hline
              $[-1,1]$,  &  1.00000 & 959.38629 & 1.00725 & 953.80766 \\
              micro      &(0.00000) &   (5.62718) & (0.09002) &  (41.40420) \\
              $[-1,1]$,&1.00000 &  87.81838 & 1.00000 &  87.25514 \\
              macro&(0.00000) &   (0.41360) & (0.00000) &   (0.54000) \\
              $[-10,10]$,&  1.11111 & 932.44444 & 4.13699 & 541.93665 \\
              micro      & (0.45812) & (154.23691) & (4.50666) & (384.89318) \\
              $[-10,10]$,& 1.00125 &  87.16246 & 1.00903 &  86.55711 \\
              macro&  (0.03528) &   (2.95538) & (0.09786) &   (7.86466) \\
              \hline
    \end{tabular}
  \end{center}
\end{table}

CE, as shown in Figure~\ref{fig:cancer:b1}, exhibited almost linear correlation between the gradient and the error. Such simple correlation implies that the CE loss surface is likely to be more searchable than the SSE loss surface from the perspective of a gradient-based optimisation algorithm. The cancer problem is known to be an easy classification problem, which must have contributed to the simplicity of the observed attractor.

Figure~\ref{fig:cancer:b10} shows the l-g clouds for the micro and macro walks initialised in the $[-10,10]$ range. The larger initialisation range once again exposed points with indefinite Hessians for both SSE and CE, i.e. points with little to no curvature. For CE, the points of no curvature aligned with the global minima. For SSE, the global minima, as well as the other two attractors, exhibited flatness. The majority of the points exhibited saddle curvature. The two zero-gradient attractors, away from the global minima, were observed for the SSE loss surface. The CE loss surface did not exhibit multiple attractors. However, multiple points of high gradient close to the global minima were sampled. This  once again indicates that CE is more prone to sharp minima (narrow valleys) than SSE.

\begin{figure}
\begin{center}
  \begin{subfloat}[{SSE, micro, $[-1,1]$, $E_t< 0.05$ }\label{fig:cancer:gen:micro:mse}]{    \includegraphics[width=0.4\textwidth]{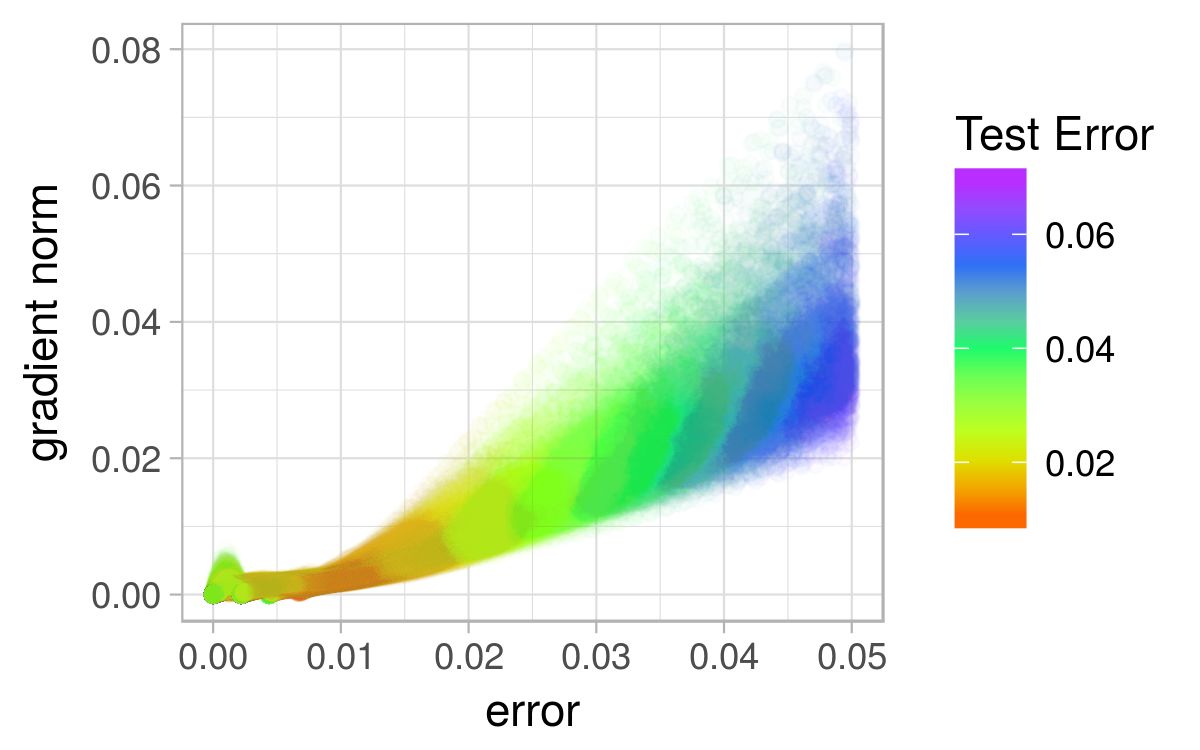}}
  \end{subfloat}
  \begin{subfloat}[{CE, micro, $[-1,1]$, $E_t< 0.05$  }\label{fig:cancer:gen:micro:ce}]{    \includegraphics[width=0.4\textwidth]{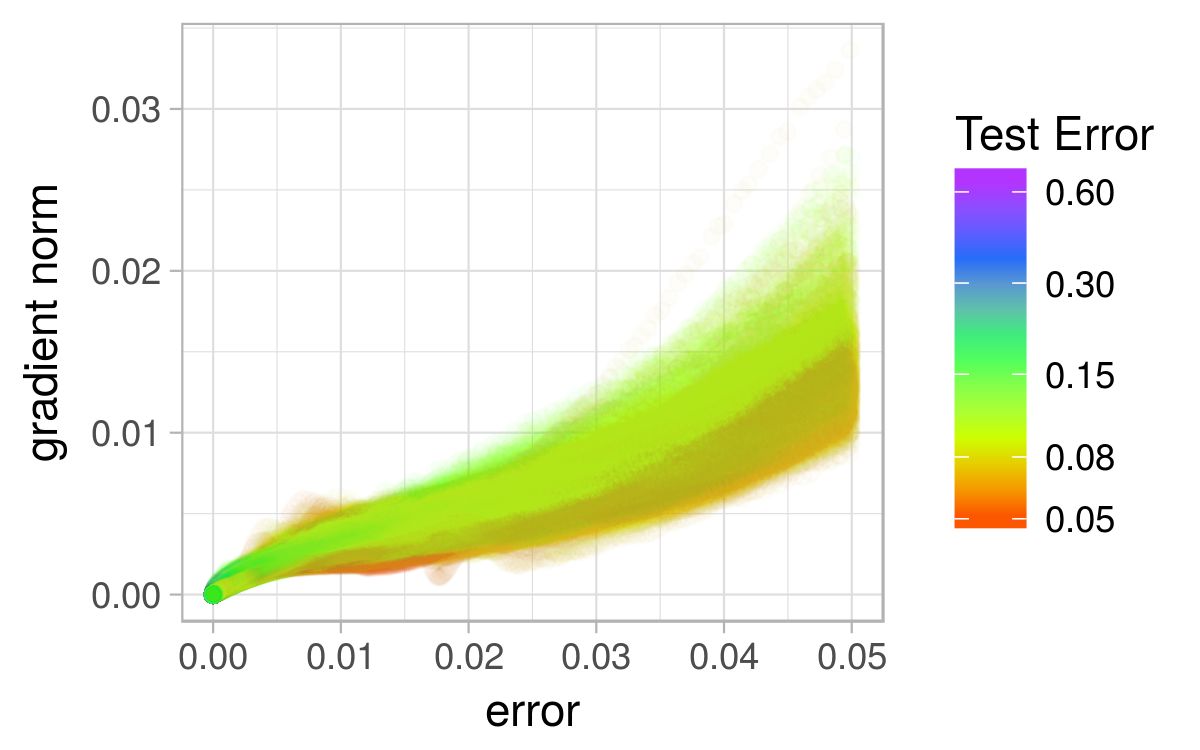}}
  \end{subfloat}
  \begin{subfloat}[{SSE, micro, $[-10,10]$, $E_t< 0.05$ }\label{fig:cancer:gen:macro:mse}]{    \includegraphics[width=0.4\textwidth]{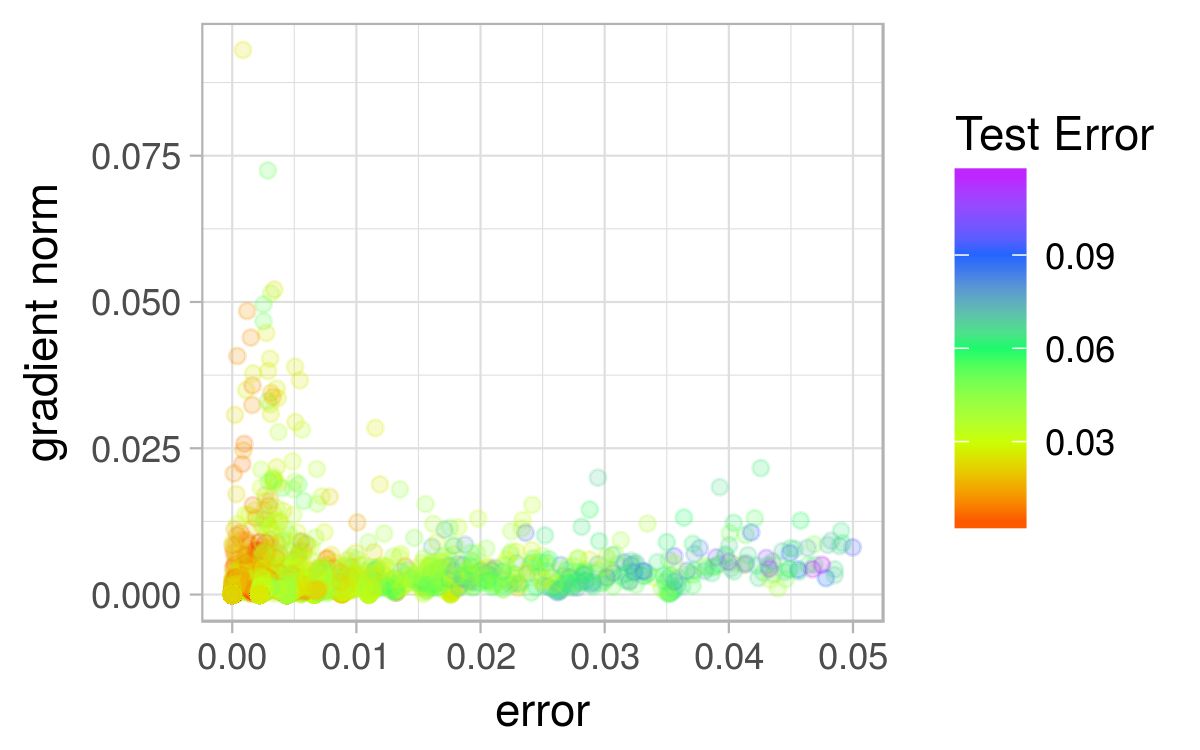}}
  \end{subfloat}
  \begin{subfloat}[{CE, micro, $[-10,10]$, $E_t< 0.1$  }\label{fig:cancer:gen:macro:ce}]{    \includegraphics[width=0.4\textwidth]{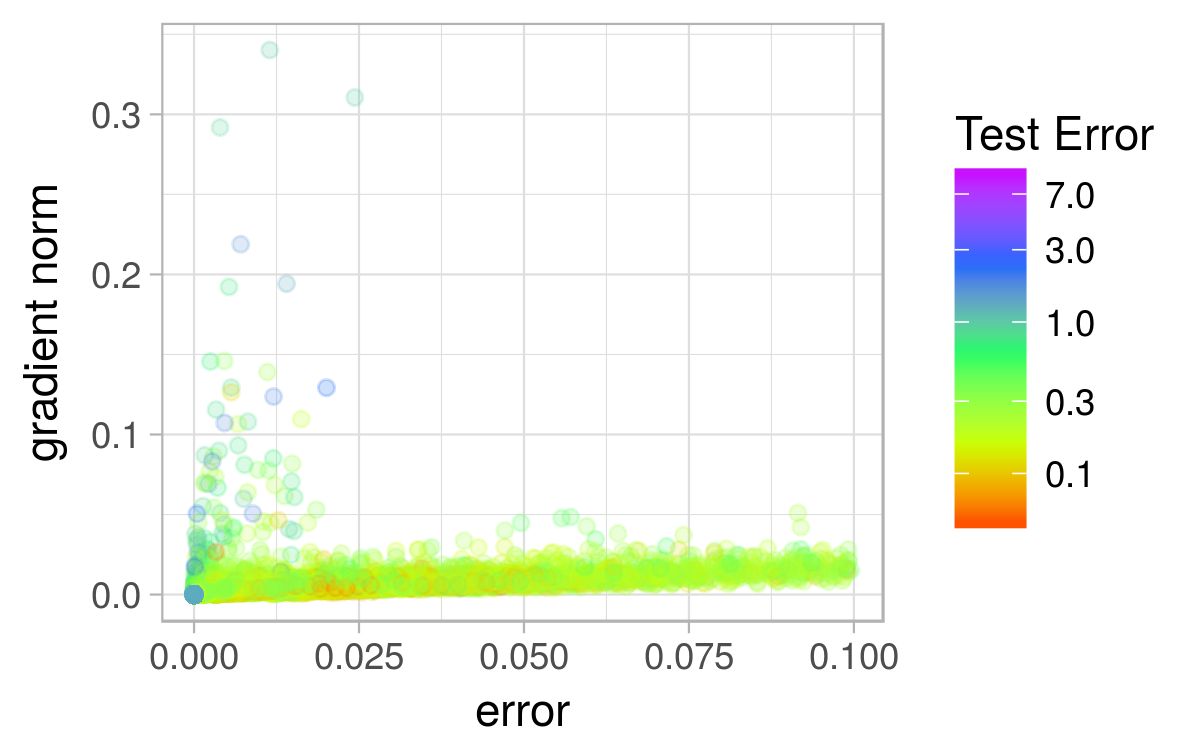}}
  \end{subfloat}
  \caption{L-g clouds colourised according to the corresponding $E_g$ values for the Cancer problem.}\label{fig:cancer:gen}
\end{center}
\end{figure}

The $n_{stag}$ and $l_{stag}$ values yielded by $E_g$ of the sampled points (Table~\ref{table:cancer:lstag}) are inconsistent with the corresponding $n_{stag}$ and $l_{stag}$ values obtained for $E_t$. To further study this inconsistency, Figure~\ref{fig:cancer:gen} presents the l-g clouds colourised according to the $E_g$ values for the points around the global minima. Due to high disparity in the $E_g$ values obtained for CE, the CE l-g clouds were colourised on logarithmic scale. Similar to the previous problems considered, the generalisation performance at the global optimum was poor for both SSE and CE. However, it is evident from Figures~\ref{fig:cancer:gen:macro:mse} and~\ref{fig:cancer:gen:macro:ce} that low error, high gradient points around the global minima generalised well for SSE, and poorly for CE. SSE in general produced weaker gradients than CE, indicating that SSE was less prone to sharp minima. Figure~\ref{fig:cancer:gen} also shows that SSE exhibited points of zero gradient for non-zero error, while CE did not. However, the observed local minima, as well as the global minima of SSE, can yield better generalisation performance than the global minima exhibited by CE.

\subsection{Heart}
\begin{figure}
\begin{center}
  \begin{subfloat}[{SSE, micro, $[-1,1]$ }\label{fig:heart:mse:micro:b1}]{    \includegraphics[height=0.19\textheight]{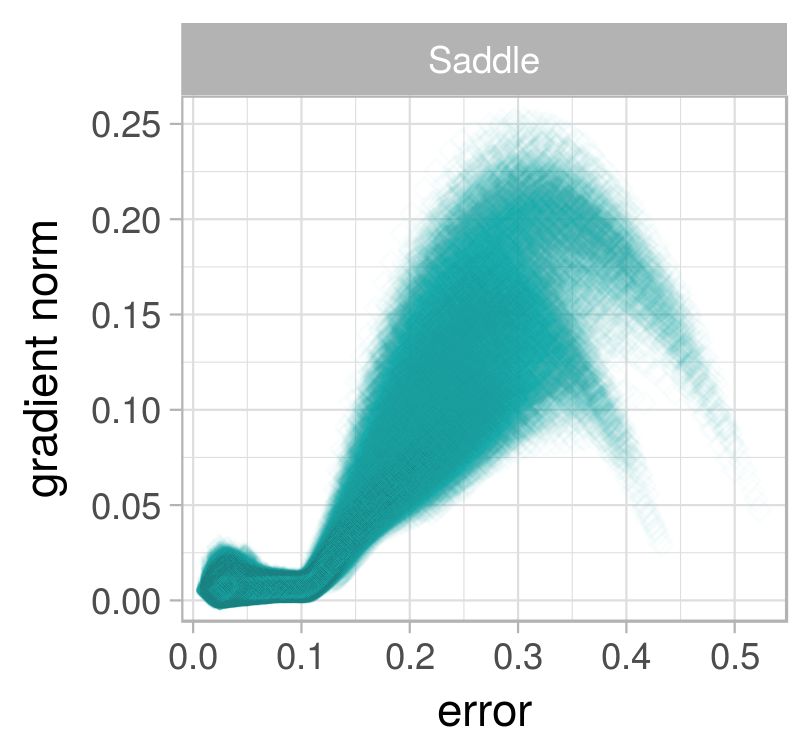}}
  \end{subfloat}
  \begin{subfloat}[{CE, micro, $[-1,1]$ }]{    \includegraphics[height=0.19\textheight]{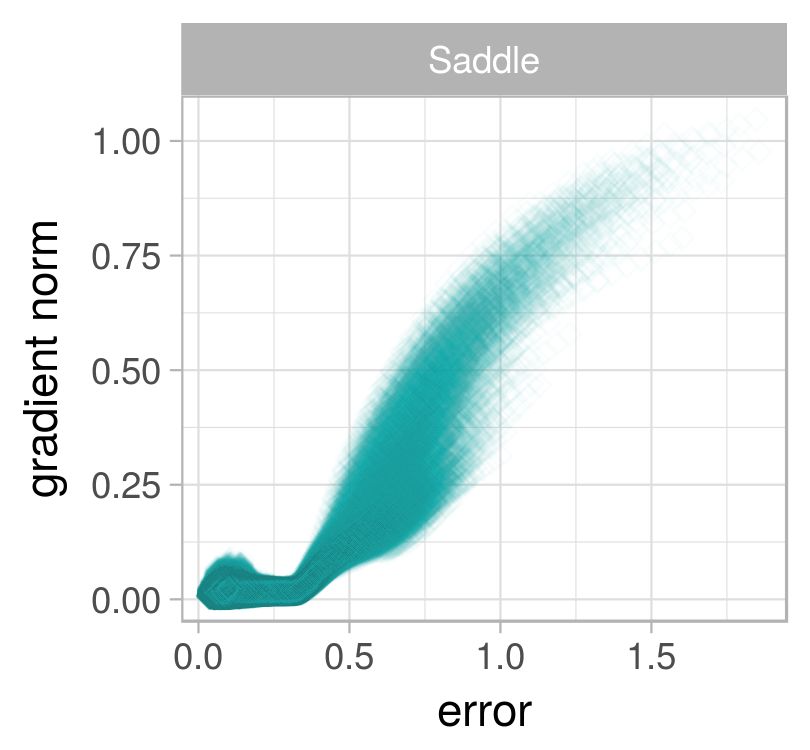}}
  \end{subfloat}\\
  \begin{subfloat}[{SSE, macro, $[-1,1]$ }\label{fig:heart:mse:macro:b1}]{    \includegraphics[height=0.19\textheight]{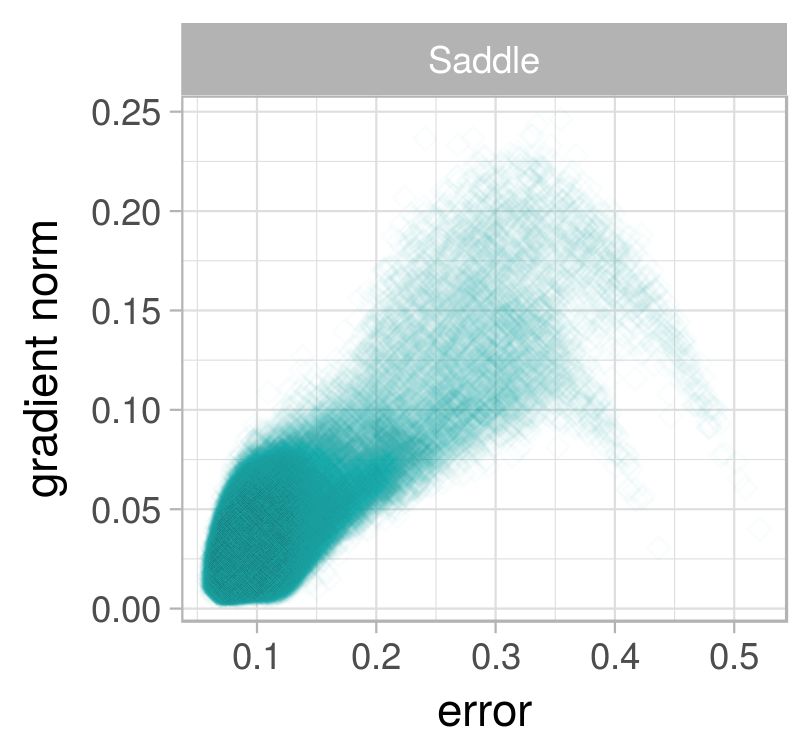}}
  \end{subfloat}
  \begin{subfloat}[{CE, macro, $[-1,1]$ }]{    \includegraphics[height=0.19\textheight]{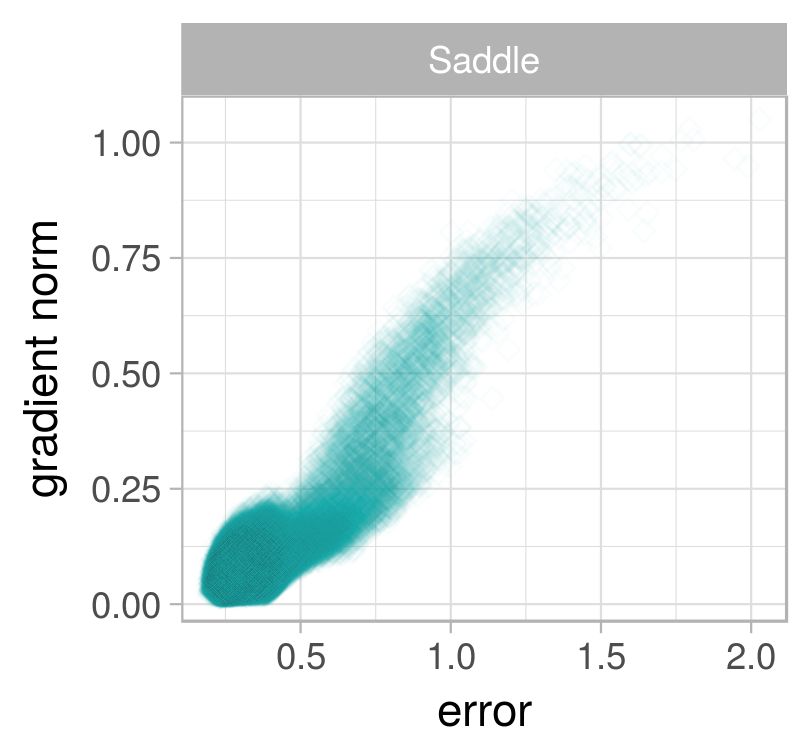}}
  \end{subfloat}
  \caption{L-g clouds for the gradient walks initialised in the $[-1,1]$ range on the Heart
    problem.}\label{fig:heart:b1}
\end{center}
\end{figure}

\begin{figure}
\begin{center}
  \begin{subfloat}[{SSE, micro, $[-10,10]$ }]{    \includegraphics[height=0.19\textheight]{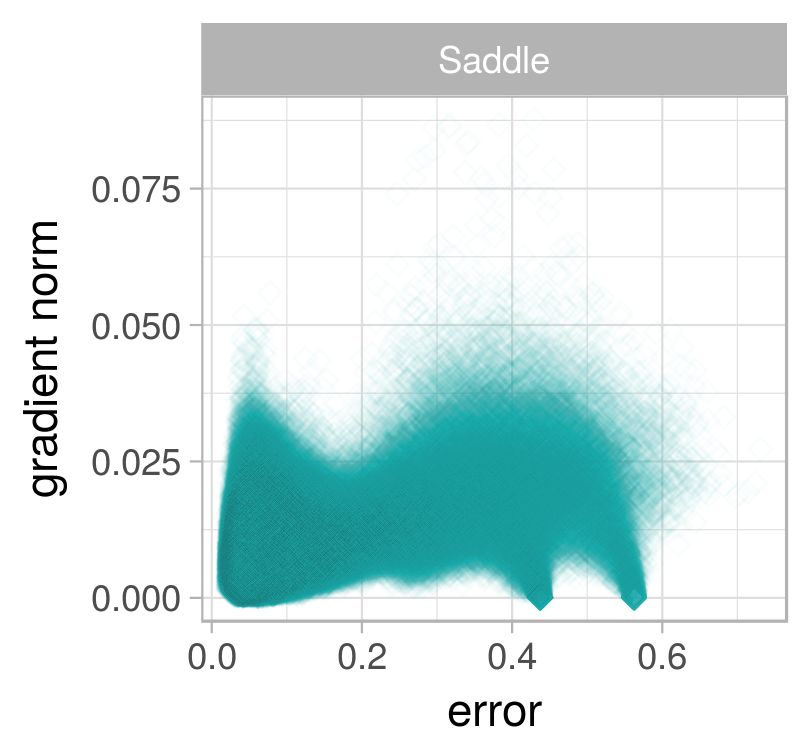}}
  \end{subfloat}
  \begin{subfloat}[{CE, micro, $[-10,10]$ }]{    \includegraphics[height=0.19\textheight]{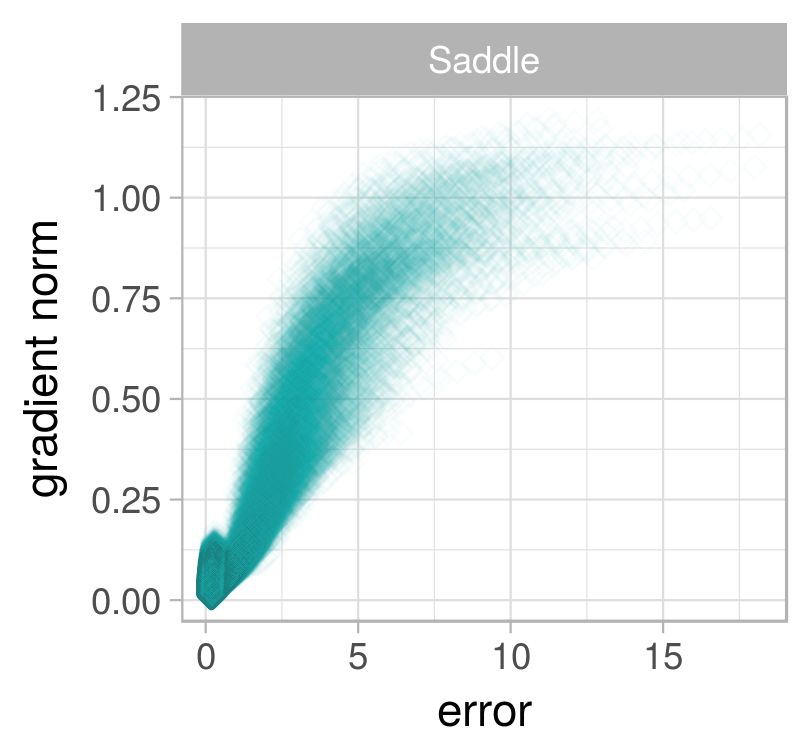}}
  \end{subfloat}
  \begin{subfloat}[{SSE, macro, $[-10,10]$ }]{    \includegraphics[height=0.19\textheight]{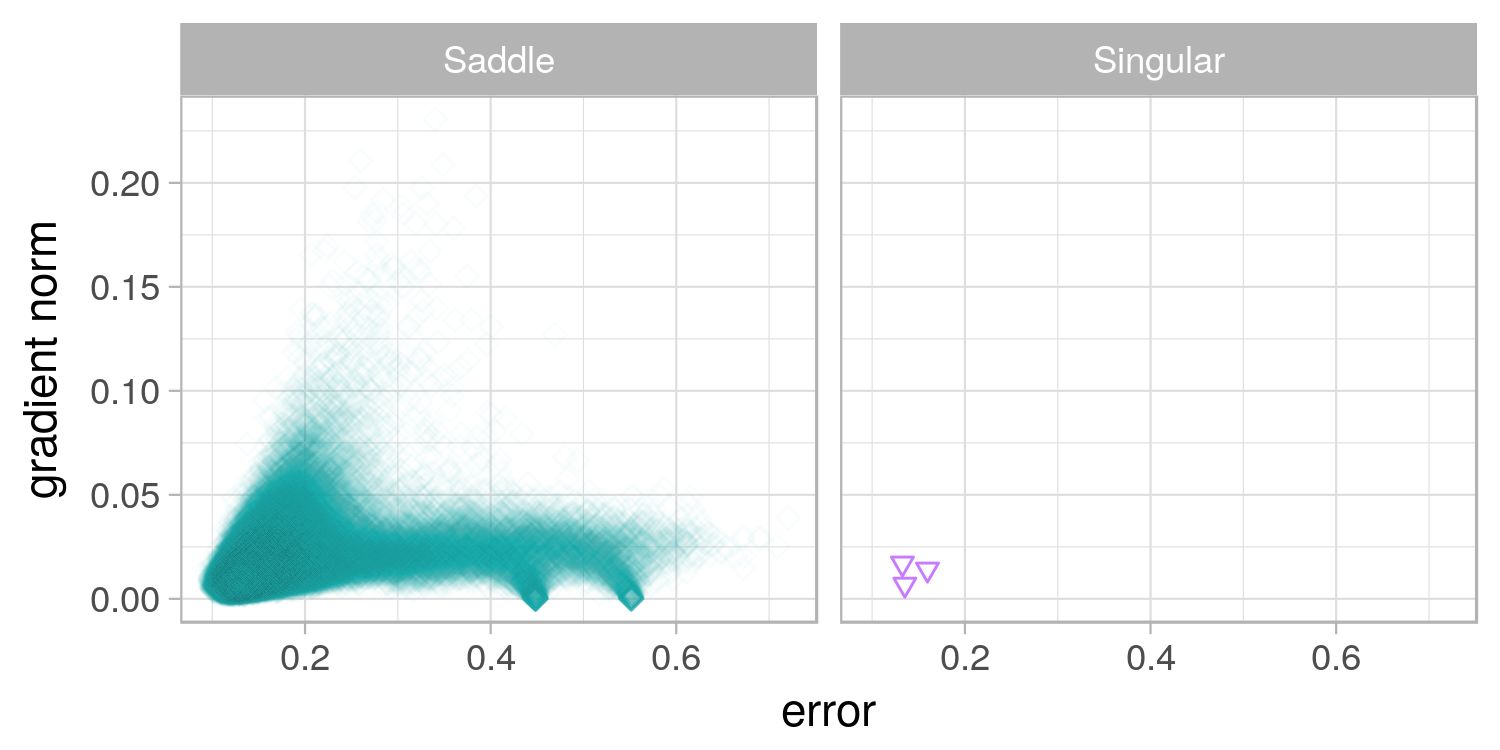}}
  \end{subfloat}
  \begin{subfloat}[{CE, macro, $[-10,10]$ }]{    \includegraphics[height=0.19\textheight]{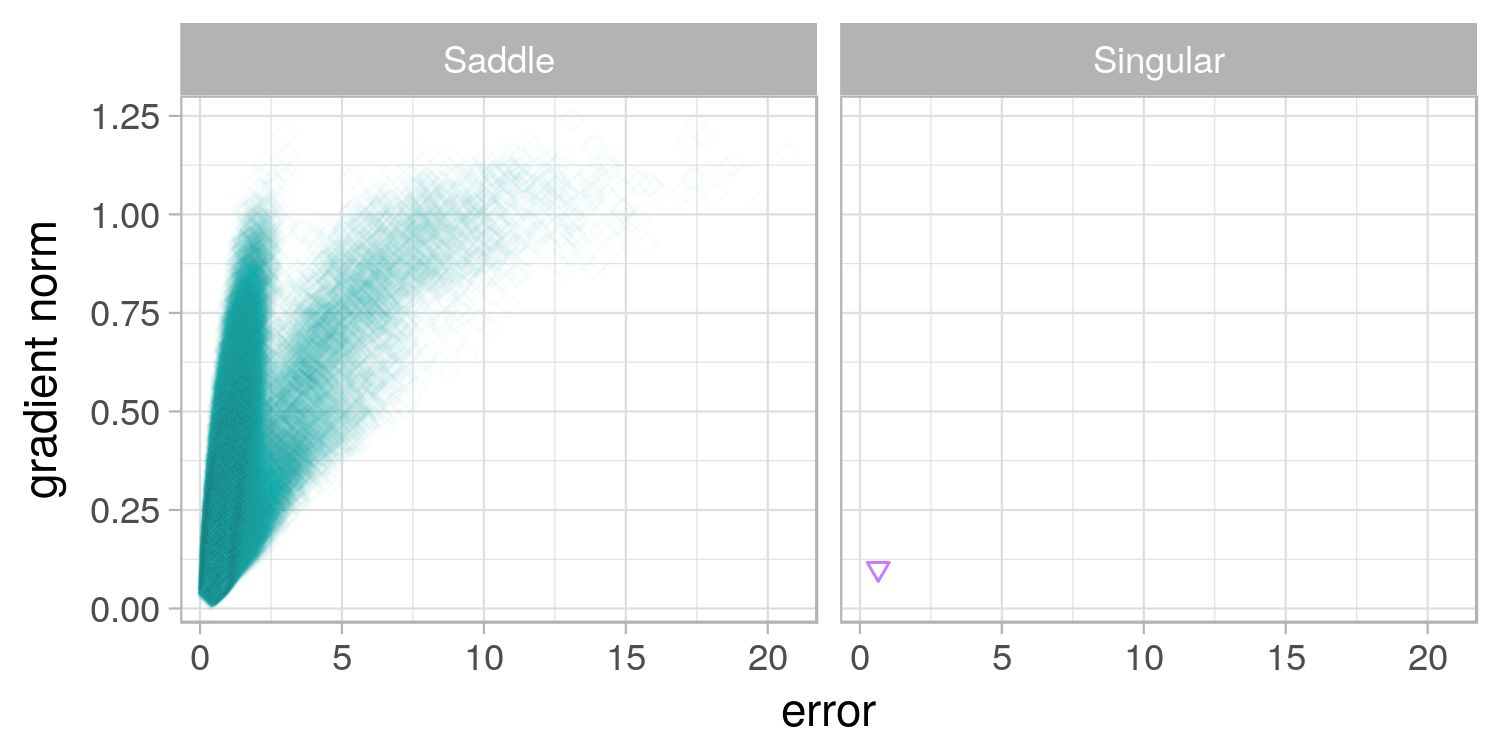}}
  \end{subfloat}
  \caption{L-g clouds for the gradient walks initialised in the $[-1,1]$ range on the Heart
    problem.}\label{fig:heart:b10}
\end{center}
\end{figure}

Figures~\ref{fig:heart:b1} and~\ref{fig:heart:b10} show the l-g clouds obtained for the Heart problem. Similar to the cancer problem, all points sampled by the $[-1,1]$ walks were classified as saddle points. The total dimensionality of the heart problem is 341, which is similar to the dimensionality of the Cancer problem. Figure~\ref{fig:heart:b1} illustrates that both SSE and CE had a single flat attractor in the general area of the global minima. In addition to this attractor, SSE exhibited two more attractors of much higher error. However, the $[-1,1]$ walks did not sample any zero-gradient (stationary) points around the high error attractors.

Larger steps and a larger initialisation range, however, allowed gradient walks to discover the stationary points of high error on the SSE loss surface, as illustrated in Figure~\ref{fig:heart:b10}. The CE loss surface sampled by the same walks did not reveal any additional attractors, but was again visibly split into two clusters leading towards the global minima: points of high gradient and low error, and points of lower gradient and higher error. This is once again indicative of narrow and wide valleys, which appears to be a common characteristic of the CE loss surface.

\begin{figure}
\begin{center}
  \begin{subfloat}[{SSE, micro, $[-1,1]$, $E_t< 0.2$ }\label{fig:heart:gen:micro:mse}]{    \includegraphics[width=0.4\textwidth]{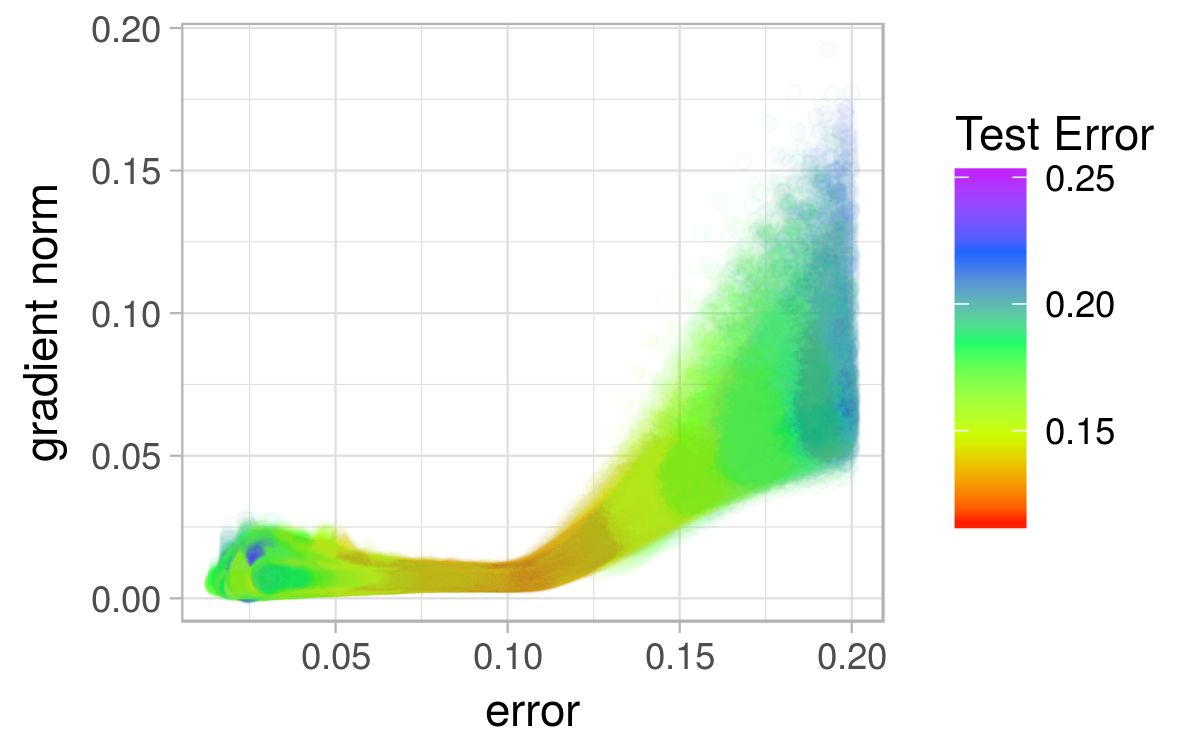}}
  \end{subfloat}
  \begin{subfloat}[{CE, micro, $[-1,1]$, $E_t< 0.5$  }\label{fig:heart:gen:micro:ce}]{    \includegraphics[width=0.4\textwidth]{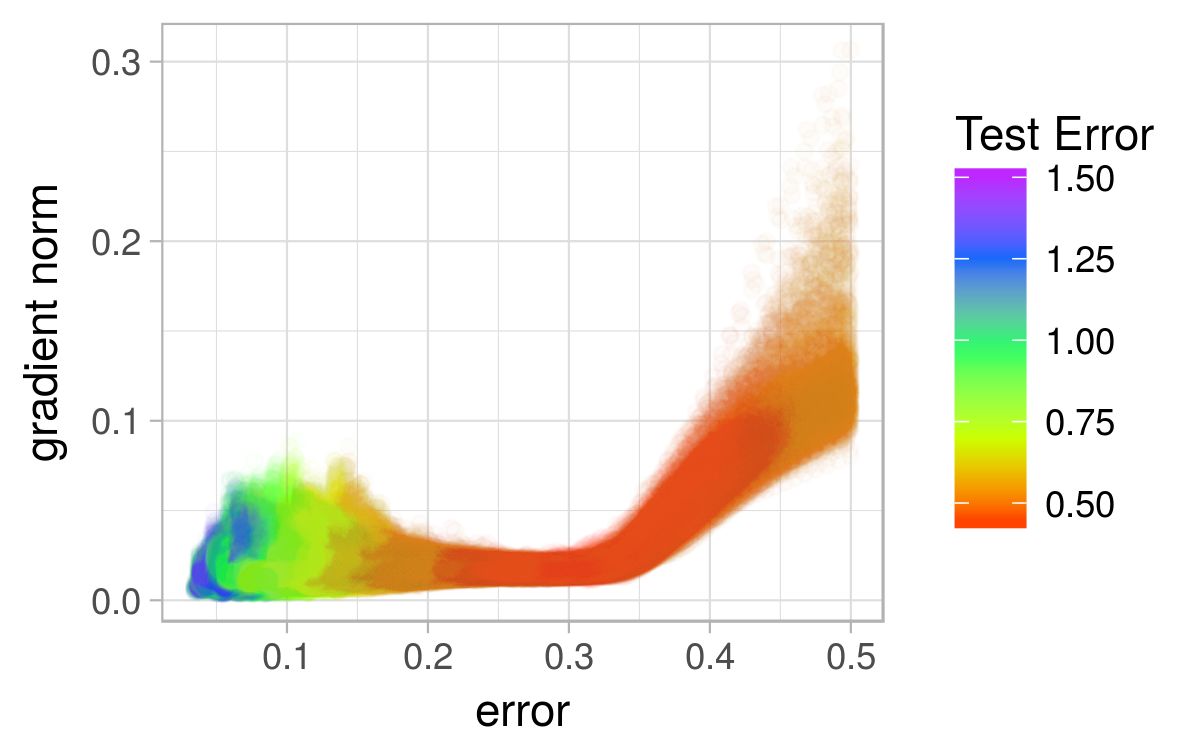}}
  \end{subfloat}
  \caption{L-g clouds colourised according to the corresponding $E_g$ values for the Heart problem.}\label{fig:heart:gen}
\end{center}
\end{figure}

The $n_{stag}$ and $l_{stag}$ values reported in 
Table~\ref{table:heart:lstag} confirm that the walks generally did not make transitions between the discovered attractors. The $n_{stag}$ and $l_{stag}$ values calculated over the $E_g$ values were again less stable than the corresponding $E_t$ values, indicating that exploiting an $E_t$ attractor does not necessarily coincide with exploiting a corresponding $E_g$ attractor. Figure~\ref{fig:heart:gen} illustrates the generalisation behaviour of the flat attractor discovered on both the SSE and CE loss surfaces by the micro $[-1,1]$ walks: the smallest $E_g$ values were observed on the rightmost side of the attractor, closest to the points of higher error and higher gradient. Exploitation around the global minima yielded superior $E_t$ values, but inferior $E_g$ values. This once again illustrates that discovering the global optimum may be unnecessary. The success of techniques such as early stopping~\cite{ref:Morgan:1990} comes precisely from preventing the algorithm from exploiting a global minimum unnecessarily.

\begin{table}
  \caption{Basin of attraction estimates calculated for the Heart problem. Standard deviation shown in parenthesis.}
  \label{table:heart:lstag}
  \begin{center}
    \begin{tabular}{l|cc|cc}
      & \multicolumn{2}{|c|}{SSE} & \multicolumn{2}{|c}{CE} \\
      \hline
          {\bf $E_t$}  &  $n_{stag}$    & $l_{stag}$ & $n_{stag}$ & $l_{stag}$ \\
          \hline          
          $[-1,1]$,  & 1.00000 & 952.36276 & 1.00000 & 947.81432 \\
          micro      & (0.00000) &    (6.49124) &  (0.00000) &    (7.04222) \\
          $[-1,1]$,  & 1.00000 &  86.70645 & 1.00000 &  86.40587 \\
          macro&   (0.00000) &    (0.66921) &  (0.00000) &    (0.97088) \\
          $[-10,10]$,& 1.02493 & 937.01486 & 1.00000 & 966.72036 \\
          micro      &   (0.15962) &   (76.63092) &  (0.00000) &    (3.70200) \\
          $[-10,10]$,&  1.00176 &  84.85293 & 1.00411 &  84.43886 \\
          macro& (0.04191) &    (3.68841) &  (0.06394) &    (7.00978) \\
          \hline\hline
              {\bf $E_g$}  &  $n_{stag}$    & $l_{stag}$ & $n_{stag}$ & $l_{stag}$ \\
              \hline
              $[-1,1]$,  & 1.00733 & 957.87269 & 2.14920 & 710.77930 \\
              micro      & (0.10669) &   (40.88520) &  (2.07157) &  (342.24282) \\
              $[-1,1]$,  &1.00000 &  87.70880 & 1.00088 &  87.54971 \\
              macro&  (0.00000) &    (0.60395) &  (0.02965) &    (1.97717) \\
              $[-10,10]$,& 1.54927 & 821.66135 & 1.00298 & 965.68084 \\
              micro      & (1.78543) &  (251.33524) &  (0.05453) &   (27.96970) \\
              $[-10,10]$,&    1.00440 &  84.55381 & 1.04956 &  79.23624 \\
              macro& (0.06618) &    (5.61240) &  (0.24735) &   (17.04847) \\
              \hline
    \end{tabular}
  \end{center}
\end{table}

\subsection{MNIST}
\begin{figure}
\begin{center}
  \begin{subfloat}[{SSE, micro, $[-1,1]$ }\label{fig:mnist:mse:micro:b1}]{    \includegraphics[width=0.4\textwidth]{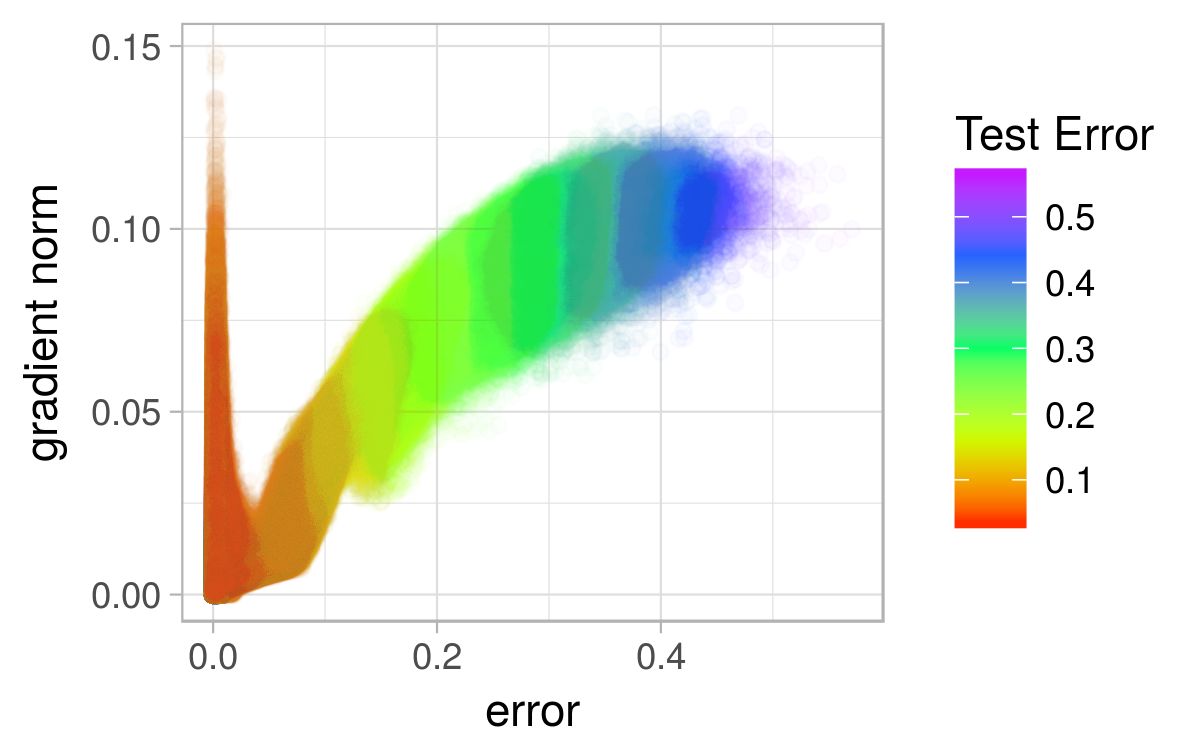}}
  \end{subfloat}
  \begin{subfloat}[{CE, micro, $[-1,1]$ }]{    \includegraphics[width=0.4\textwidth]{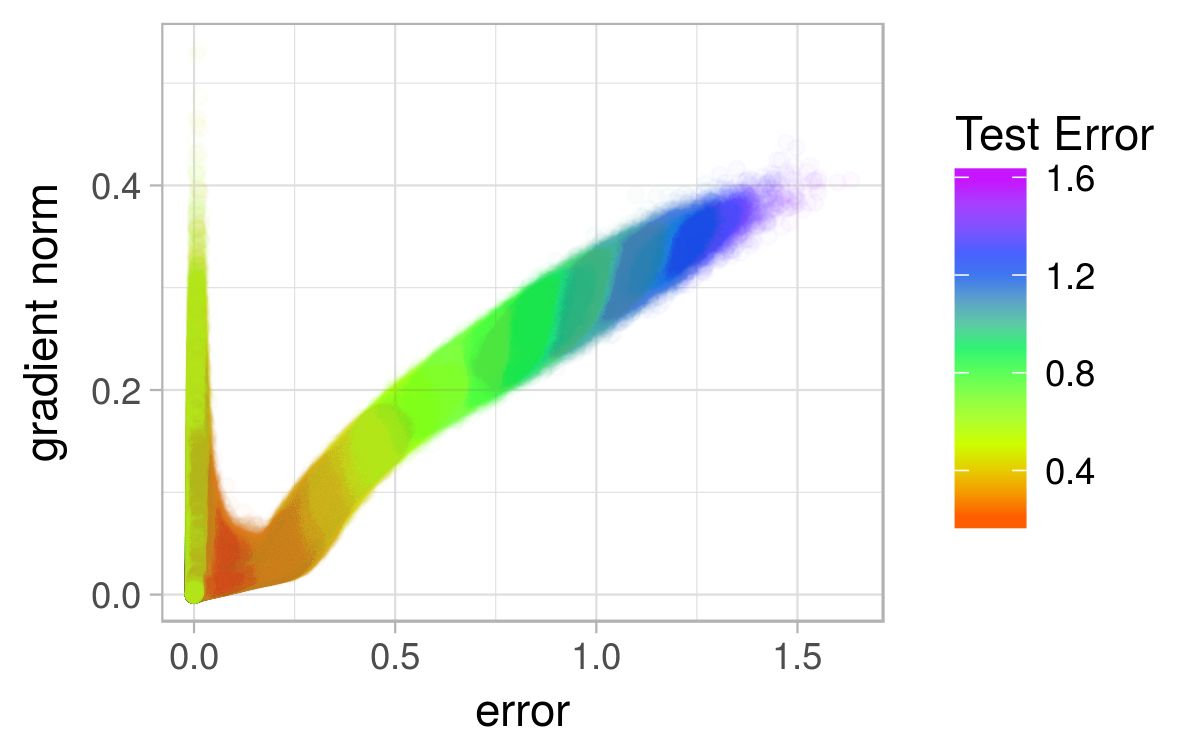}}
  \end{subfloat}\\
  \begin{subfloat}[{SSE, macro, $[-1,1]$ }\label{fig:mnist:mse:macro:b1}]{    \includegraphics[width=0.4\textwidth]{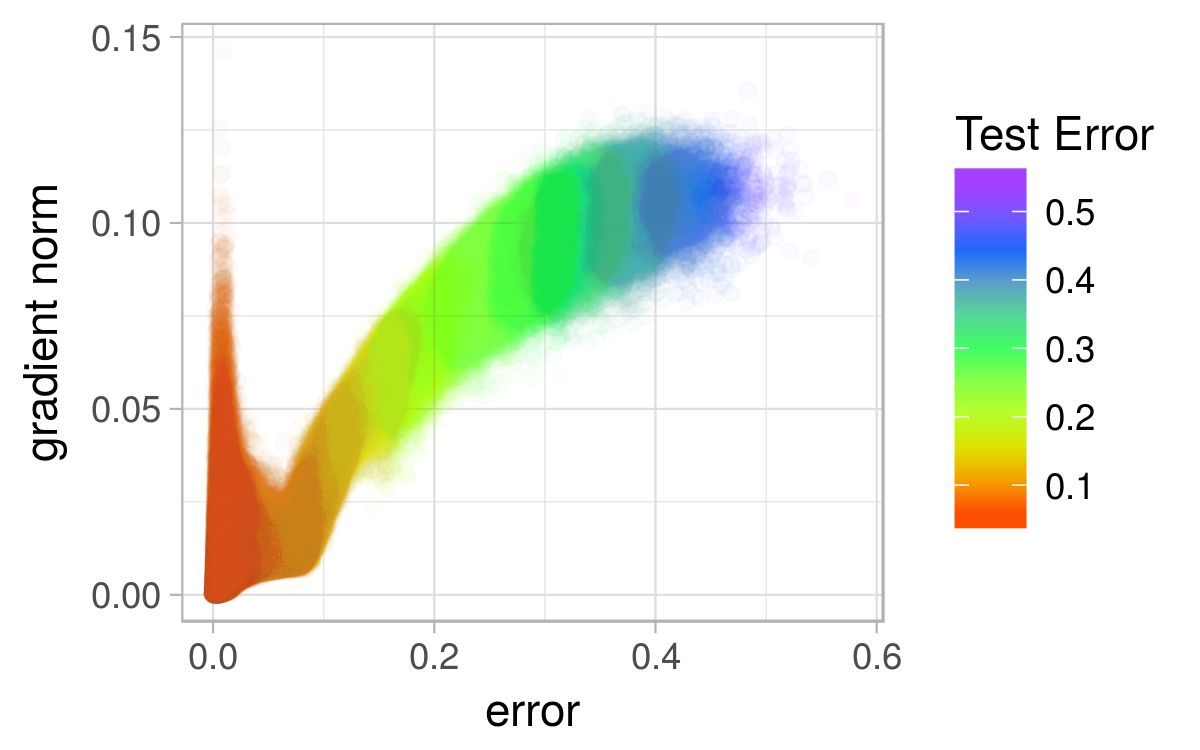}}
  \end{subfloat}
  \begin{subfloat}[{CE, macro, $[-1,1]$ }]{    \includegraphics[width=0.4\textwidth]{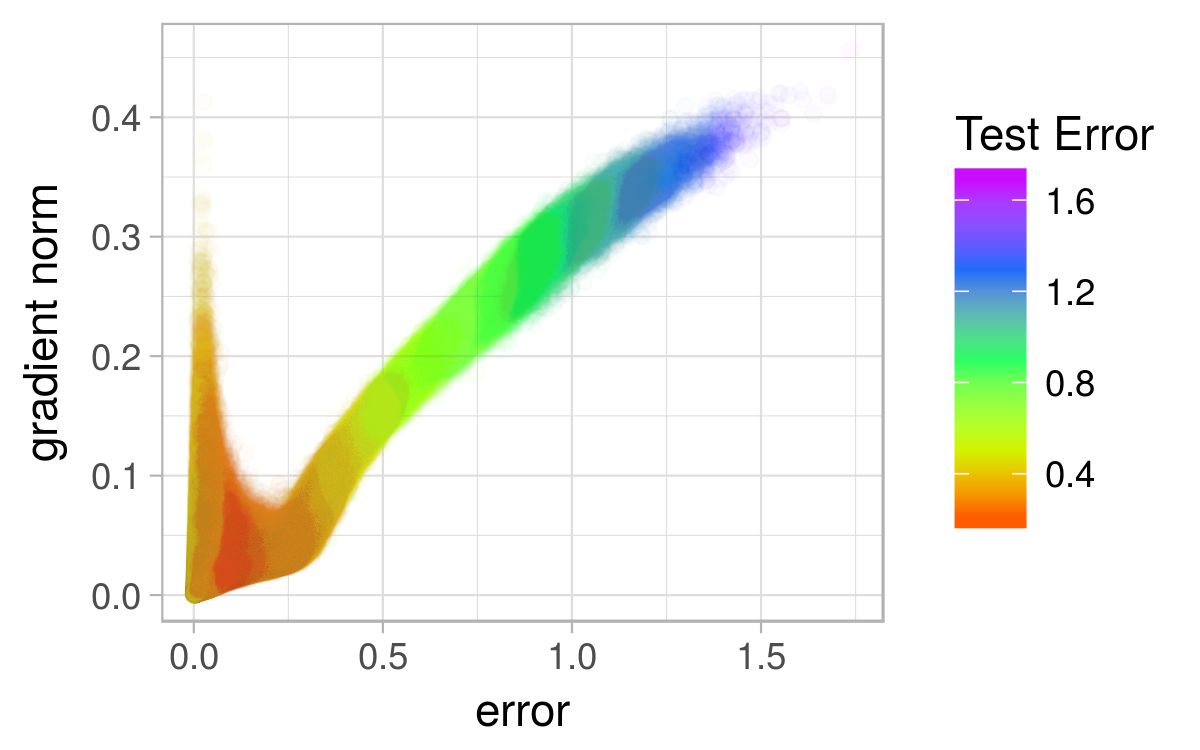}}
  \end{subfloat}
  \caption{L-g clouds for the gradient walks initialised in the $[-1,1]$ range on the MNIST
    problem.}\label{fig:mnist:b1}
\end{center}
\end{figure}

\begin{figure}
\begin{center}
  \begin{subfloat}[{SSE, micro, $[-10,10]$ }\label{fig:mnist:mse:micro:b10}]{    \includegraphics[width=0.4\textwidth]{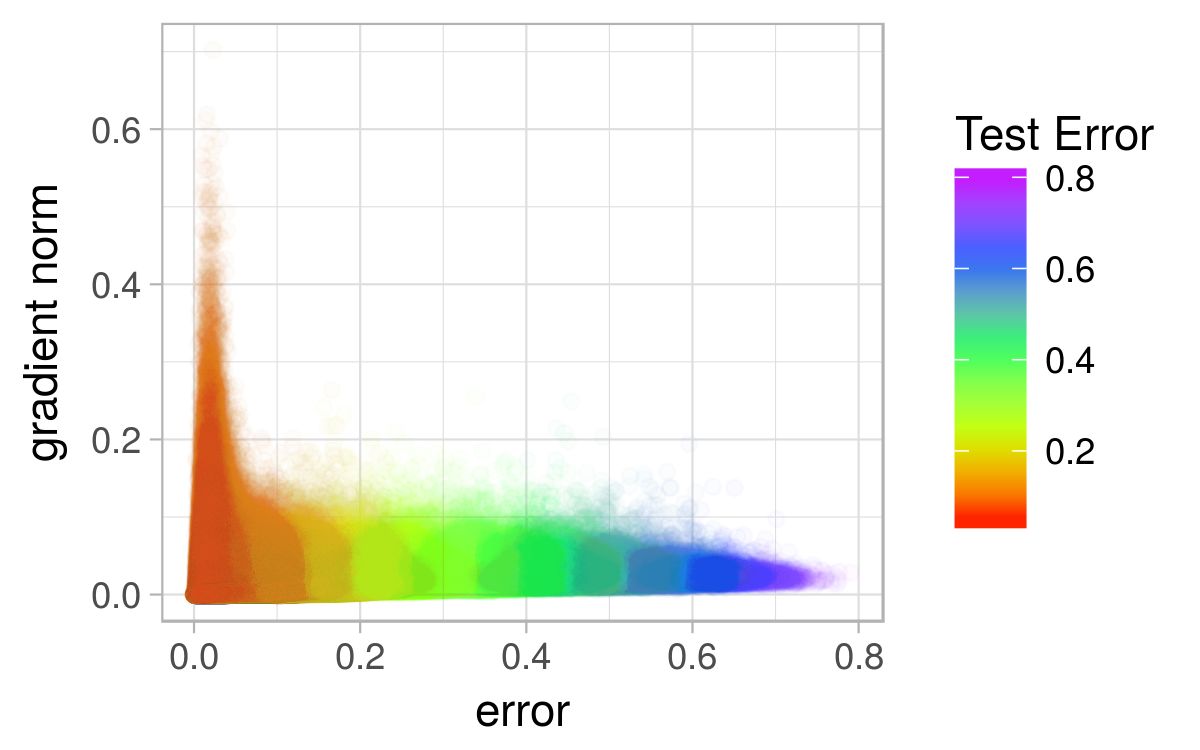}}
  \end{subfloat}
  \begin{subfloat}[{CE, micro, $[-10,10]$ }]{    \includegraphics[width=0.4\textwidth]{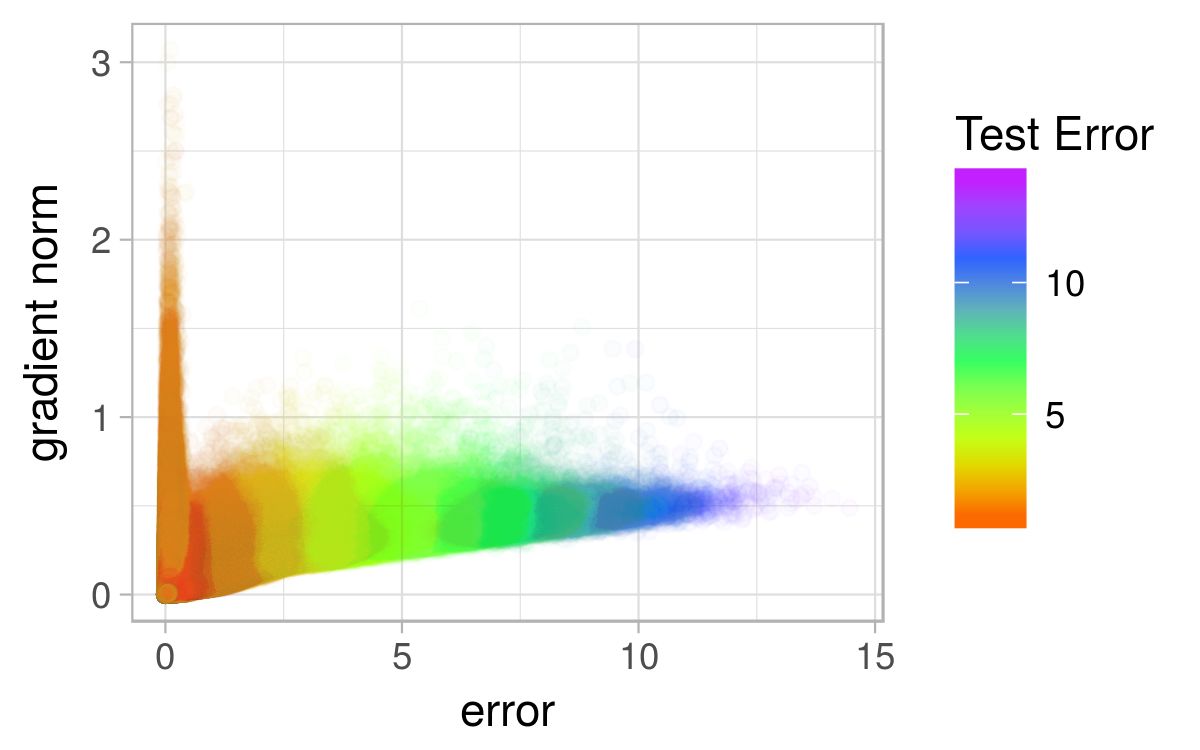}}
  \end{subfloat}\\
  \begin{subfloat}[{SSE, macro, $[-10,10]$ }\label{fig:mnist:mse:macro:b10}]{    \includegraphics[width=0.4\textwidth]{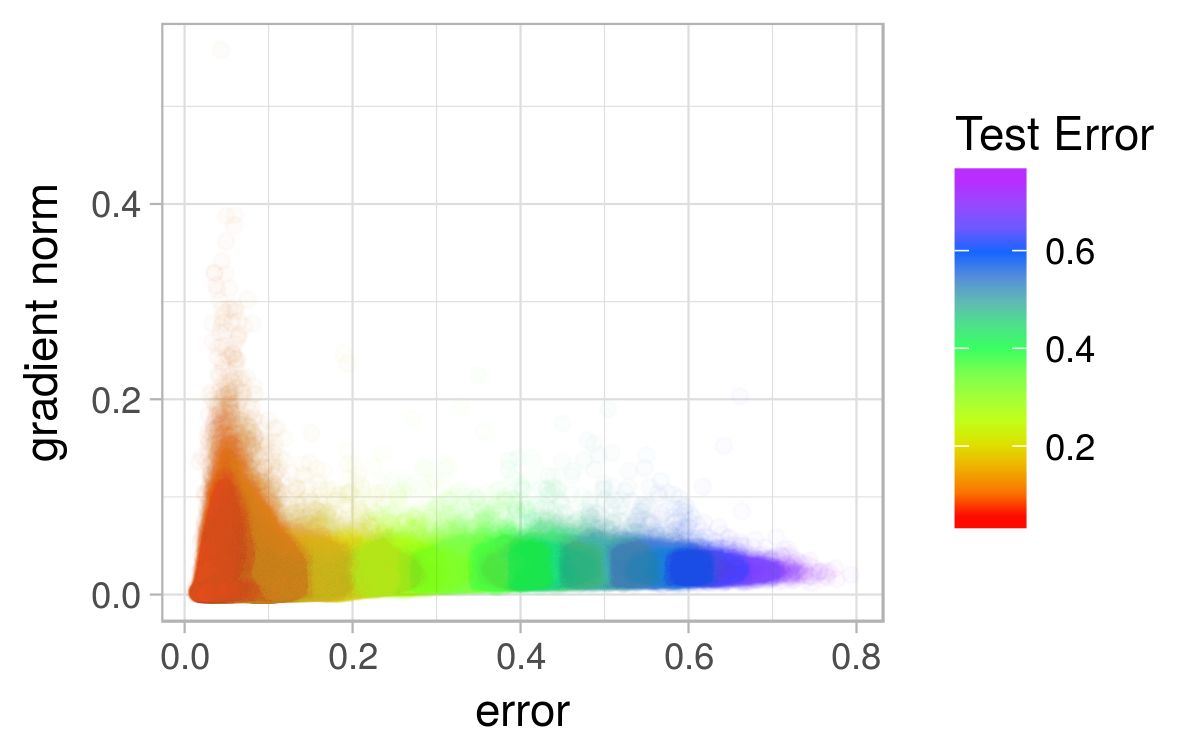}}
  \end{subfloat}
  \begin{subfloat}[{CE, macro, $[-10,10]$ }]{    \includegraphics[width=0.4\textwidth]{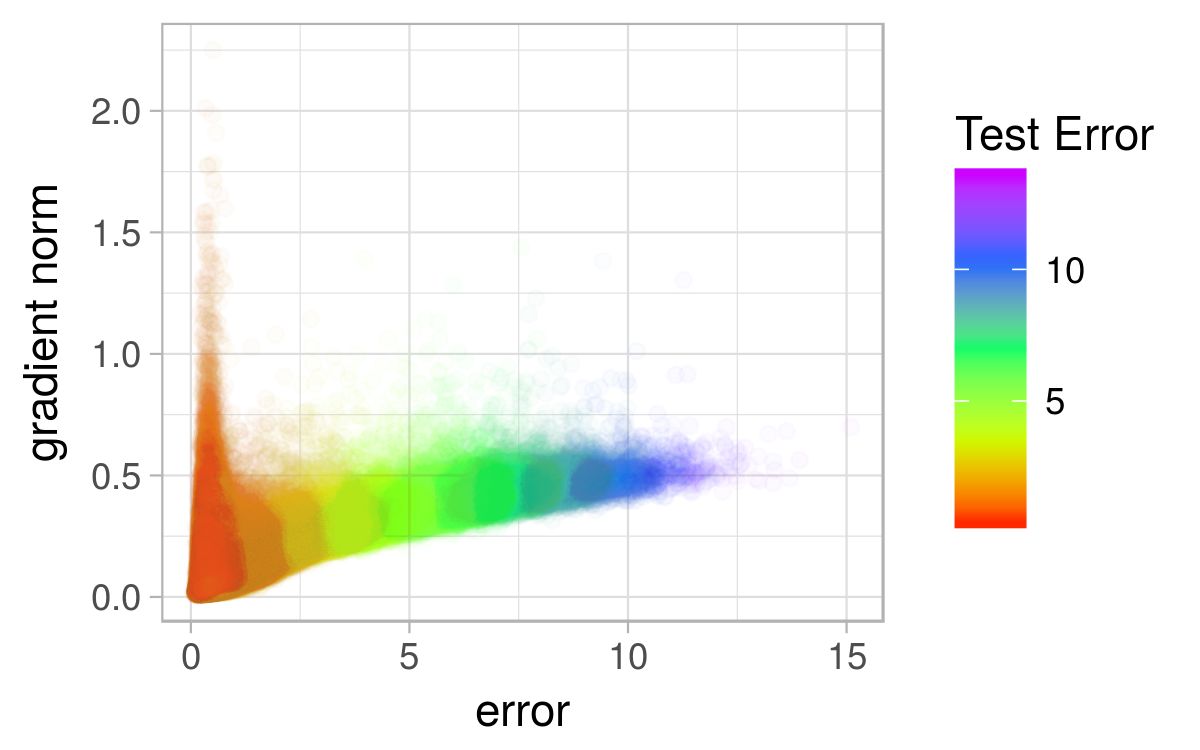}}
  \end{subfloat}
  \caption{L-g clouds for the gradient walks initialised in the $[-10,10]$ range on the MNIST
    problem.}\label{fig:mnist:b10}
\end{center}
\end{figure}

Figures~\ref{fig:mnist:b1} and~\ref{fig:mnist:b10} show the l-g clouds for the MNIST problem. Due to the prohibitively expensive memory requirements, the Hessian matrices were not computed for the MNIST dataset. Thus, the curvature of the loss functions for the MNIST dataset is not reported in this study. The reader is referred to the previous studies of the MNIST Hessians~\cite{ref:Sagun:2017} for a discussion of curvature characteristics, where it was shown that the gradient descent algorithm  discovered points of saddle and singular curvature only.

Figure~\ref{fig:mnist:b1} shows that both SSE and CE exhibited one major attractor around the global minima. Additionally, SSE exhibited two more attractors of non-zero gradient. Thus, the error landscape of CE was more searchable than the error landscape of SSE. The $n_{stag}$ and $l_{stag}$ results reported in Table~\ref{table:mnist:lstag} indicate that most walks have discovered a single attractor only, which corresponds to the results in Figures~\ref{fig:mnist:b1} and~\ref{fig:mnist:b10}.

\begin{figure}
\begin{center}
  \begin{subfloat}[{SSE, macro, $[-10,10]$, $E_t< 0.2$ }\label{fig:mnist:gen:macro:mse}]{    \includegraphics[width=0.4\textwidth]{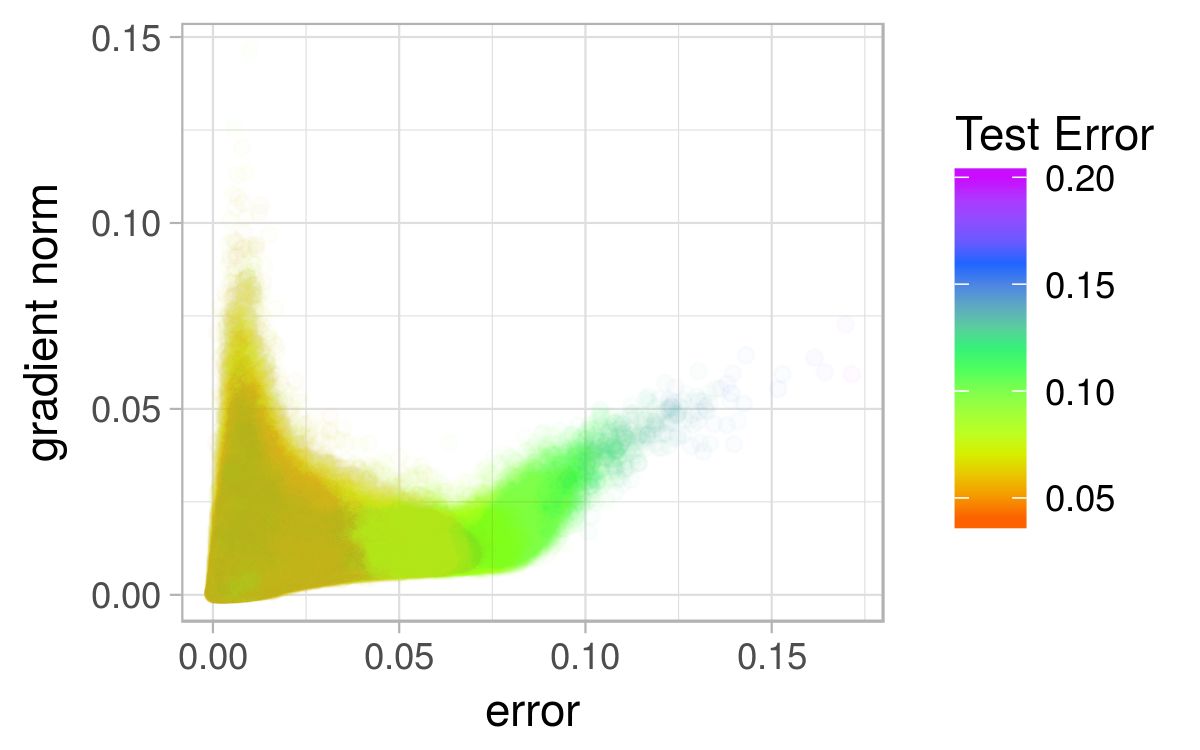}}
  \end{subfloat}
  \begin{subfloat}[{CE, macro, $[-10,10]$, $E_t< 0.25$  }\label{fig:mnist:gen:macro:ce}]{    \includegraphics[width=0.4\textwidth]{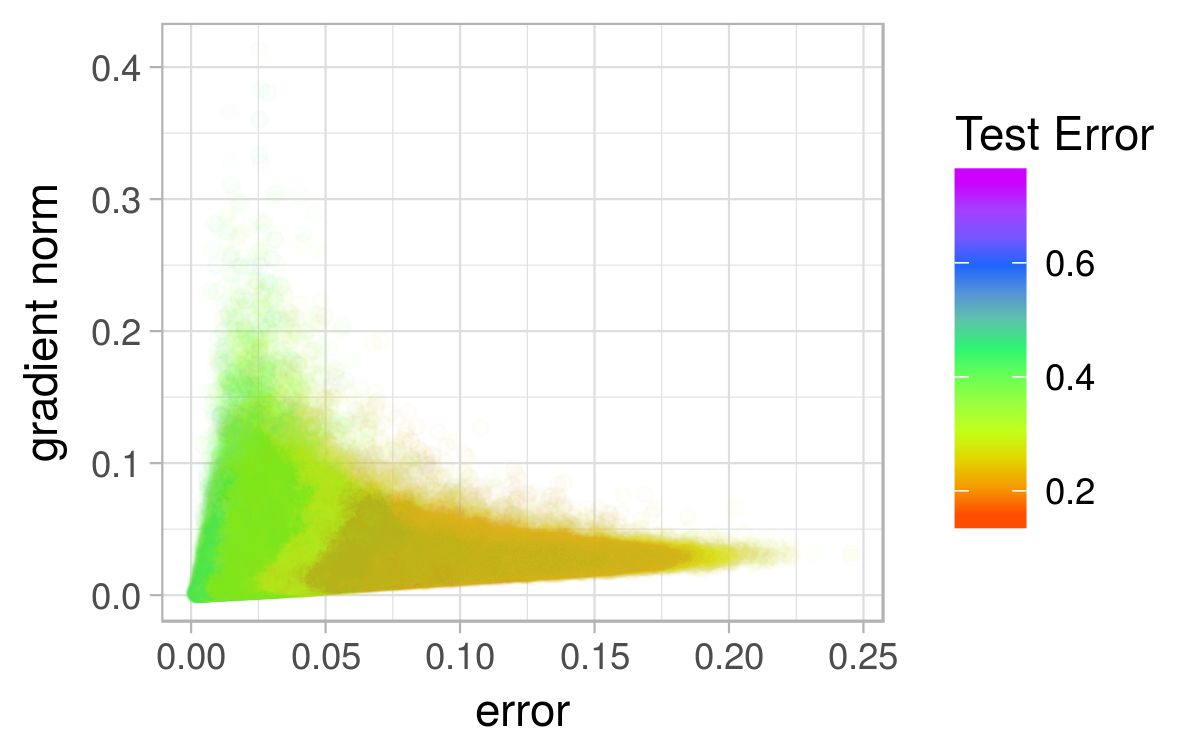}}
  \end{subfloat}
  \caption{L-g clouds colourised according to the corresponding $E_g$ values for
    the MNIST problem.}\label{fig:mnist:gen}
\end{center}
\end{figure}

A cluster of values of high gradient and low error can be observed for both SSE and CE, indicating that both exhibited sharp minima. SSE, however, exhibited lower gradients overall. Figure~\ref{fig:mnist:b1} illustrates that the generalisation performance improved as the error approached the global minima. Figure~\ref{fig:mnist:gen} shows the generalisation performance of the points sampled around the global minima. SSE once again exhibited better generalisation performance around the global minima than CE, confirming the earlier made hypothesis that SSE is less prone to overfitting due to weaker  gradients. The classification error results reported in Table~\ref{table:mnist:class}, however, indicate that, although SSE yielded a smaller disparity between the $E_t$ and $E_g$ values, both loss functions performed similarly in terms of final classification.

\begin{table}
  \caption{Basin of attraction estimates calculated for the MNIST problem. Standard deviation shown in parenthesis.}
  \label{table:mnist:lstag}
  \begin{center}
    \begin{tabular}{l|cc|cc}
      & \multicolumn{2}{|c|}{SSE} & \multicolumn{2}{|c}{CE} \\
      \hline
          {\bf $E_t$}  &  $n_{stag}$    & $l_{stag}$ & $n_{stag}$ & $l_{stag}$ \\
          \hline          
          $[-1,1]$,  & 1.00003 & 948.96269 & 1.00020 & 943.65445 \\
          micro      & (0.00557)&(7.73257)&(0.01418)&(10.31210)\\
          $[-1,1]$,  & 1.00000 & 89.59761 & 1.00000 & 88.94583\\
          macro&   (0.00000)&(0.62686)&(0.00000)&(0.79698)\\
          $[-10,10]$,& 1.00338 & 944.23606 & 1.00020 & 955.48716\\
          micro      & (0.06420)&(29.25325)&(0.01418)&(9.14026)\\
          $[-10,10]$,& 1.00004 & 90.19884 & 1.00001 & 90.24536\\
          macro& (0.00614)&(1.02171)&(0.00354)&(1.09101) \\
          \hline\hline
              {\bf $E_g$}  &  $n_{stag}$    & $l_{stag}$ & $n_{stag}$ & $l_{stag}$ \\
              \hline
              $[-1,1]$,  & 1.00028 & 944.25561 & 2.84430 & 570.85913 \\
              micro      & (0.01762)&(10.01749)&(2.71734)&(334.02547) \\
              $[-1,1]$,  &1.00000 & 90.04197 & 1.01201 & 85.19988 \\
              macro&  (0.00000)&(0.63536)&(0.11234)&(7.54223)\\
              $[-10,10]$,& 1.00408 & 943.53542 & 1.26670 & 878.55561 \\
              micro      & (0.11775)&(29.40333)&(1.10976)&(191.85100) \\
              $[-10,10]$,& 1.00005 & 90.31260 & 1.00881 & 88.80587 \\
              macro& (0.00709)&(1.00828)&(0.10216)&(6.14361)\\
              \hline
    \end{tabular}
  \end{center}
\end{table}

Figure~\ref{fig:mnist:b10} indicates that SSE exhibited a much weaker correlation between the gradient and the error when sampled by gradient walks initialised in the $[-10,10]$ interval. For CE, the positive correlation was still clearly manifested. Thus, CE exhibited a more searchable landscape when sampled by the $[-10,10]$ walks. Both loss functions clustered along the error and the gradient axis.

The landscape properties exhibited by the MNIST problem were thus very similar to the landscape properties exhibited by the problems of lower dimensionality. The CE loss surface was more searchable for all problems considered, and exhibited fewer non-global attractors. SSE, however, exhibited somewhat better generalisation capabilities under some of the considered scenarios. Perhaps the two loss functions should be combined to construct an error landscape that is both searchable and robust to overfitting.

\section{Conclusions}\label{sec:conclusions}
This study presented a visual and numerical analysis of local minima and the associated basins of attraction for two common NN loss functions, i.e. quadratic loss and entropic loss. The study was performed by analysing the samples obtained by a number of progressive gradient walks proportionate to the dimensionality of the problems. The gradient walks were not restricted to any specific search space bounds, but were initialised in two distinct intervals, i.e. $[-1,1]$ and $[-10,10]$. Additionally, two granularity settings were considered for the gradient walks, namely micro and macro.

This study proposed an intuitive visualisation of the local minima and the associated basins of attraction, namely the loss-gradient clouds. By plotting the sampled loss values against the corresponding gradient vector magnitudes, stationary points could be easily identified. To classify the identified stationary points as minima, maxima, or saddles, Hessian matrix information were used to identify the curvature of each sampled point.

Additionally, this study proposed two simple metrics to quantify the number and extent of attraction basins as sampled by the walks. Calculation of statistical metrics over a number of walks provides an idea of the connectedness of the various basins, as well as the likelihood of escaping from the basins.

Both loss functions exhibited convex local minima for the XOR problem. The amount of observed convexity decreased with the increase in problem dimensionality. Saddle curvature was the most prevalent curvature observed, and some higher-dimensional problems considered exhibited only saddle curvature for all sampled points.

SSE consistently exhibited more local stationary points and associated attractors than CE. Analysis of the individual walks further revealed that transition between different attractors was unlikely, and that the paths connecting different attractors exhibited singular Hessian matrices, indicative of flatness. Thus, CE exhibited a more consistent and searchable structure across the selection of problems considered in this study.

With an increase in problem dimensionality, the number of zero or low gradient attractors decreased. The majority of the problems exhibited a single main attractor around the global optimum. For CE, the gradient was for the most part positively correlated to the error value, indicating that the CE loss surface is highly searchable from the perspective of gradient-based methods. This study did not attempt to quantify the number of optima, but the obtained results clearly indicated that the majority of the optima exhibited similar loss values.

The results confirmed previously made observations of the presence of valley-shaped optima in NN error landscapes. For the majority of the problems, descending into a valley was easily accomplished by the walks. Travelling down the bottom of the valley towards the global minimum yielded a decrease in generalisation performance for both SSE and CE. CE exhibited stronger gradients than SSE in all experiments conducted, and the stronger gradients promoted overfitting in CE. For some of the problems, SSE exhibited better generalisation performance. It can be speculated that the CE loss surface is more prone to sharp minima (narrow valleys) than SSE; thus, CE is more easily overfitted. The experiments revealed the tendency for the points sampled on CE to fall into two major clusters: points of low error and high gradients, and points of higher error and low gradients. These are hypothesised to represent 
narrow and wide valleys, respectively. The results of this study confirmed that superior generalisation performance was exhibited by the points in the wide valleys.

An analysis of the progressive gradient samples thus illustrated a number of current theories regarding the shape of NN error surfaces, and highlighted the differences between SSE and CE loss surfaces, confirming that FLA is a viable method for visualisation and analysis of NN error landscapes. Future research will apply FLA to analyse the influence of various activation functions, as well as NN architectures, on the resulting stationary points and attraction basins.

The observation that thee SSE landscape may have superior generalisation properties suggests that a hybrid of SSE and CE may produce a landscape that combines the searchability of CE with the robustness of SSE. Additionally, the presence of a single attractor in the majority of the problems considered suggests that an exploitative rather than an exploratory approach should be taken for the purpose of NN training. This observation has strong implications for population-based training algorithms, which so far failed to be effectively applied to high-dimensioal NN training problems. A population-based approach designed with exploitation rather than exploration in mind may perform competitively, especially if gradient information is used as one of the guides for the population. This hypothesis is further supported by a recent study of particle swarm optimisation in high-dimensional spaces~\cite{oldewage2018perils}, were the efficacy of exploitation over exploration in high-dimensional spaces was observed. Investigation of exploitative population-based techniques applied to NNs is an interesting topic for future research.

Another interesting observation is the impressive ability of a randomised algorithm to find the global optima, when guided by nothing besides the direction of the gradient. As \ref{appendix:class} indicates, the average classification error calculated at the last step of the gradient walks approached 100\% accuracy on most problems under at least one of the granularity settings. Perhaps gradient-guided stochastic training algorithms should be considered for deeper, more complex problems.

\section*{Acknowledgements}
The authors would like to thank the Centre for High Performance
Computing (CHPC) (http://www.chpc.ac.za) for the use of their cluster to
obtain the data for this study. 

\bibliography{dissertation}

\appendix
\section{Pseudocode for basin of attraction estimates}\label{appendix:lstag}\noindent Two estimates to quantify the basins of attraction are proposed in
this study:
\begin{enumerate}
\item The average number of times stagnation observed, $n_{stag}$.
\item The average length of the stagnant sequence, $l_{stag}$.
\end{enumerate}
Pseudocode given in Algorithms~\ref{algo:stationary} and~\ref{algo:compute_stationary} summarises the
proposed method to obtain both metrics.

\begin{algorithm}[!htb]
	\caption{Basins of attraction estimates}
	\label{algo:stationary}
	\begin{algorithmic}
		\STATE Initialise $n_{stag}$, average number of basins, to 0;
		\STATE Initialise $l_{stag}$, average basin size, to 0;
		\STATE Initialise $n_w$ to the number of walks to perform;
		\STATE Initialise \textit{walk}, the sample, to $\emptyset$;
		\FOR{$\forall i \in \{1,...,n_w\}$}
			\STATE \textit{walk} $\gets$ sample the input problem using a progressive gradient walk~\cite{ref:Bosman:2018};
			\STATE Normalise the sample fitness range in \textit{walk} to $[0,1]$;
			\STATE Initialise $n_{stag, i}$ and $l_{stag, i}$ to 0 for walk $i$;                        
                        \FOR{$\forall j \in \{6,8,...,18,20\}$}
                                \STATE \textit{walk} $\gets$ calculate the exponentially weighted moving average of \textit{walk}, $\alpha = 2/(j + 1)$
                                \STATE $epsilon \gets$ calculate the standard deviation of \textit{walk}
                                \STATE \textit{stdev\_walk} $\gets$ calculate the sequence of moving standard deviations of \textit{walk}, $w = j$
                                \STATE Obtain a list of stagnant regions, $l$, using algorithm \ref{algo:compute_stationary} with inputs \textit{stdev\_walk}, $epsilon$.
                                \IF{average length of regions in $l > l_{stag, i}$}
                                        \STATE $l_{stag, i} \gets$ average($l$)
                                        \STATE $n_{stag, i} \gets$ number of regions in $l$
			        \ENDIF
                        \ENDFOR
                        \STATE $n_{stag} \gets n_{stag} + n_{stag, i}$
                        \STATE $l_{stag} \gets l_{stag} + l_{stag, i}$
		\ENDFOR
		\STATE return $n_{stag}/n_w$, $l_{stag}/n_w$
	\end{algorithmic}
\end{algorithm}

\begin{algorithm}[!htb]
	\caption{Basins of attraction identification}
	\label{algo:compute_stationary}
	\begin{algorithmic}
		\STATE Inputs: \textit{stdev\_walk}, $epsilon$;
		\STATE Initialise $l_{stag}$, average basin size, to 0;
		\STATE Initialise \textit{stuck} to \FALSE
		\STATE Initialise $len$, length of a stagnant region, to 0;
		\STATE Initialise $l$, the list of stagnant regions, to $\emptyset$;
		\FOR{each step $s_i$ in \textit{stdev\_walk}}
			\IF{\textit{stuck}}
			        \IF{$s_i < \epsilon$}
                                \STATE $len \gets len + 1$
                                \ELSE
                                \STATE $stuck \gets$\FALSE
                                \STATE $l \gets$ add $len$ to the list $l$
                                \STATE $len \gets 0$
                                \ENDIF
                        \ELSE
                            \IF{$s_i < \epsilon$}   
                            \STATE $len \gets len + 1$
                            \STATE $stuck \gets$\TRUE
                            \ENDIF 
			\ENDIF
		\ENDFOR
                \IF{$len > 0$}     
                \STATE $l \gets$ add $len$ to the list $l$
		\ENDIF
		\STATE return $l$
	\end{algorithmic}
\end{algorithm}

\section{Classification errors}\label{appendix:class}\noindent The average classification errors arrived at by the gradient walks are reported in this appendix. Averages are calculated across the error values as observed at the last step of each walk. The classification error of the training set is referred to as $C_t$, and the classification error of the test set is referred to as $C_g$. Tables~\ref{table:iris:class},
\ref{table:diab:class}, \ref{table:glass:class},
\ref{table:cancer:class}, \ref{table:heart:class},
and~\ref{table:mnist:class} list the average $C_t$ and $C_g$ values obtained for the iris, diabetes, glass, cancer, and MNIST problems, respectively. Standard deviation is shown in parenthesis.

\begin{table}
  \small\addtolength{\tabcolsep}{-5pt}
  \caption{Iris, classification errors.}\label{table:iris:class}
  \begin{center}
  \begin{tabular}{l|cc|cc|cc|cc}
    \hline & \multicolumn{4}{|c|}{micro} & \multicolumn{4}{c}{macro}\\\hline
    & \multicolumn{2}{|c|}{SSE} &  \multicolumn{2}{|c|}{CE} & \multicolumn{2}{|c|}{SSE} & \multicolumn{2}{|c}{CE} \\\hline
    &$C_t$ & $C_g$ & $C_t$ & $C_g$ & $C_t$ & $C_g$ & $C_t$ & $C_g$  \\ \hline
    $[-1,1]$& 1.00000 & 0.91819 & 0.98352 & 0.93333 & 0.96638 & 1.00000 & 0.97557 & 0.96667 \\
    & (0.00000) & (0.01660) & (0.00125) & (0.00000) &  (0.00881) & (0.00000) & (0.00646) & (0.00000) \\
    $[-10,10]$& 0.97252 & 0.97105 & 0.99245 & 0.90581 & 0.92155 & 0.92829 & 0.92857 & 0.92457 \\
    &(0.07622) & (0.08790) & (0.00734) & (0.05097) & (0.09578) & (0.10521) & (0.05844) & (0.05806) \\
    \hline
  \end{tabular}
  \end{center}
\end{table}


\begin{table}
  \small\addtolength{\tabcolsep}{-5pt}
  \caption{Diabetes, classification errors.}\label{table:diab:class}
  \begin{center}
  \begin{tabular}{l|cc|cc|cc|cc}
    \hline & \multicolumn{4}{|c|}{micro} & \multicolumn{4}{c}{macro}\\\hline
    & \multicolumn{2}{|c|}{SSE} &  \multicolumn{2}{|c|}{CE} & \multicolumn{2}{|c|}{SSE} & \multicolumn{2}{|c}{CE} \\\hline
    &$C_t$ & $C_g$ & $C_t$ & $C_g$ & $C_t$ & $C_g$ & $C_t$ & $C_g$  \\ \hline
    $[-1,1]$& 0.91094 & 0.66913 & 0.85453 & 0.73725 & 0.81141 & 0.73586 & 0.81187 & 0.74165 \\
    & (0.01002) & (0.02400) & (0.00991) & (0.02684) & (0.00959) & (0.01712) & (0.01017) & (0.02307) \\
    $[-10,10]$& 0.85434 & 0.74521 & 0.83494 & 0.69911 & 0.79669 & 0.68657 & 0.71915 & 0.66424 \\
    & (0.01441) & (0.02648) & (0.01480) & (0.03019) & (0.02970) & (0.03580) & (0.06485) & (0.05338) \\
    \hline
  \end{tabular}
  \end{center}
\end{table}

\begin{table}
  \small\addtolength{\tabcolsep}{-5pt}
  \caption{Glass, classification errors.}\label{table:glass:class}
  \begin{center}
  \begin{tabular}{l|cc|cc|cc|cc}
    \hline & \multicolumn{4}{|c|}{micro} & \multicolumn{4}{c}{macro}\\\hline
    & \multicolumn{2}{|c|}{SSE} &  \multicolumn{2}{|c|}{CE} & \multicolumn{2}{|c|}{SSE} & \multicolumn{2}{|c}{CE} \\\hline
    &$C_t$ & $C_g$ & $C_t$ & $C_g$ & $C_t$ & $C_g$ & $C_t$ & $C_g$  \\ \hline
    $[-1,1]$& 0.94400 & 0.55738 & 0.93769 & 0.62740 & 0.79600 & 0.68398 & 0.79793 & 0.67744 \\
    &(0.01636)&(0.04912)&(0.01551)&(0.04766)&(0.02738)&(0.04128)&(0.02285)&(0.05012) \\
    $[-10,10]$& 0.79578 & 0.60513 & 0.90388 & 0.62657 & 0.71373 & 0.58626 & 0.69585 & 0.55828 \\
	&(0.08762)&(0.08170)&(0.02785)&(0.05711)&(0.06541)&(0.06897)&(0.07784)&(0.08112) \\
    \hline
  \end{tabular}
  \end{center}
\end{table}


\begin{table}
  \small\addtolength{\tabcolsep}{-5pt}
  \caption{Cancer, classification errors.}\label{table:cancer:class}
  \begin{center}
  \begin{tabular}{l|cc|cc|cc|cc}
    \hline & \multicolumn{4}{|c|}{micro} & \multicolumn{4}{c}{macro}\\\hline
    & \multicolumn{2}{|c|}{SSE} &  \multicolumn{2}{|c|}{CE} & \multicolumn{2}{|c|}{SSE} & \multicolumn{2}{|c}{CE} \\\hline
    &$C_t$ & $C_g$ & $C_t$ & $C_g$ & $C_t$ & $C_g$ & $C_t$ & $C_g$  \\ \hline
    $[-1,1]$& 0.99944 & 0.97298 & 1.00000 & 0.97322 & 0.99451 & 0.97656 & 0.99633 & 0.97487 \\
    &  (0.00096)&(0.00788)&(0.00000)&(0.00861)&(0.00227)&(0.00759)&(0.00275)&(0.00887) \\
    $[-10,10]$& 0.99813 & 0.96206 & 1.00000 & 0.96408 &  0.99539 & 0.96574 & 0.99357 & 0.97335 \\
   &(0.00150)&(0.01170)&(0.00000)&(0.00685)&(0.00279)&(0.01006)&(0.00612)&(0.00961) \\
    \hline
  \end{tabular}
  \end{center}
\end{table}

\begin{table}
  \small\addtolength{\tabcolsep}{-5pt}
  \caption{Heart, classification errors.}\label{table:heart:class}
  \begin{center}
  \begin{tabular}{l|cc|cc|cc|cc}
    \hline & \multicolumn{4}{|c|}{micro} & \multicolumn{4}{c}{macro}\\\hline
    & \multicolumn{2}{|c|}{SSE} &  \multicolumn{2}{|c|}{CE} & \multicolumn{2}{|c|}{SSE} & \multicolumn{2}{|c}{CE} \\\hline
    &$C_t$ & $C_g$ & $C_t$ & $C_g$ & $C_t$ & $C_g$ & $C_t$ & $C_g$  \\ \hline
    $[-1,1]$& 0.97447 & 0.78274 & 0.97918 & 0.77466 & 0.91038 & 0.83086 & 0.90601 & 0.82772 \\
    & (0.00477)&(0.02250)&(0.00648)&(0.02138)&( 0.00925)&(0.01538)&(0.00915)&(0.01743) \\
    $[-10,10]$&  0.95409 & 0.76148 & 0.93496 & 0.80585 &  0.85821 & 0.83363 & 0.80135 & 0.74857 \\
    & (0.00910)&(0.02149)&(0.01096)&(0.02425)&(0.01829)&(0.02063)&(0.05700)&(0.05340) \\
    \hline
  \end{tabular}
  \end{center}
\end{table}

\begin{table}
  \small\addtolength{\tabcolsep}{-5pt}
  \caption{MNIST, classification errors.}\label{table:mnist:class}
  \begin{center}
  \begin{tabular}{l|cc|cc|cc|cc}
    \hline & \multicolumn{4}{|c|}{micro} & \multicolumn{4}{c}{macro}\\\hline
    & \multicolumn{2}{|c|}{SSE} &  \multicolumn{2}{|c|}{CE} & \multicolumn{2}{|c|}{SSE} & \multicolumn{2}{|c}{CE} \\\hline
    &$C_t$ & $C_g$ & $C_t$ & $C_g$ & $C_t$ & $C_g$ & $C_t$ & $C_g$  \\ \hline
    $[-1,1]$& 0.98922 &0.56534 &0.99846 &0.57332 & 0.96611 &0.60834 &0.98404& 0.61831 \\
   &(0.01097) &(0.04874) &(0.00395)&(0.04828)&(0.01829)& (0.04600) &(0.01334)& (0.04547) \\
    $[-10,10]$&  0.87408& 0.49537& 0.95444 &0.52401 & 0.74988 &0.46981 &0.70077& 0.44259 \\
   &(0.05040) &(0.05590) &(0.02668)&(0.05063)&(0.06487)& (0.06232) &(0.07736)& (0.06666) \\
    \hline
  \end{tabular}
  \end{center}
\end{table}

\end{document}